%% file: main.tex
\documentclass[accepted]{uai2021} %

\usepackage[british]{babel}
\usepackage[dvipsnames, table]{xcolor}

\usepackage{natbib} %
\usepackage{custom_commands}
\usepackage{hyperref}
\begin{document}
\title{Asynchronous \eps-Greedy Bayesian Optimisation}

\author[1]{George {De Ath}}
\author[1]{Richard M. Everson}
\author[1]{Jonathan E. Fieldsend}
\affil[1]{%
  Department of Computer Science\\
  University of Exeter\\
  Exeter, United Kingdom
}

\maketitle

\begin{abstract}
  Batch Bayesian optimisation (BO) is a successful technique for the 
  optimisation of expensive black-box functions. Asynchronous BO can reduce
  wallclock time by starting a new evaluation as soon as another finishes, thus
  maximising resource utilisation. To maximise resource allocation, we develop
  a novel asynchronous BO method, AEGiS (Asynchronous $\epsilon$-Greedy Global
  Search) that combines greedy search, exploiting the surrogate's mean
  prediction, with  Thompson sampling and random selection from the approximate
  Pareto set describing the trade-off between exploitation (surrogate mean
  prediction) and exploration (surrogate posterior variance).
  We demonstrate empirically the efficacy of AEGiS on synthetic benchmark
  problems, meta-surrogate hyperparameter tuning problems and real-world
  problems, showing that AEGiS generally outperforms existing
  methods for asynchronous BO. When a single worker is available
  performance is no worse than BO using expected improvement.
\end{abstract}

\section{Introduction}
Bayesian Optimisation (BO) is a popular sequential approach for optimising
time-consuming or costly black-box functions that have no closed-form 
expression or derivative information \citep{brochu:BO:2010,
snoek:practical:2012}. BO comprises two main steps which are repeated until
convergence or the budget is exhausted. Firstly, a surrogate model is
constructed using previous evaluations; this is typically a Gaussian process
(GP), due to its strength in uncertainty quantification and function 
approximation \citep{rasmussen:gpml:2006, shahriari:ego:2016}. Secondly, an
acquisition function is optimised to select the next location to expensively
evaluate. Such functions attempt to balance the exploitation of locations with
a good predicted value with the exploration of locations with high uncertainty.

In real-world problems it is often possible to use hardware capabilities to
run multiple evaluations in parallel. For example, the number of hyperparameter
configurations that can be evaluated in parallel when training neural networks
\citep{chen:alphago:2018} is only limited by the computational resources
available. The desire to speed up optimisation wallclock time through parallel
evaluation led to the development of parallel BO in which $q$ locations are
jointly selected to be evaluated synchronously. It has been successfully
applied in many areas such as drug discovery
\citep{hernandez-lobato:parTS:2017}, heat treatment, 
\citep{gupta:eespace:2018}, and hyperparameter configuration
\citep{kandasamy:ts:2018}.

When the runtime for evaluations varies, such as when optimising the
number of units in the layers of a neural network, or indeed the number of
layers themselves, synchronous parallel approaches will result in resource
underutilisation because all evaluations must finish before the next $q$
locations can be suggested. To counteract this, asynchronous BO approaches have
been proposed that allow for the evaluation of functions asynchronously; \ie,
as soon as an evaluation has completed, another can be submitted. In order to
take into account the uncertainty in the surrogate model at and around 
locations which are still being evaluated, a number of
methods have been proposed to penalise either the surrogate model
itself or the acquisition function in order to limit the selection of locations
near places that are currently under evaluation
\citep{ginsbourger:qei:2008, gonzalez:lp:2016, alvi:playbook:2019}.

\citet{kandasamy:ts:2018} presented an alternative approach based on 
Thompson sampling (TS) that draws surrogate function realisations from the GP posterior
and optimises these to suggest new locations to evaluate without the need to
penalise the model or acquisition function. However, TS relies on there being
enough stochasticity to create sufficiently diverse function draws with
meaningfully different optima. Therefore, TS tends to be insufficiently
exploratory in low dimensions (as draws are very similar to the mean
prediction) and over-exploratory in high dimensions where the posterior
uncertainty is large across much of the domain.

$\epsilon$-greedy methods have been shown to be remarkably successful in
sequential and synchronous parallel optimisation \citep{death:eshotgun:2020,
death:egreedy:2021}. Most of the time these exploit the mean prediction of
the surrogate model by expensively evaluating the location with the best
posterior mean, ignoring the uncertainty in the surrogate's prediction, but
making deliberately exploratory samples with probability $\epsilon$. In this work
we combine a greedy exploitation of the mean prediction with Thompson sampling
and random selection from the approximate Pareto set that maximally trades off
exploration and exploitation. Specifically, we introduce an $\epsilon$-greedy
scheme that, with probability $\epsilon_T$ performs  Thompson sampling or
with probability $\epsilon_P$ random selection from the approximate Pareto set, and exploits the surrogate
model's posterior mean prediction the rest of the time, \ie with probability
$1 - (\epsilon_T + \epsilon_P)$. For convenience we write $\epsilon =
\epsilon_T + \epsilon_P$. One of the reasons for the success of $\epsilon$-greedy
methods is that model inaccuracy introduces some accidental exploration even
when exploiting the mean prediction. This effect becomes dominant as the
dimension increases, and we therefore investigate how $\epsilon$ can be reduced
with increasing dimension.

\begin{itemize}[leftmargin=12pt]
\item We present AEGiS (Asynchronous $\epsilon$-Greedy Global Search), a
  novel asynchronous BO algorithm that combines greedy exploitation of the
  posterior mean prediction with Thompson sampling and random selection
  from the approximate Pareto set between exploration and exploitation.
\item We present results on fifteen well-known benchmark functions, three
  meta-surrogate hyperparameter tuning problems and two robot pushing
  problems. Empirically, we demonstrate that AEGiS outperforms existing
  methods for asynchronous BO on synthetic and hyperparameter optimisation
  problems.
\item We investigate how the degree of deliberate exploration $\epsilon$
  should be controlled with increasing dimension, $d$, showing empirically
  that $\epsilon = \min(2/\sqrt{d}, 1)$ yields superior performance to
  increased or decreased rates of exploration with respect to problem
  dimensionality.
\item We present an ablation study that verifies the importance of all 
  three components, namely the Pareto set selection, Thompson sampling, and
  exploitation, showing that AEGiS achieves performance that is
  greater than the sum of its individual components.
\item We show empirically that selection from the exploration-exploitation
  Pareto set is superior to selection from the entire problem domain. 
\end{itemize}

We begin in Section~\ref{sec:bo} by reviewing sequential, synchronous and
asynchronous BO, and Thompson sampling, before introducing AEGiS in 
Section~\ref{sec:AEGiS}. Section~\ref{sec:results} details extensive empirical
evaluation of AEGiS against other popular methods on well-known test problems,
and also includes an ablation study. We finish with concluding remarks in
Section~\ref{sec:conclusion}.

\section{Bayesian Optimisation}
\label{sec:bo}
Our goal is to minimise a black-box function $f : \mX \mapsto \Real$, defined
on a compact domain $\mX  \subset \Real^d$. The function itself is unknown, but
we have access to the result of its evaluation $f(\bx)$ at any location
$\bx \in \mX$. We are particularly interested in cases where the evaluations
are expensive, either in terms of time or money or both. We therefore seek to
minimise $f$ in either as few evaluations as possible to incur as little cost
as possible or for a fixed budget.

\subsection{Sequential Bayesian Optimisation}
\label{sec:sequential-bo}
Bayesian Optimisation (BO) is a global search strategy that sequentially
samples the problem domain $\mX$ at locations that are likely to contain the
global optimum, taking into account both the predictions of a probabilistic
surrogate model and its corresponding uncertainty \citep{jones:ego:1998}. It 
starts by generating $M$ initial sample locations $\{\bx_i\}_{i=1}^M$ with a
space filling algorithm, such as Latin hypercube sampling 
\citep{mckay:lhs:1979}, and expensively evaluates them with the function
$f_i = f(\bx_i)$. This set of observations $\Data_M = \{(\bx_n, f_n
\triangleq f(\bx_n)) \}_{n=1}^M$ is used for initial training of the
surrogate model.  Following  training, and at
each iteration of BO, the next location to be expensively evaluated is
chosen as the location that maximises an acquisition function
$\alpha(\bx)$. Thus $f(\cdot)$ is next evaluated at $\xnext = \argmax_{\bx
  \in \mX} \alpha(\bx)$.
The dataset is augmented with $\xnext$ and $f(\xnext)$ and the
process is repeated until the budget is exhausted. The value of the global
minimum is then estimated to be the best function evaluation seen during
optimisation, \ie $\fstar = \min_i \{f_i\}$.

\paragraph{Gaussian Processes}
\label{sec:gp}
Gaussian processes (GPs) are commonly used as the surrogate model for $f$.
A GP defines a prior distribution over functions, such that any finite number
of function values have a joint Gaussian distribution
\citep{rasmussen:gpml:2006} with mean $m(\bx)$ and a covariance
$\kappa(\bx, \bx' \given \btheta)$ with hyperparameters $\btheta$. Henceforth,
w.l.o.g. we use a zero-mean prior $m(\bx) = 0\, \forall\, \bx \in \mX$.
Conditioning the prior distribution on data consisting of $f$ evaluated at $t$
sampled locations $\Data_t$,
the posterior distribution of $f$ is also a GP with posterior mean and variance
\begin{align}
    \mu(\bx \given \Data, \btheta)
     & = m(\bx) + \bkappa( \bx, X ) K^{-1} \bff
    \label{eqn:gp:pred}                                                  \\
    \sigma^2(\bx \given \Data, \btheta)
     & = \kappa(\bx, \bx) - \bkappa( \bx, X)^\top K^{-1} \bkappa(X, \bx).
    \label{eqn:gp:var}
\end{align}
Here, $X \in \Real^{t \times d}$ and $\bff = (f_1, f_2, \dots, f_t)^\top$ are
the matrix of design locations and corresponding vector of function evaluations
respectively.
The kernel matrix $K \in \Real^{t \times t}$ is 
$K_{ij} = \kappa(\bx_i, \bx_j \given \btheta)$ and
$\bkappa(\bx, X) \in \Real^t$ is given by
$[\bkappa(\bx, X)]_i = \kappa(\bx, \bx_i \given \btheta)$.
Kernel hyperparameters $\btheta$ are learnt by maximising the log marginal likelihood
 \citep{rasmussen:gpml:2006}.

\paragraph{Acquisition Functions}
The choice of where to evaluate next in BO is determined by an acquisition
function ${\alpha : \Real^d \mapsto \Real}$. Perhaps the most commonly used
acquisition function is Expected Improvement (EI) \citep{jones:ego:1998}, which
is defined as the expected positive improvement over the best-seen 
function value $\fstar$ thus far:
\begin{equation}
    \label{eqn:ei}
    \alpha_{\text{EI}}(\bx) = \mathbb{E}_{p(f (\bx) \given \Data)}
        \left[ \max\left( \fstar - f(\bx), \, 0 \right) \right].
\end{equation}
However, many other acquisition functions have also been proposed,
including probability of improvement \citep{kushner:ego}, 
optimistic strategies such as UCB \citep{srinivas:ucb:2010}, 
$\epsilon$-greedy strategies \citep{bull:convergence:2011, death:egreedy:2021},
and information-theoretic approaches \citep{scott:kg:2011, hennig:es:2012, 
henrandez-lobato:pes:2014, wang:mes:2017, ru:fitbo:2018}.

\subsection{Parallel Bayesian Optimisation}
\label{sec:parallel-bo}
In parallel BO, the algorithm has access to $q$ workers that can
evaluate $f$ in parallel and the focus is on minimising the optimisation
time, acknowledging that the total computational cost will be greater than
for strictly sequential evaluations. The goal is to select a batch of $q$ 
promising locations to be expensively evaluated in parallel. In the synchronous
setting, the BO algorithm sends out a batch of $q$ locations to be evaluated in
parallel, and waits for all $q$ evaluations to be completed before submitting
the next batch. However, in the asynchronous setting, as soon as a worker has
finished evaluating $f$ a new location is submitted to it for evaluation. 
Therefore, if $f$ takes a variable time to evaluate, the asynchronous version
of BO will result in more evaluations of $f$ for the same wallclock time.

One of the earliest synchronous approaches was the qEI method of
\citet{ginsbourger:qei:2008}, a generalisation of the sequential EI acquisition
function~\eqref{eqn:ei} that jointly proposes a batch of $q$ locations to 
evaluate. Similarly, the parallel predictive entropy search
\citep{shah:ppes:2015} and the parallel knowledge gradient \citep{wu:pkg:2016}
also jointly propose $q$ locations and are based on their sequential
counterparts. However, these methods require the optimisation of a $d \times q$
problem that becomes prohibitively expensive to solve as $q$ increases
\citep{daxberger:dgpo:2017}.

Consequently, the prevailing strategy has become to  sequentially select the
$q$ batch locations. Penalisation-based methods penalise the regions of either
the surrogate model \citep{azimi:bbo:2010, desautels:ee:2014} or the 
acquisition function that correspond to locations that are currently under
evaluation. The Kriging Believer method of \citet{ginsbourger:pbo:2010}, for
example, hallucinates the results of pending evaluations by using the surrogate
model's mean prediction as its observed value. The local penalisation methods
of \citet{gonzalez:lp:2016} and \citet{alvi:playbook:2019} directly penalise an
acquisition function. \citet{death:eshotgun:2020} show that an 
$\epsilon$-greedy method that usually exploits the most promising region, with
occasional exploratory moves, is effective.

Asynchronous BO has received less attention. \citet{janusevskis:aEI:2012}
proposed an asynchronous version of the qEI method that uses the qEI criterion
assuming that the $p$ locations currently under evaluation are fixed, and
optimises the position of the remaining $q-p$ locations.
\citet{kandasamy:ts:2018} suggest using Thompson sampling to
select locations to evaluate. Thompson sampling (TS) \citep{thompson:ts:1933} 
is a randomised strategy for sequential decision-making under uncertainty
\citep{kandasamy:ts:2018}. At each iteration, TS selects the next location to
evaluate $\xnext$ according to the posterior probability that it is the 
optimum. If $p_*(\bx \given \Data)$ denotes the probability that $\bx$ 
minimises the surrogate model posterior, then
\begin{align}
p_*(\bx \given \Data) &= \int p_*(\bx \given g) \, p(g \given \Data) \, dg  \\
                      &= \int \delta ( \bx - \argmin_{\bx \in \mX} g(\bx)) 
                         \, p(g \given \Data) \, dg,
\end{align}
where $\delta$ is the Dirac delta distribution. Thus, all the probability mass
of $p_*(\bx \given \Data)$ is at the minimiser of $g$. In the context of BO,
this corresponds to sampling a function $g(\cdot)$ from the surrogate model's
posterior $p(\cdot \given \bx, \Data)$ and selecting
$\xnext = \argmin_{\bx \in \mX} g(\bx)$ as the next location to evaluate. This
can then be trivially applied in the sequential, synchronous and asynchronous
settings by sampling and minimising as many function realisations as required.
Thus, the BO loop using TS involves building a surrogate model with previously
evaluated locations (ignoring those that may be currently under evaluation),
drawing as many function realisations as there are free workers, finding the
minimiser of each realisation and evaluating $f$ at those locations.

\citet{kandasamy:ts:2018} analyse Thompson sampling in terms of maximum
information gain. They show that both synchronous and asynchronous TS
algorithms making $n$ function evaluations are almost as good as if the $n$
evaluations were made sequentially. Furthermore, they extend the notion of
simple regret to incorporate the (random) number of evaluations made in a
given time. Under this metric they show that asynchronous TS outperforms
synchronous and sequential algorithms using a variety of evaluation-time
models.

Traditionally, $g$ has been approximately optimised by picking the minimum of a
randomly sampled set of discrete points \citep{kandasamy:ts:2018}. Recently,
\citet{wilson:samplepath:2020} have identified an elegant decoupled sampling
method, which we use here, that decomposes the GP posterior into a sum of a
weight-space prior term that is approximated by random Fourier features, and a
pathwise update in function space. Unlike other fast approximations, such as
\citep{rahimi:rff:2008}, this method does not suffer from variance starvation
\citep{calandriello:gpo:2019}. It scales linearly with the number of locations
queried during optimisation of $g$, and is pathwise differentiable, meaning
that gradient-based optimisation can be employed.

Recently, \citet{depalma:sampTS:2019} proposed to draw realisations of
acquisition functions instead, by drawing a new set of surrogate model
hyperparameters from their posterior distribution for each acquisition
function. An advantage of these TS-based methods is that each draw from either
the model or hyperparameter posterior defines a new distinct acquisition
function. A drawback is that they rely on the uncertainty in the surrogate
model to give sufficient diversity in position to the locations selected,
unlike acquisition functions which implicitly or explicitly balance exploration
and exploitation.

\section{AEGiS}
\label{sec:AEGiS}

It is well recognised that the fundamental problem in global optimisation is to
balance exploitation of promising locations found thus far with exploration of
new areas which have not been explored. While extremely appealing in its
simplicity, Thompson sampling (TS) tends to over-exploit in low dimensions
(because after a number of evaluations the posterior variance is reduced and
most sampled realisations are similar to the posterior mean), but under-exploit
in high dimensions (because there are insufficient samples to drive the
posterior variance away from the prior, resulting in effectively random 
search). The efficacy of $\epsilon$-greedy methods in sequential and 
synchronous BO over a range of dimensions \citep{death:eshotgun:2020,
death:egreedy:2021} motivates an $\epsilon$-greedy strategy to switch between
greedy exploitation of the surrogate model, exploration, and TS. Specifically,
at each step we optimise a TS function draw with probability $\epsilon_T$ or perform exploration each with
probability $\epsilon_P$, and exploit the surrogate model's posterior mean
prediction $\mu(\bx)$ of the rest of the time, \ie with probability 
$1 - (\epsilon_T + \epsilon_P)$.
\input{algorithm.tex}

The AEGiS method is outlined in Algorithm \ref{alg:ets}. With probability
$1-(\epsilon_T + \epsilon_P)$, the mean prediction from the surrogate model is minimised to
find the next location $\xnext$ to evaluate (line \ref{alg:optimise-mu}).
Alternatively, with probability $\epsilon_T$, a TS step taken: the next location
is determined by sampling a realisation~$g(\cdot)$ from the surrogate posterior
and finding its minimiser (lines~\ref{alg:ts-sample} and~\ref{alg:optimise-ts}). 
Otherwise (lines \ref{alg:eS:moo_optimise} and
\ref{alg:eS:choice_PF}), with probability $\epsilon_P$, the location to evaluate
is chosen from the approximate Pareto set of all locations which maximally
trade off between exploitation (minimising~$\mu(\bx)$) and exploration
(maximising~$\sigma^2(\bx)$). A full discussion is given by
\cite{death:egreedy:2021}. Briefly, a location $\bx_1$ dominates $\bx_2$
(written $\bx_1 \succ \bx_2$) iff $\mu(\bx_1) \le \mu(\bx_2)$ and 
$\sigma(\bx_1) \ge \sigma(\bx_2)$, and at least one of the inequalities is
strict. If both $\bx_1 \nsucc \bx_2$ and $\bx_2 \nsucc \bx_1$ then $\bx_1$
and $\bx_2$ are said to be mutually non-dominating. The maximal set of
mutually non-dominating solutions is known as the Pareto set
$\mathcal{P} = \{ \bx \in \mX \given \bx' \not\succ \bx \, \forall \bx' \in
\mX \}$ and comprises the optimal trade-off between exploratory and
exploitative locations. Since it is infeasible to enumerate $\mX$, we
sample a location from the approximate Pareto set 
$\Papprox \approx \mathcal{P}$ which is found using a standard multi-objective
optimiser (lines~\ref{alg:eS:moo_optimise}~and~\ref{alg:eS:choice_PF}).

Selecting a random location from $\Papprox$ ensures that no other location
could have been selected that has either more uncertainty given the same
predicted value or a lower predicted value given the same amount of
uncertainty. Therefore, when considering the two objectives of both reducing
the overall model uncertainty and finding the location with the minimum value,
these locations are optimal with respect to a specific exploitation-exploration
trade-off. Indeed, the maximisers of both EI and UCB are members of
$\mathcal{P}$ \citep{death:egreedy:2021}.

Rather than selecting from $\Papprox$, a more exploratory procedure is to
select from the entire feasible space $\mX$. That is, lines
\ref{alg:eS:moo_optimise} and \ref{alg:eS:choice_PF} are replaced with
$\xnext \gets \mathtt{randomChoice}(\mX)$. We denote this %
variant  by AEGiS-RS. However, as shown below, it is less effective than
selection from $\Papprox$.

At the beginning of an optimisation run, the initial $q$ locations to be
evaluated are chosen by first selecting the most exploitative location
(line~\ref{alg:optimise-mu}). The remaining $q - 1$ locations are each chosen
by performing either Thompson sampling or selection from the approximate Pareto
set $\Papprox$ with probabilities $\epsilon_T / (\epsilon_T + \epsilon_P)$ and
$\epsilon_P / (\epsilon_T + \epsilon_P)$ respectively. This ensures that the
most exploitative location is only chosen once during initialisation. The
remaining evaluated locations in an optimisation run are chosen according to
Algorithm~\ref{alg:ets}.

As discussed above, inaccuracies in the surrogate model provide an element of
exploration even when disregarding the uncertainty in the posterior
$p(f \given \Data_t)$. On the other hand, TS tends to be over-exploitative in
low dimensions ($d \lesssim 4$) and too exploratory in high dimensions. The
performance of sequential $\epsilon$-greedy algorithms is relatively
insensitive to the precise choice of $\epsilon$ \citep{death:egreedy:2021}.
As we show in section \ref{sec:results:additional}, the performance of
AEGiS is relatively insensitive to the ratio of $\epsilon_T : \epsilon_P$.
For simplicity, we therefore initially set $\epsilon_T = \epsilon_P = \epsilon/2$ and 
$\epsilon = \min(2/\sqrt{d}, 1)$.  This allows for more
exploitation to occur as the dimensionality of the problem increases. As shown
empirically in section \ref{sec:results:additional}, this rate of decay in the
amount of exploration carried out with respect to the problem dimensionality is
more effective than faster or slower rates.

The dominant contributor to computational cost of finding the approximate Pareto set,
carried out by NSGA-II \citep{deb:nsga2:2002}, is the non-dominated sort of the
population of solutions and their offspring. With two objectives, as
is the case in AEGiS, this has complexity
$\mathcal{O}(G N \log N)$ \citep{jensen:nsga2runtime:2003}, where $N$ is the
total population size and $G$ is the number of generations for which the
evolutionary algorithm runs. Empirically we find that this cost is similar
to that of TS, whose computational cost is dominated by the gradient descent
method used to optimise the sample path $g$; \eg $\mathcal{O}(L d^2)$ for BFGS
\citep{byrd:lbfgs:1995}, where $L$ is the maximum number of steps taken in
the search. However, we
note that both of these costs are negligible compared to fitting the GP.

\section{Experimental Evaluation}
\label{sec:results}
\begin{figure*}[t] %
\centering%
\includegraphics[width=0.495\linewidth, clip, trim={5 44 5 7}]{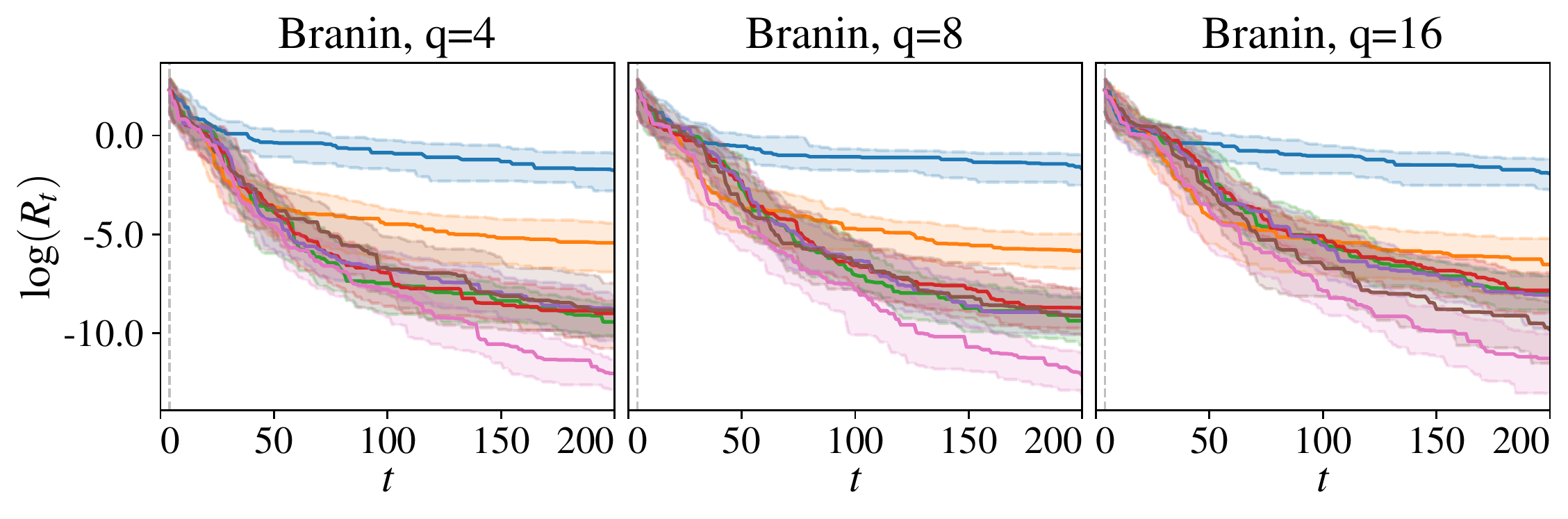}\hfill%
\includegraphics[width=0.495\linewidth, clip, trim={5 44 5 7}]{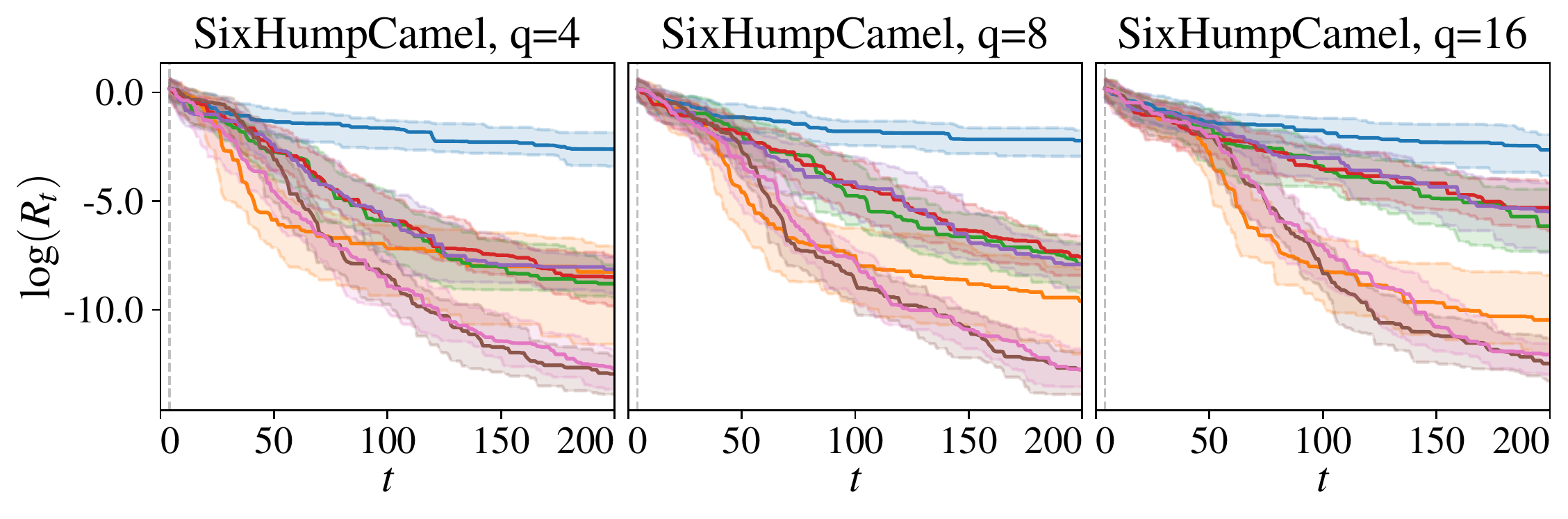}\\[1mm]
\includegraphics[width=0.495\linewidth, clip, trim={5 0 5 7}]{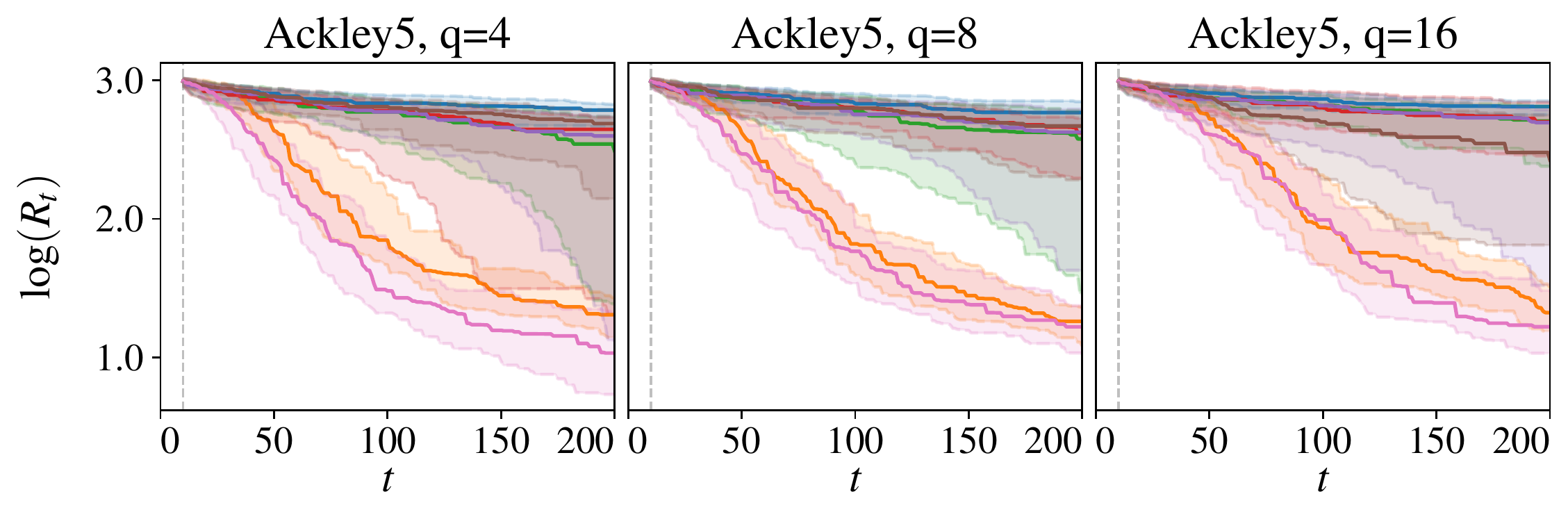}\hfill%
\includegraphics[width=0.495\linewidth, clip, trim={5 0 5 7}]{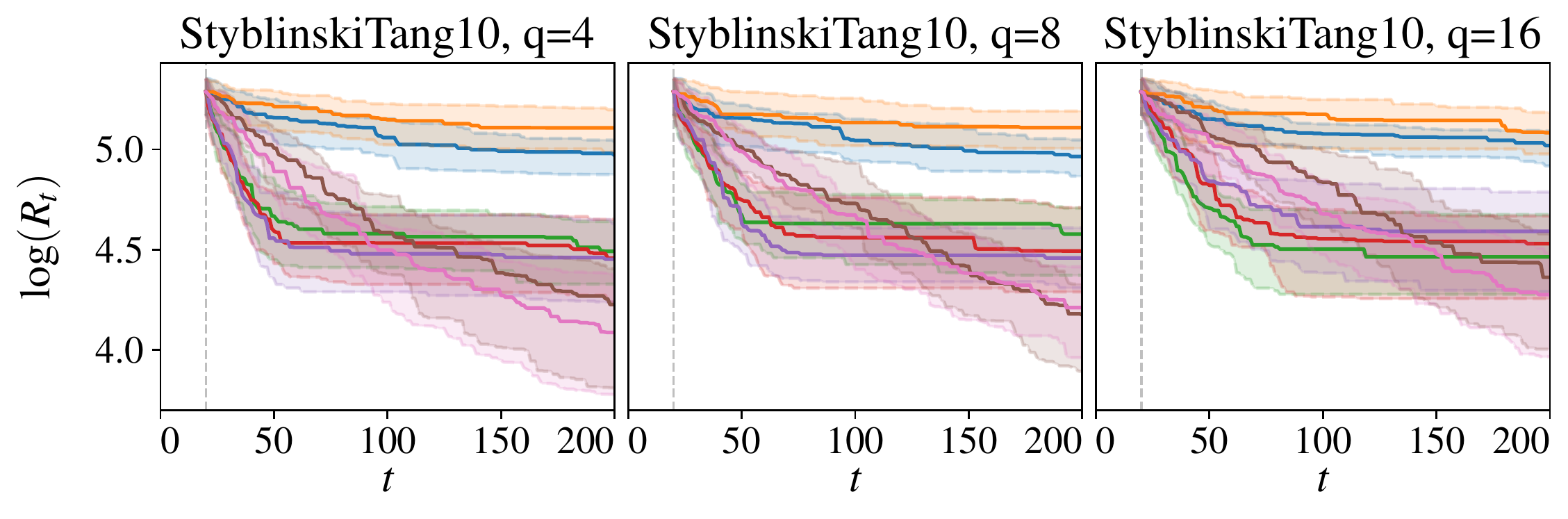}\\
\includegraphics[width=0.85\linewidth, clip, trim={10 15 10 13}]{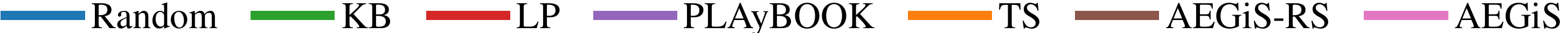}%
\caption{Illustrative convergence plots for four benchmark problems with
$q \in \{4,8,16\}$ asynchronous workers. Each plot shows the median log simple
regret, with shading representing the interquartile range over $51$ runs.
}
\label{fig:synthetic:conv}
\end{figure*}
We compare AEGiS with four well-known asynchronous BO algorithms: two acquisition
function-based penalisation methods, Local Penalisation (LP)
\citep{gonzalez:lp:2016} and the more recent PLAyBOOK
\citep{alvi:playbook:2019}, as well as the Kriging Believer (KB)
\citep{ginsbourger:pbo:2010}, all of which are based on the expected
improvement acquisition function~\eqref{eqn:ei}. We also include the standard
Thompson sampling (TS) approach of \citet{kandasamy:ts:2018} and a purely
random approach using Latin hypercube sampling (Random). The  
optimisation pipeline was constructed using GPyTorch
\citep{gardner:gpytorch:2018} and BoTorch \citep{balandat:botorch:2020}, 
and we have made this available
online\footnote{\url{http://www.github.com/georgedeath/aegis}}.

A zero-mean Gaussian process surrogate model with an isotropic \Matern~$5/2$
kernel, as recommended by \citet{snoek:practical:2012} for modelling realistic functions,  was used in all the experiments. Input variables were rescaled to reside
in $[0, 1]^d$ and observations were rescaled to have zero-mean and unit variance.
The models were initially trained on $M = 2d$ observations generated by maximin
Latin hypercube sampling and then optimised for a further $200 - 2d$ function
evaluations. Each optimisation run was repeated $51$ times from a different set
of Latin hypercube samples. Sets of initial locations were common across all
methods to enable statistical comparison. At each iteration, before new
locations were selected, the hyperparameters of the GP were optimised by
maximising the log likelihood using a multi-restart strategy
\citep{shahriari:ego:2016} with L-BFGS-B \citep{byrd:lbfgs:1995} and $10$
restarts (Algorithm~\ref{alg:ets}, line~\ref{alg:eS:fitgp}).

All experiments
were repeated for $q \in \{4, 8, 16\}$ asynchronous workers. We  followed the
procedure of \citet{kandasamy:ts:2018} to sample an evaluation time for each
task, thereby simulating the asynchronous setting. A half-normal distribution 
was used for this with a scale parameter of $\sqrt{\pi / 2}$, giving a mean 
runtime of 1 \citep{alvi:playbook:2019}.

The KB, LP and PLAyBOOK methods all used the EI acquisition function,
which was selected because of its popularity. The expected improvement, the sampled function from the
GP posterior in both TS and AEGiS, and the posterior mean in AEGiS were all
optimised using the typical strategy
\citep{balandat:botorch:2020} of uniformly sampling $1000d$ locations in $\mX$,
optimising the best $10$ of these using L-BFGS-B, and selecting the best as the
optimal location. In AEGiS and TS, $2000$ Fourier features were used for the
decoupled sampling method of \citet{wilson:samplepath:2020}. The approximate
Pareto set $\Papprox$ in AEGiS was found using NSGA-II
\citep{deb:nsga2:2002}; see supplementary material.

Here, we report performance in terms of the logarithm of the simple regret
$R_t$, which is the difference between the true minimum value $f(\bx^*)$ and
the best seen function evaluation up to iteration $t$:
$\log(R_t) = \log (\min \{ f_1, \dots, f_t \} - f(\bx^*) )$.

\subsection{Synthetic Experiments}
\label{sec:results:synthetic}
\begin{table}[t]
\centering
\caption{Benchmark Functions and Dimensionality.}
\label{tbl:function_details}
\begin{tabular}[t]{lr p{4pt} lr}
\addlinespace[-\aboverulesep]%
    \cmidrule[\heavyrulewidth]{1-2}%
    \cmidrule[\heavyrulewidth]{4-5}%
    \textbf{Name}  & $d$ && \textbf{Name}  & $d$      \\
\cmidrule[\lightrulewidth]{1-2}\cmidrule[\lightrulewidth]{4-5}
    Branin         & 2   && Ackley         & 5, 10    \\
    Eggholder      & 2   && Michalewicz    & 5, 10    \\
    GoldsteinPrice & 2   && StyblinskiTang & 5, 7, 10 \\
    SixHumpCamel   & 2   && Hartmann6      & 6        \\
    Hartmann3      & 3   && Rosenbrock     & 7, 10    \\
    \cmidrule[\heavyrulewidth]{1-2}%
    \cmidrule[\heavyrulewidth]{4-5}%
\addlinespace[-\belowrulesep]
\end{tabular}
\end{table}
The methods were evaluated on the $15$ synthetic benchmark functions listed in
Table~\ref{tbl:function_details}.\footnote{Formulae for all functions can be
found at: \url{http://www.sfu.ca/~ssurjano/optimization.html}.}
Figure~\ref{fig:synthetic:conv} shows four illustrative convergence plots for
the Branin, SixHumpCamel, Ackley, and StyblinskiTang benchmark functions for
$q \in \{4,8,16\}$ asynchronous workers. Convergence plots and tabulated
results for all functions are available in the supplementary material. As can
be seen from the four sets of convergence plots, AEGiS (pink) is almost always better 
than TS (orange) on all four problems and values of $q$. Contrastingly, TS is
sometimes worse than random search (blue); this  is particularly evident on the
StyblinskiTang problem. This highlights the efficacy of both additional
exploration and exploitation in AEGiS relative to the TS algorithm. The
SixHumpCamel convergence plots show an interesting trend with respect to the
penalisation-based methods, namely that they deteriorate at an accelerated
rate in comparison to the TS-based methods. We suspect that this is because
errors in estimating the correct degree of penalisation accumulate more
noticeably as $q$ increases and therefore more function evaluations are placed
in unpromising regions.

\begin{figure}[t] %
\includegraphics[width=\linewidth, clip, trim={7 7 7 7}]{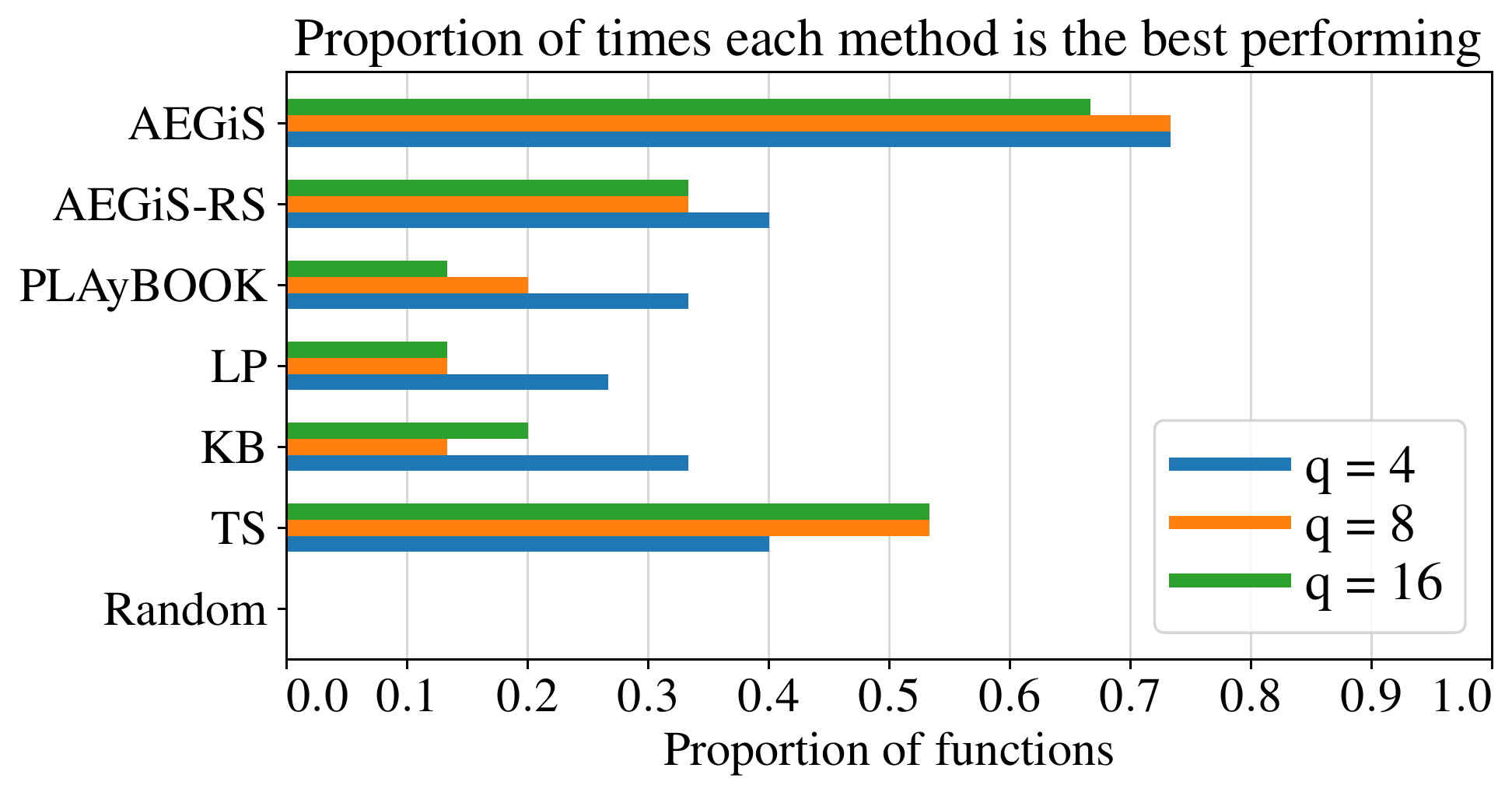}%
\caption{Synthetic function optimisation summary. Bar lengths correspond to the
proportion of times that a method is best or statistically equivalent to the 
best method across the 15 synthetic functions, for $q \in \{4,8,16\}$
 workers.}
\label{fig:bestmethods}
\end{figure}

Figure~\ref{fig:bestmethods} summarises the performance of each of the
evaluated methods for different numbers of workers~($q$). It shows the number
of times each method is the best, \ie has the lowest median regret over the
$51$ optimisation runs, or is statistically indistinguishable from the best
method, according to a one-sided, paired Wilcoxon signed-rank test
\citep{knowles:testing:2006} with Holm-Bonferroni 
correction~\citep{holm:test:1979} (${p \geq 0.05}$). As is clear from the 
figure, AEGiS achieves a strong level of performance, and is the best (or
equivalent to the best) on $10$ out of the $15$ functions. AEGiS-RS, which
samples exploratory locations from the entire feasible space,  is at best
equivalent to AEGIS and generally worse. This indicates that using a more
informative random sampling scheme, such as selecting from approximate Pareto
set of the trade-off between exploitation ($\mu(\bx)$) and exploration
($\sigma^2(\bx)$), provides a meaningful improvement to performance.
In contrast to both AEGiS and TS, which barely decrease in performance as the
batch size $q$ increases, the EI-based methods (PLAyBOOK, LP and KB) show a
much larger reduction in relative performance.

\subsection{Hyperparameter Optimisation}
\label{sec:results:meta}
Like \citet{souza:priorBO:2020}, we optimise the hyperparameters of three
meta-surrogate optimisation tasks corresponding to a SVM, a fully connected neural network (FC-NET) and  XGBoost.
These  are part of the Profet benchmark \citep{klein:profet:2019, paleyes:emukit:2019} and
are drawn from generative models built using performance data on multiple
datasets. They aim to mimic the landscape of expensive hyperparameter
optimisation tasks, preserving the global landscape characteristics of the
modelled methods and giving local variation between function draws from the
generative model. This allows far more evaluations to be carried out than
would be possible with the modelled problems. 

Here, we optimise $51$ instances of each meta-surrogate, repeating the
optimisation $N=20$ times per instance with different paired training data for
each of the $N$ runs. Details of the construction of each problem are given in
the supplementary material. The SVM problem had $2$ parameters; the FC-NET
problem $6$; and the XGBoost problem $8$.

\begin{figure}[t] %
\centering%
\includegraphics[width=\linewidth, clip, trim={5 43 5 7}]{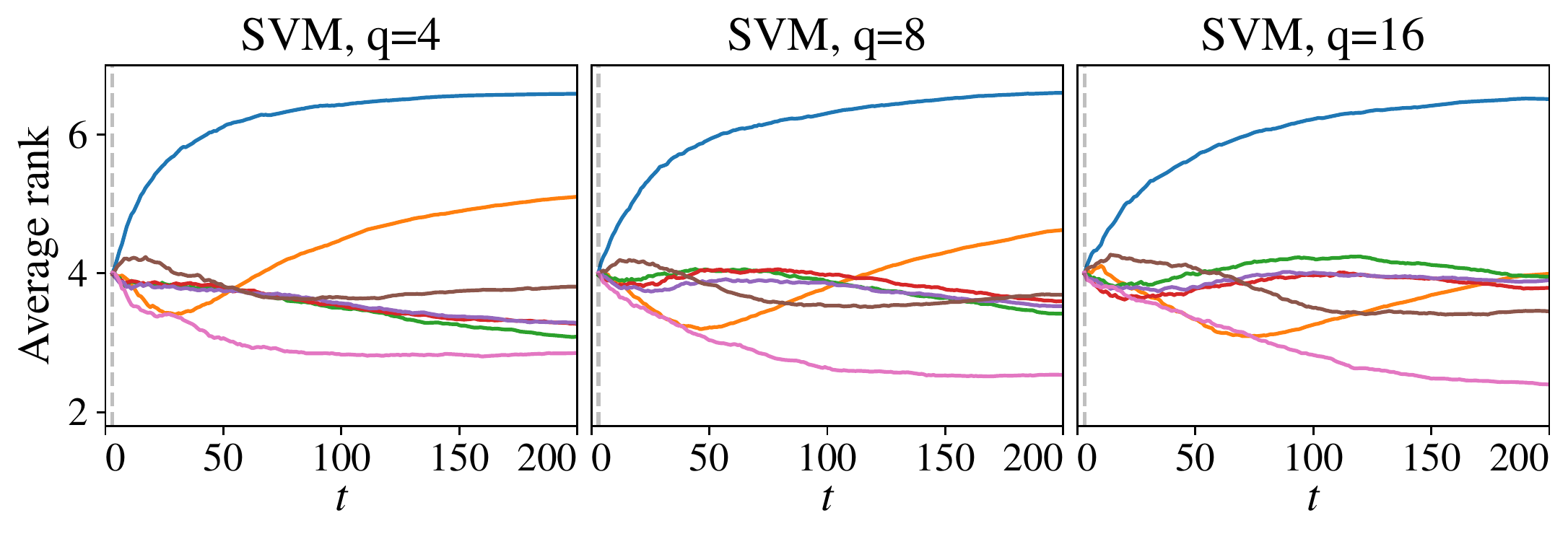}\\
\includegraphics[width=\linewidth, clip, trim={5 43 5 7}]{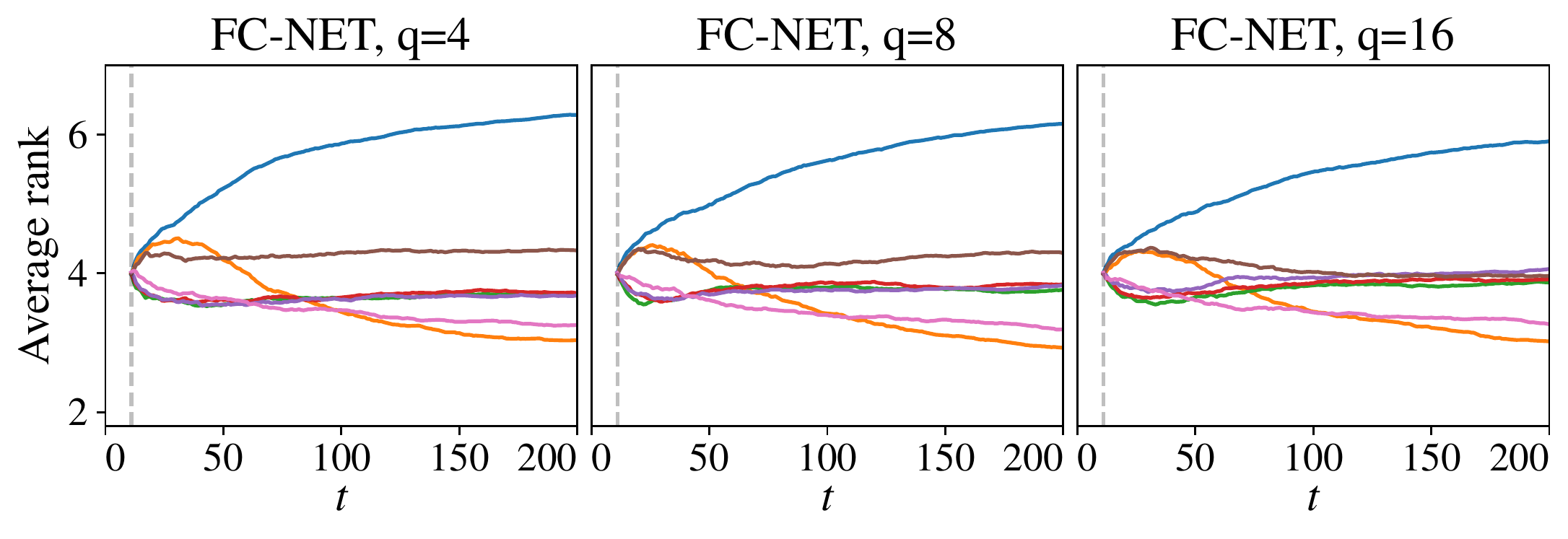}\\
\includegraphics[width=\linewidth, clip, trim={5 7 5 7}]{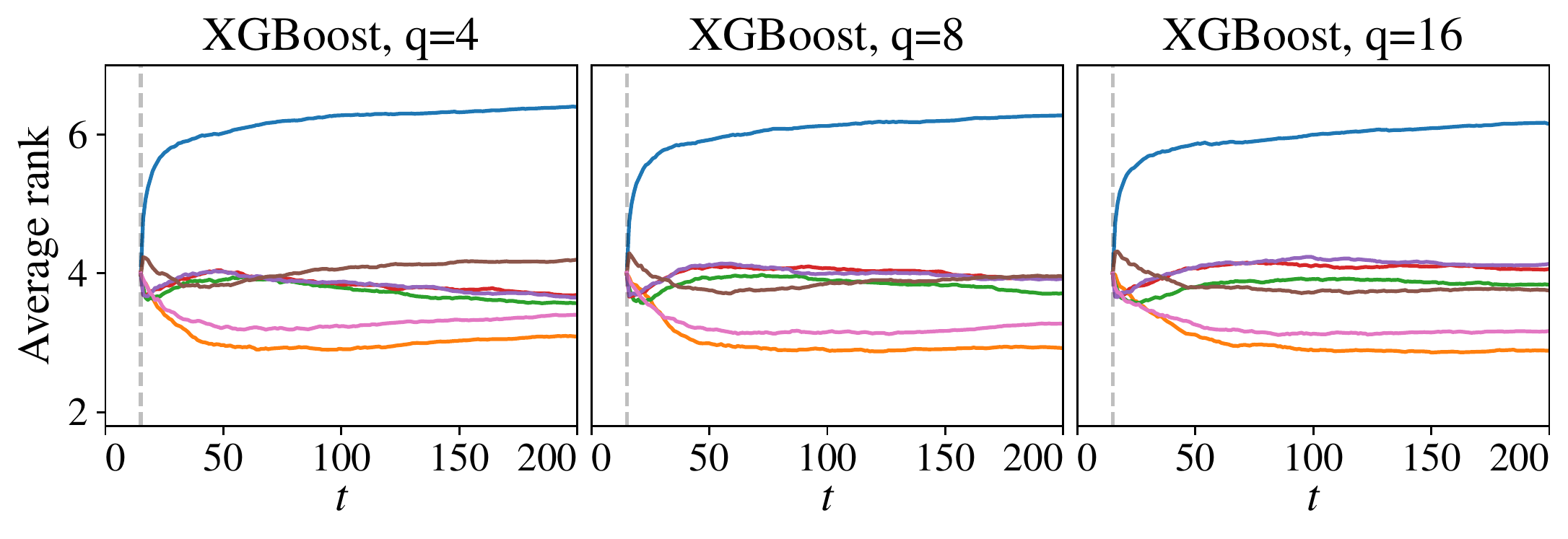}\\
\includegraphics[width=1.\linewidth, clip, trim={10 15 10 13}]{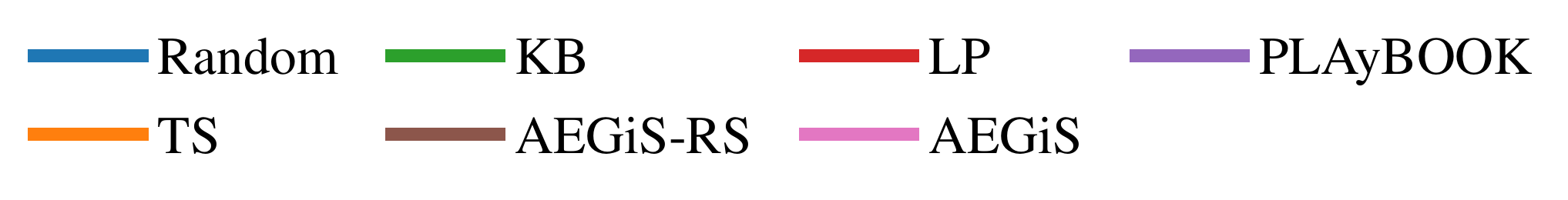}%
\caption{Average ranking scores for the three meta-surrogate benchmarks for 
$q \in \{4,8,16\}$.}
\label{fig:instance:ranks}
\end{figure}
We compare each evaluated method by computing the average ranking score in
every iteration for each problem instance. We follow 
\citet{feurer:hyperparams:2015} and compute this by, for each problem instance,
drawing a bootstrap sample of $1000$ runs out of the $7^N$ possible
combinations and calculating the fractional ranking after each of the $200$
iterations. The ranks are then averaged over all $51$ problem instances.

Figure~\ref{fig:instance:ranks} shows the average ranks of the methods on all
three problems for $q \in \{4, 8, 16\}$. Similarly to the synthetic functions,
all three penalisation methods (KB, LB and PLAyBOOK) perform comparably and 
AEGiS is consistently superior to them after roughly 50 function evaluations.
On the FC-NET and XGBoost problems TS performs similarly to AEGiS, and indeed
is better on the XGBoost problem.  However, AEGiS-RS performs consistently 
worse, supporting the contention that choosing random locations that lie in the 
Pareto set is  beneficial.

\subsection{Active Learning for Robot Pushing}
\label{sec:results:robots}
Following \citet{wang:mes:2017} and \citet{death:egreedy:2021}, we optimise the
control parameters for two instance-based active learning robot pushing
problems \citep{wang:pushing:2018}. In push4, a robot pushes an
object towards an unknown target location. It receives the object-target
distance once it has finished pushing. The location of the robot, the
orientation of its pushing hand and for how long it pushes are the parameters
to be optimised with the goal of minimising this distance. The push8 problem
has two robots that receive the distance between their respective objects and
targets after pushing --- we minimise the sum of these distances. In both
problems the target location(s) are chosen randomly, with the constraint that
they cannot overlap. We note that the push8 problem is much more difficult
because the robots can block each other, and so each problem instance may not
be completely solvable. Further details of the problems are provided in the
supplementary material.

\begin{figure}[t] %
\centering%
\includegraphics[width=\linewidth, clip, trim={5 42 5 7}]{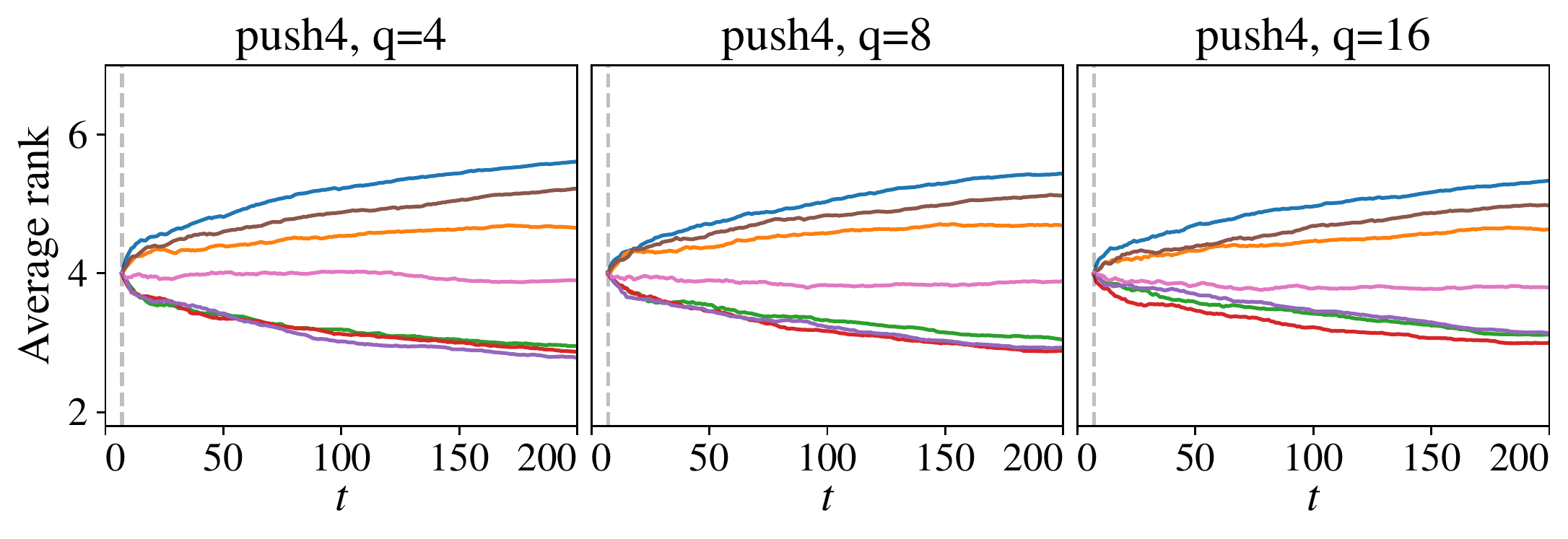}\\
\includegraphics[width=\linewidth, clip, trim={5 7 5 7}]{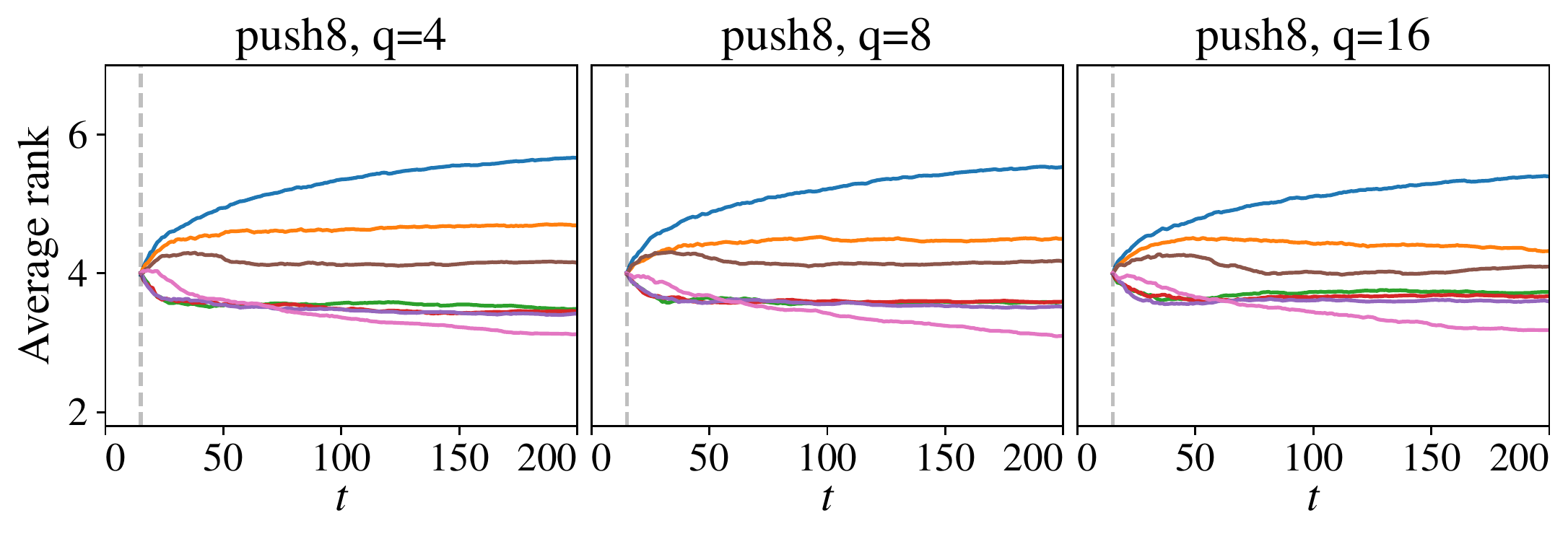}\\
\includegraphics[width=1.\linewidth, clip, trim={10 15 10 13}]{figs/legend_onecol}%
\caption{Average ranking scores for the two robot pushing problems for 
$q \in \{4,8,16\}$.}
\label{fig:instance:push}
\end{figure}
Average rank score plots, calculated in the same way as described in 
Section~\ref{sec:results:meta}, are shown in Figure~\ref{fig:instance:push}.
The EI-based methods continue to have very similar performance to one another,
and are the best methods on push4, and, even though it is not the best method,
AEGiS still out-performs TS consistently. On push8, the EI-based methods
initially improve on AEGiS in terms of average rank but then stagnate and are
overtaken by AEGiS after roughly 75 function evaluations.

\subsection{Additional Experiments}
\label{sec:results:additional}
Lastly, we describe an ablation study, experiments on setting the degree of
deliberate exploration ($\epsilon = \epsilon_T +\epsilon_P$), together
with the optimal ratio $\epsilon_T : \epsilon_P$. Finally, we compare
AEGiS, EI and TS in the sequential BO setting ($q=1$). See supplementary
material for full results.

\paragraph{\textbf{Ablation Study}}
\begin{figure}[t]
\includegraphics[width=\linewidth, clip, trim={7 7 7 7}]
                {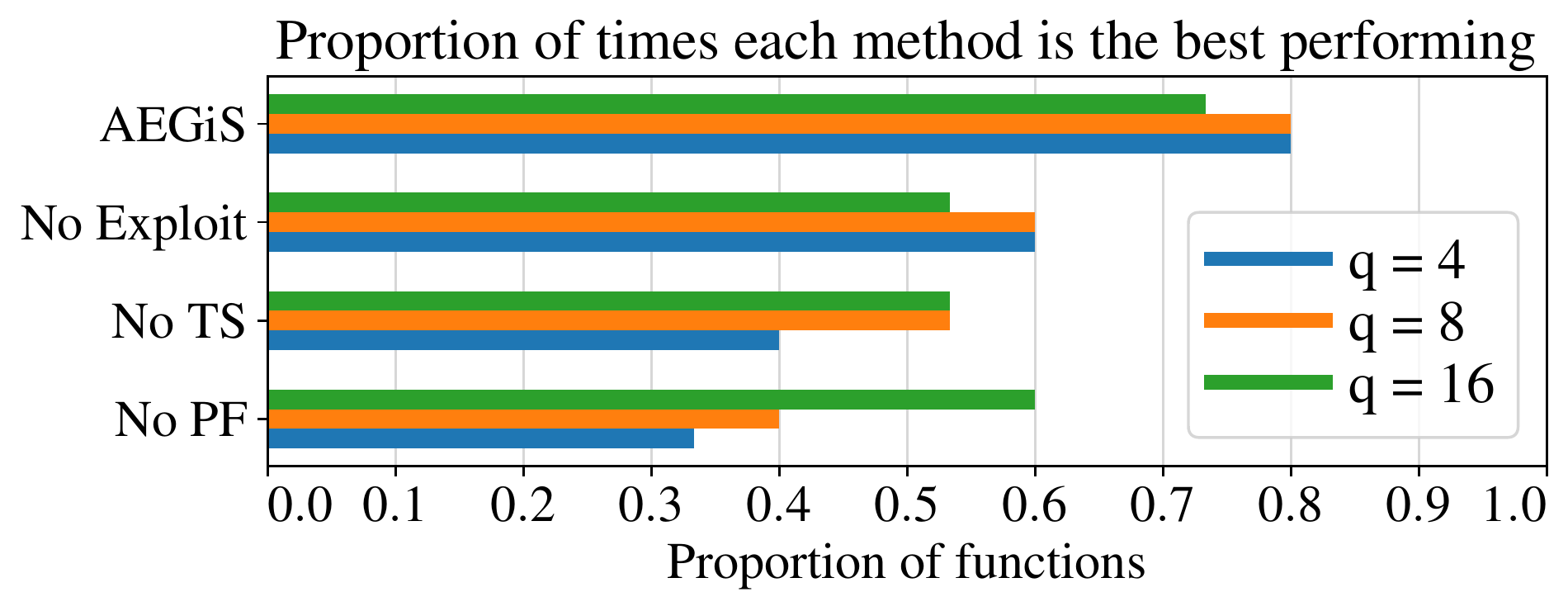}%
\caption{Ablation study summary. The bars correspond to the proportion of times
that a method is best or statistically equivalent to the best method across the
$15$ synthetic functions, for $q \in \{4,8,16\}$.}
\label{fig:ablation:bestmethods}
\end{figure}
Here, we conduct an ablation study on AEGiS using the $15$ synthetic benchmark
functions used in Section~\ref{sec:results:synthetic}. Since AEGiS-RS is
consistently outperformed by AEGiS, we omit it from this study. We compare to
AEGiS in the following ways:  \emph{No Exploit}, without the exploitation ($\epsilon_T =
\epsilon_P = 1/2$); \emph{No TS}, with no Thompson sampling ($\epsilon_T =
0; \epsilon_P = \epsilon$);
\emph{No PF}, without Pareto set selection ($ \epsilon_T =
\epsilon, \epsilon_P = 0$).  In all cases $\epsilon = \min(2 / \sqrt{d}, 1)$.

Figure~\ref{fig:ablation:bestmethods} shows that all three components combine
to give the best results and that removing any of the three results in an
inferior algorithm. As can be seen, the removal of either TS or Pareto set
selection considerably reduces the performance of the algorithm. Note that the
results for \emph{No Exploit} are inflated because the method is equivalent to
AEGiS on the 5 out of the 15 benchmark functions that have $d \leq 4$ dimensions.

\begin{figure}[t] %
\includegraphics[width=\linewidth, clip, trim={7 7 7 7}]{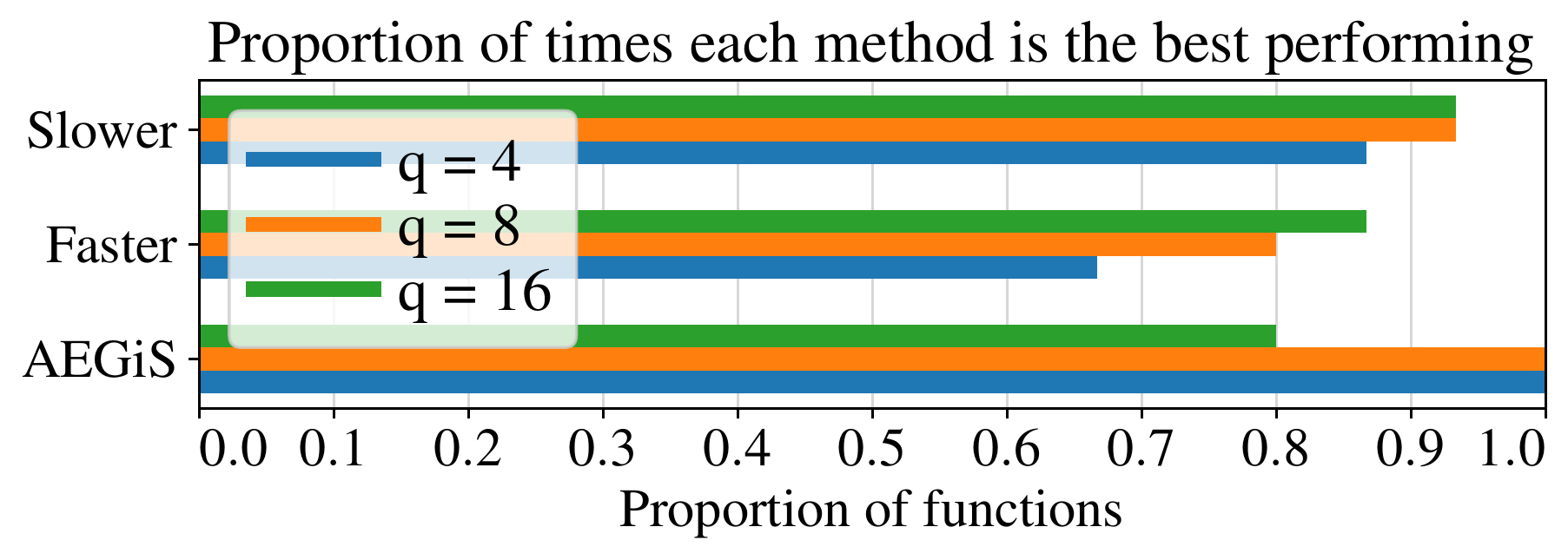}%
\caption{Selecting $\epsilon$: \emph{faster} and \emph{slower} correspond to 
quicker and slower rates of $\epsilon$ decay than AEGiS, resulting in more or
less exploitation respectively.}
\label{fig:bracketing}
\end{figure}

\paragraph{\textbf{Setting $\epsilon$}}
We investigate the rate at which the degree of deliberate exploitation of the
surrogate model's posterior mean function should increase with problem
dimensionality $d$. In AEGiS, exploitation is carried out 
$1 - (\epsilon_T + \epsilon_P)$ proportion of the time. As above we chose
$\epsilon_T = \epsilon_P = \epsilon/2$ and $\epsilon$ was chosen
\emph{a priori} to decay like
$1/\sqrt{d}$, \ie $\epsilon = \min(2/\sqrt{d}, 1)$. Here, we bracket
this decay rate with a quicker decay (more exploitation) and a slower decay
(more exploration). Specifically, we evaluate AEGiS on the $15$ synthetic test
functions using $\epsilon = \min(2 / (d - 2), 1)$ and
$\epsilon = \min(2 / \log(d + 3), 1)$ labelled \emph{slower} and
\emph{faster} respectively. These decay functions were chosen because they
match AEGiS and provide no exploitation when $d \leq 4$, \ie $\epsilon = 0.5$.
See the supplementary material for a visual comparison of the rates.
Figure~\ref{fig:bracketing} summarises the results on the $15$ test functions.
When using a smaller number of workers ($q \in \{4, 8\}$) the $2/\sqrt{d}$
decay gives superior performance. However, when  $q = 16$ there is little
to differentiate between the three rates.  We note that when $d$ is large the
faster rate of the decay (more exploitation for a given $d$) is superior; see
the supplementary material for full details.

\paragraph{\textbf{Proportion of TS to PF selection}}
\begin{figure}[t] %
\includegraphics[width=\linewidth, clip, trim={7 7 7 7}]{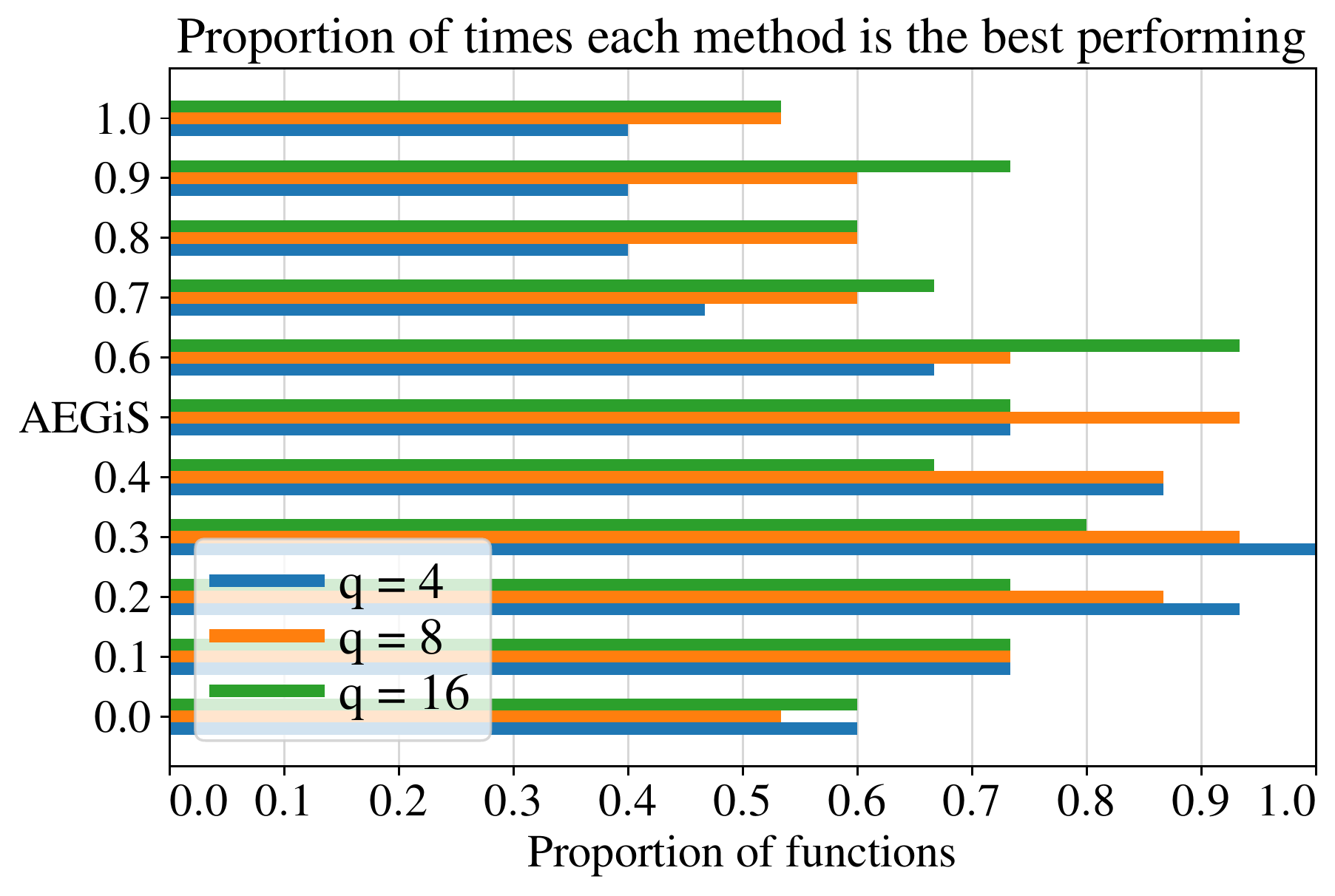}%
\caption{Proportion of TS to Pareto set selection for proportions
$\gamma \in \{0, 0.1, \dots, 1\}$ (vertical axis). Note that the label
\emph{AEGiS} corresponds to $\gamma = 0.5$.}
\label{fig:TStoPF}
\end{figure}
In the above evaluations TS and PF were selected with equal probability
$\epsilon_T = \epsilon_P$.  Here, we
investigate the proportion of times that TS should be chosen over Pareto set
selection. Specifically we evaluate AEGiS on the synthetic benchmark functions
for $q \in \{4,8,16\}$ with the split between exploitation, TS and Pareto set
selection being $1 - \epsilon$, $\epsilon_T = \gamma \epsilon$ and
$\epsilon_P = (1 - \gamma) \epsilon$ respectively and 
$\gamma \in \{ 0, 0.1, \dots, 1 \}$.
Note that two of the methods evaluated in the earlier ablation
study, \emph{No TS} and \emph{No PF}, correspond to $\gamma=0$ and $\gamma=1$
respectively. Figure~\ref{fig:TStoPF} summarises the results on the 15
synthetic benchmark functions. It shows that using more TS than Pareto
set selection appears detrimental to AEGiS, whereas using less TS, \ie a
smaller value of $\gamma$ leads to slightly improved average performance over $0.5$.
Interestingly, while AEGiS generally performed better with a smaller value of
$\gamma$, for some test functions, \eg Rosenbrock, larger values were
preferred -- see the supplementary material for all convergence plots and
tabulated results.  In general we recommend the standard setting
$\epsilon_T = \epsilon_P = \epsilon/2$.

\paragraph{\textbf{Sequential BO}}
\begin{figure}[t]
\centering
\includegraphics[width=1\linewidth, clip, trim={7 7 7 7}]{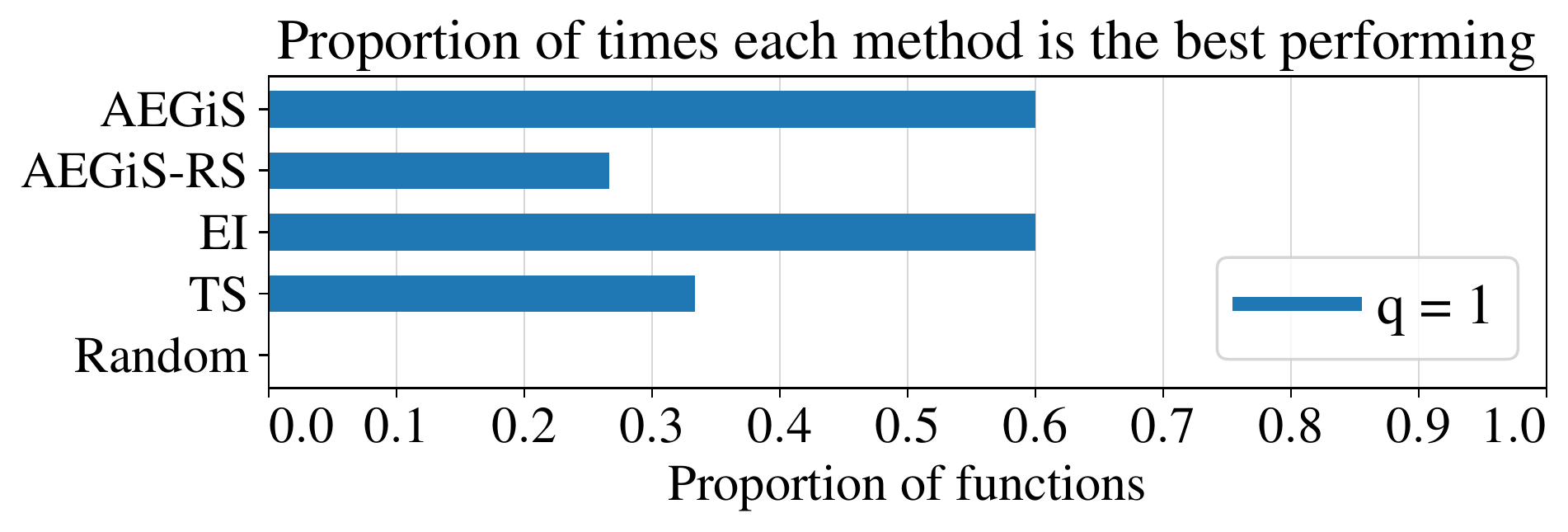}%
\caption{Synthetic function optimisation summary for the sequential setting
($q=1$). Bar lengths correspond to the proportion of times that a method is
best or statistically equivalent to the best method across the 15 synthetic
functions.}
\label{fig:sequential:equaltobest}
\end{figure}
Finally, we evaluate AEGiS in the sequential setting, \ie with $q=1$. In this
setting KB, LP, and PLAyBOOK are equivalent because they all select the next
(and only) location to be evaluated using  EI. Therefore, we compare AEGiS and
AEGiS-RS to TS, EI, and Latin hypercube sampling. As shown in 
Figure~\ref{fig:sequential:equaltobest}, AEGiS and EI had approximately 
equivalent performance, and they were both superior to AEGiS-RS and TS. This
again emphasises the importance of sampling from $\Papprox$. An ablation study
was also carried out with similar findings: removing the sampling from
$\Papprox$ led to a large reduction in performance. This was a larger drop in
performance in comparison to the ablation study using larger values of $q$. We
suspect that this is because the model will naturally be of a higher quality at
any iteration for $q=1$ and thus selection from $\Papprox$ will be even more
informative to the optimisation process. Different rates of $\epsilon$ decay
were also investigated as before, and similar results were found: both
decreasing and increasing the rate of decay led to worse performance, with the
largest decrease coming from increasing the rate and thus increasing
exploitation further.

\section{Conclusion}
\label{sec:conclusion}
Practical optimisation of expensive functions can often make use of
parallel hardware to rapidly obtain results. The AEGiS method makes best
use of hardware resources through asynchronous Bayesian optimisation by
combining greedy exploitation of the surrogate mean prediction with 
Thompson sampling and exploratory moves from the approximate Pareto set
that maximally trades off exploration and exploitation. The ablation study
verifies the importance of each of these  components. With only a single
worker this simple algorithm is no worse than BO using expected
improvement. As the problem dimension increases deliberate exploratory
moves are less necessary because of the inadvertent exploration due to
modelling inaccuracies. We showed empirically on a wide range of problems
that setting $\epsilon \propto 2/\sqrt{d}$ is more efficient than faster or
slower rates of reducing the amount of exploration carried out as the problem
dimensionality $d$ increases.

Unlike other methods, such as PLAyBOOK and Thompson sampling, AEGiS cannot be
trivially transformed into a synchronous batch BO method, and therefore it
cannot be directly compared to a synchronous version of itself. The closest
method in spirit is the $\epsilon$-shotgun method of
\citet{death:eshotgun:2020}. It scatters $q-1$ samples around a central
location, where the samples are distributed according the properties of the
surrogate model's landscape. The central location is chosen to be either the
most exploitative location or a randomly selected location from the Pareto set.
Interestingly the authors of both PLAyBOOK and Thompson sampling empirically
found that, even though there is less information available for the selection
of each asynchronous location, their methods outperformed their synchronous
counterparts. We note that AEGiS is more effective than both PLAyBOOK and
Thompson sampling and, therefore, also their synchronous equivalents.

Although AEGiS is robust to both the amount of exploration
($\epsilon_T + \epsilon_P$) and the ratio $\epsilon_T : \epsilon_P$, the
combination of greedy exploitation, Thompson sampling and selection from the
exploration-exploitation Pareto set is a challenge to obtaining non-trivial
bounds on the convergence rate. While future work will concentrate on providing
such theoretical guarantees, our extensive empirical evaluation shows that
AEGiS is a simple, practical and robust method for asynchronous batch Bayesian
optimisation.

\begin{acknowledgements} %
This work was supported by Innovate UK, grant number 104400. The authors would
like to acknowledge the use of the University of Exeter High-Performance
Computing (HPC) facility.
\end{acknowledgements}

\bibliography{de-ath_227}

\end{document}


\title{Asynchronous \eps-Greedy Bayesian Optimisation (Supplementary material)}

\author[1]{George {De Ath}}
\author[1]{Richard M. Everson}
\author[1]{Jonathan E. Fieldsend}
\affil[1]{%
  Department of Computer Science\\
  University of Exeter\\
  Exeter, United Kingdom
}

\onecolumn
\maketitle
\appendix

\section{Introduction}
\label{sec:intro}
In this supplementary materials document we provide details of all additional
experiments carried out in the course of this work. In 
Sections~\ref{sec:synthres}, \ref{sec:instanceres} and \ref{sec:ablation} we
show the results of the synthetic problems, instance-based problems and
the ablation study respectively. Section~\ref{sec:settingeps} details the
\eps-setting experiments and Section~\ref{sec:sequential} contains the results
of the experiments, ablation study and \eps-setting experiments in the
sequential setting (\ie $q = 1$).

\section{Experimental details}
\label{sec:experimental-details}

Here we give additional details of the algorithms used.

\subsection{NSGA-II}
To find the approximate Pareto set we used NSGA-II \citep{deb:nsga2:2002}
with a population size of $100d$, a mutation rate of $d^{-1}$, a crossover
rate of $0.8$, and mutation and distribution indices of
$\eta_c = \eta_m = 20$.

\section{Additional Results: Synthetic Functions}
\label{sec:synthres}
\begin{figure}[t]
\centering
\includegraphics[width=0.5\linewidth, clip, trim={5 0 5 7}]{figs/Branin}%
\includegraphics[width=0.5\linewidth, clip, trim={5 0 5 7}]{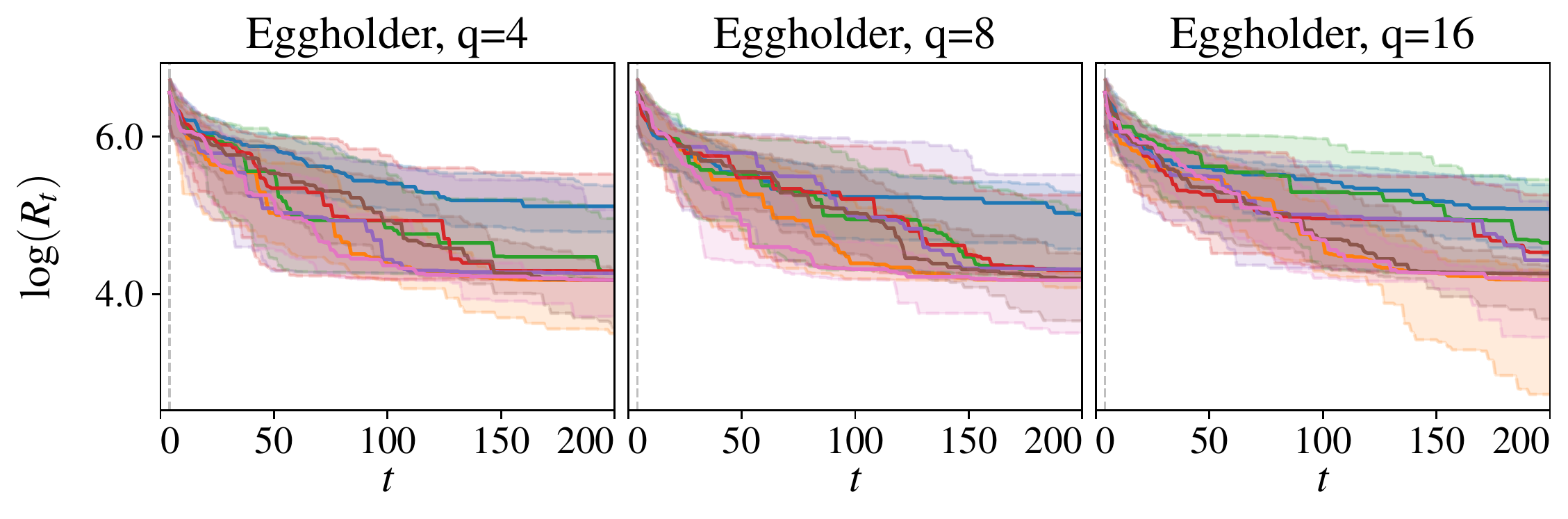}\\
\includegraphics[width=0.5\linewidth, clip, trim={5 0 5 7}]{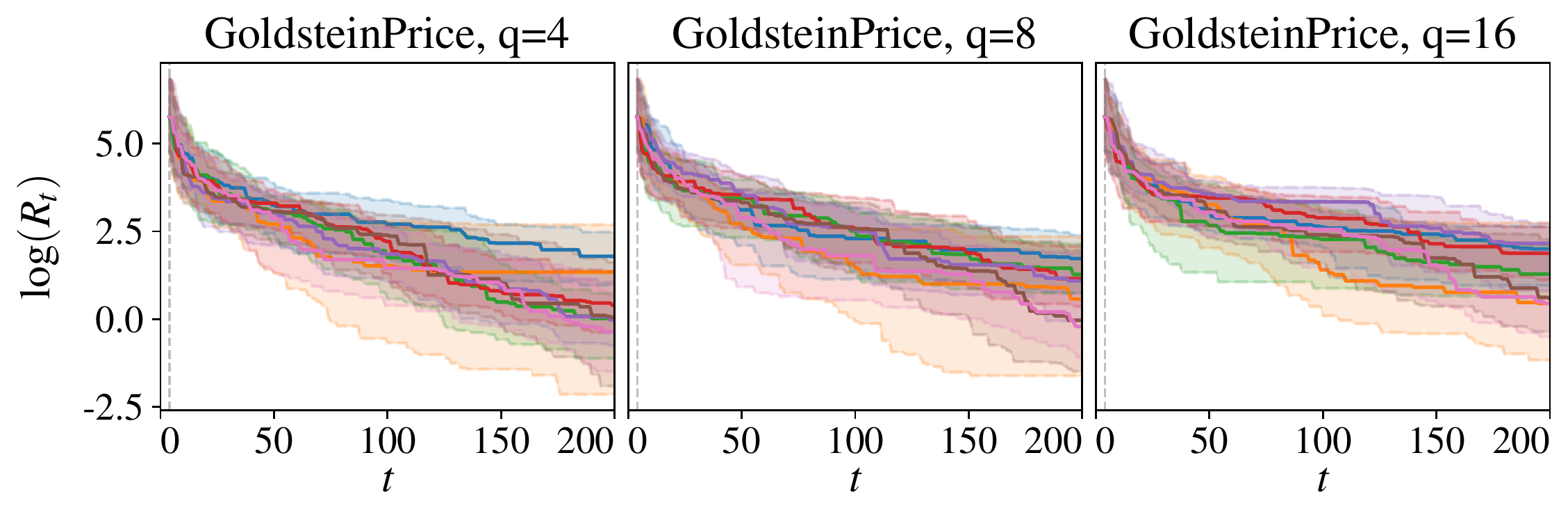}%
\includegraphics[width=0.5\linewidth, clip, trim={5 0 5 7}]{figs/SixHumpCamel}\\
\includegraphics[width=0.8\linewidth, clip, trim={10 15 10 13}]{figs/legend_twocol}%
\caption{Convergence results for the synthetic test problems.}
\label{fig:results:synthetic1}
\end{figure}

\begin{figure}[t] %
\centering
\includegraphics[width=0.5\linewidth, clip, trim={5 0 5 7}]{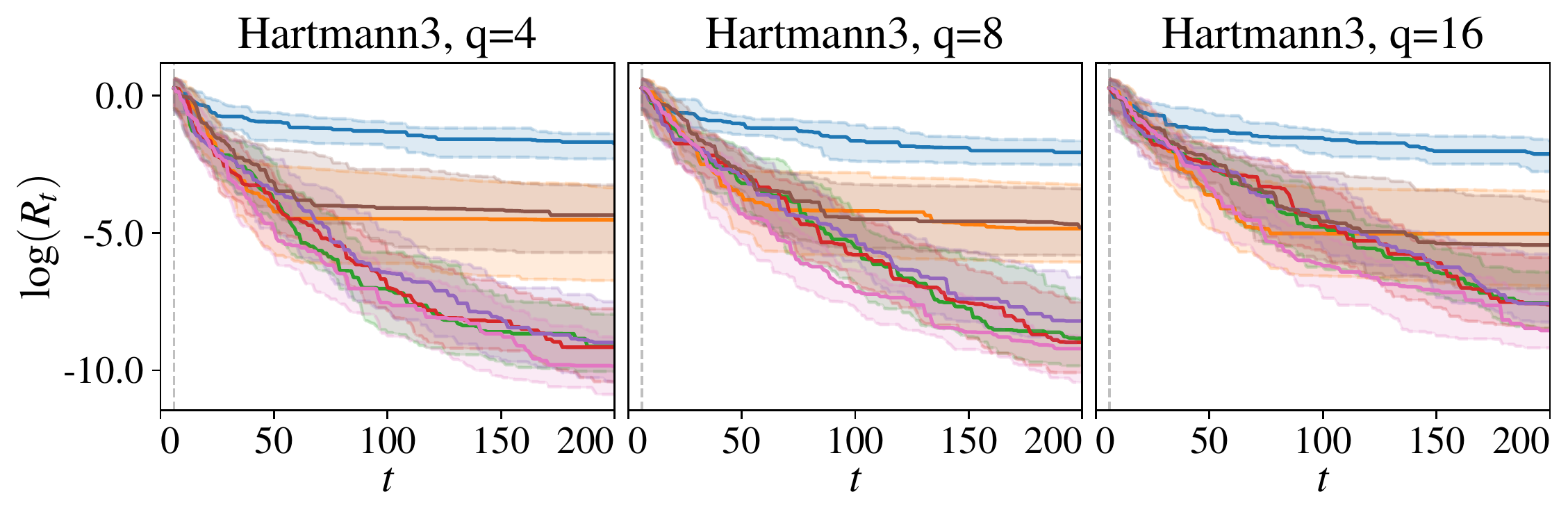}%
\includegraphics[width=0.5\linewidth, clip, trim={5 0 5 7}]{figs/Ackley5}\\
\includegraphics[width=0.5\linewidth, clip, trim={5 0 5 7}]{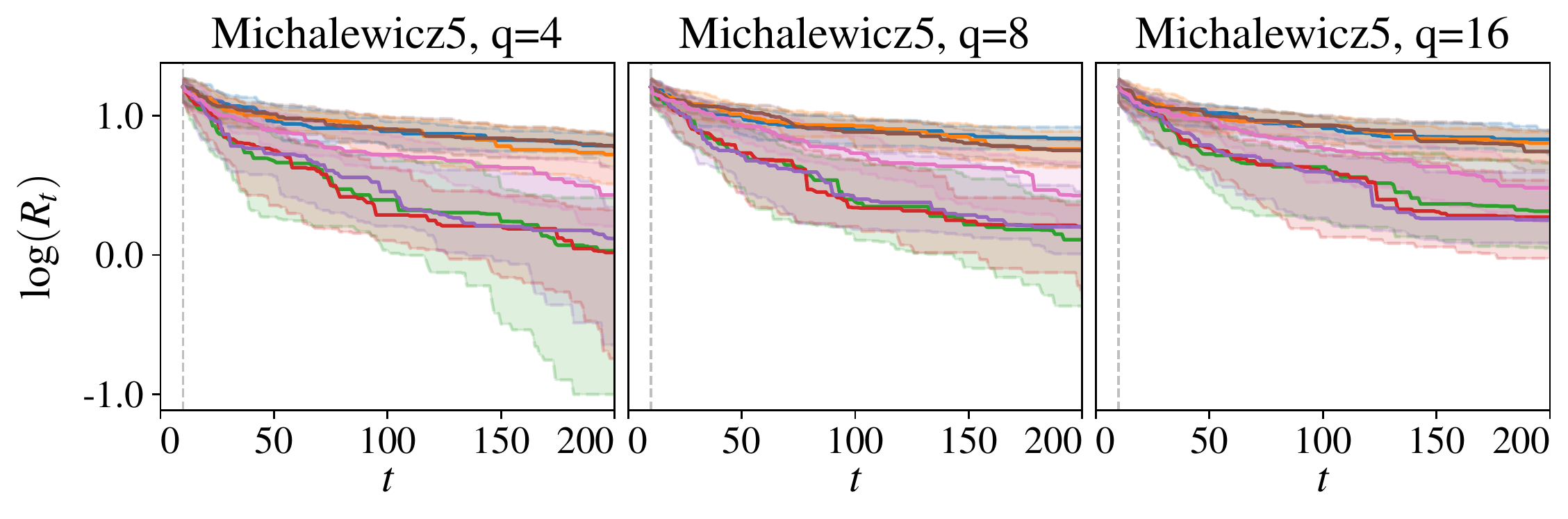}%
\includegraphics[width=0.5\linewidth, clip, trim={5 0 5 7}]{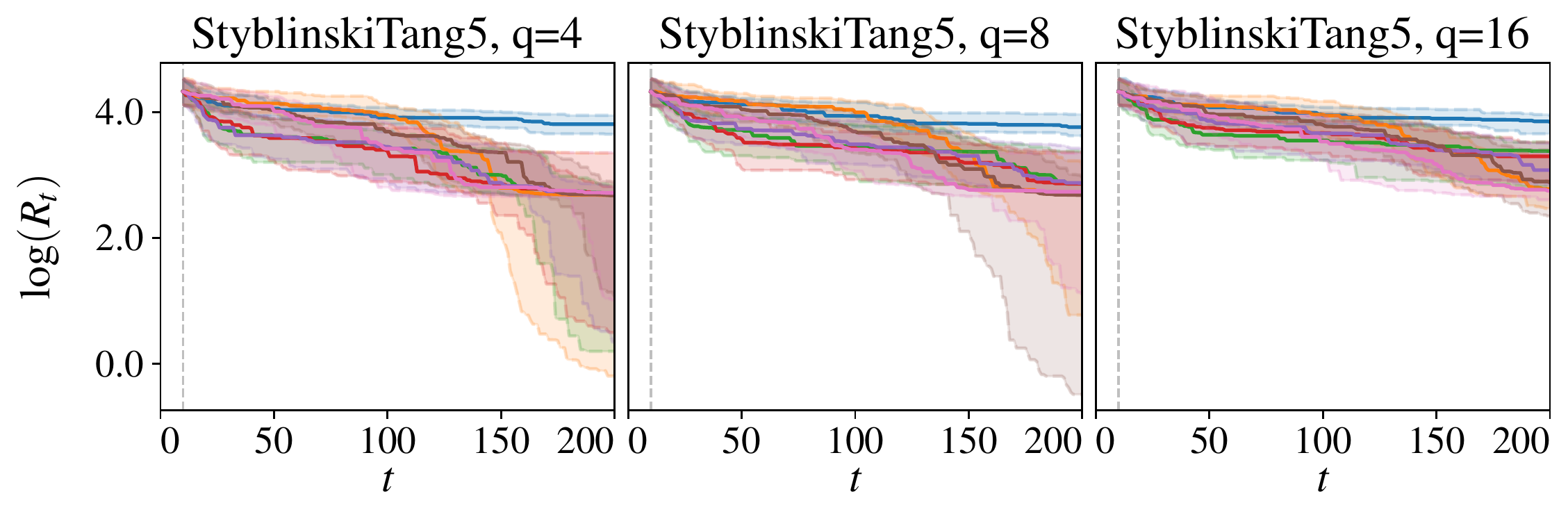}\\
\includegraphics[width=0.8\linewidth, clip, trim={10 15 10 13}]{figs/legend_twocol}%
\caption{Convergence results for the synthetic test problems.}
\label{fig:results:synthetic2}
\end{figure}

\begin{figure}[t] %
\centering
\includegraphics[width=0.5\linewidth, clip, trim={5 0 5 7}]{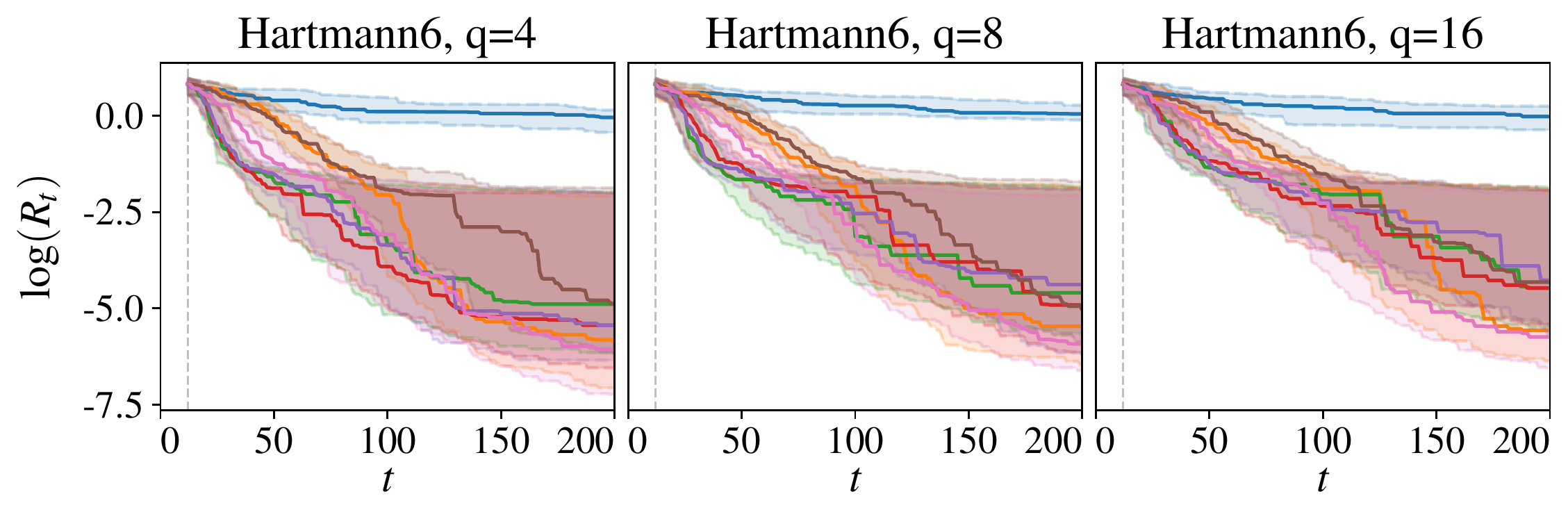}%
\includegraphics[width=0.5\linewidth, clip, trim={5 0 5 7}]{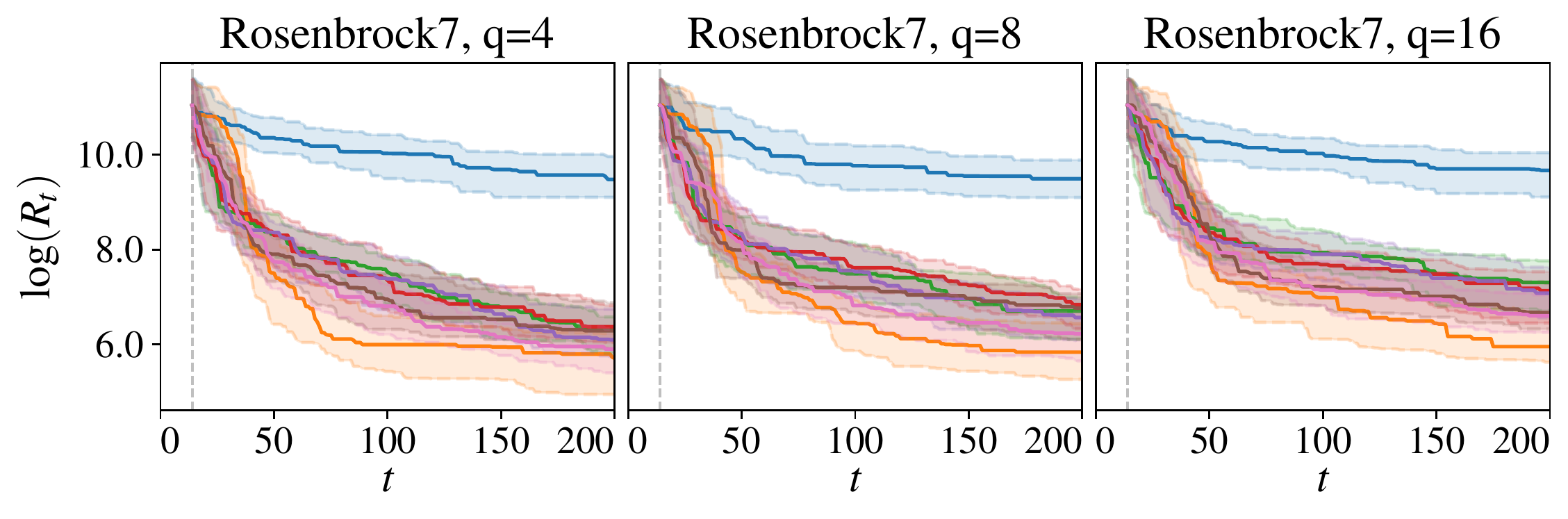}\\
\includegraphics[width=0.5\linewidth, clip, trim={5 0 5 7}]{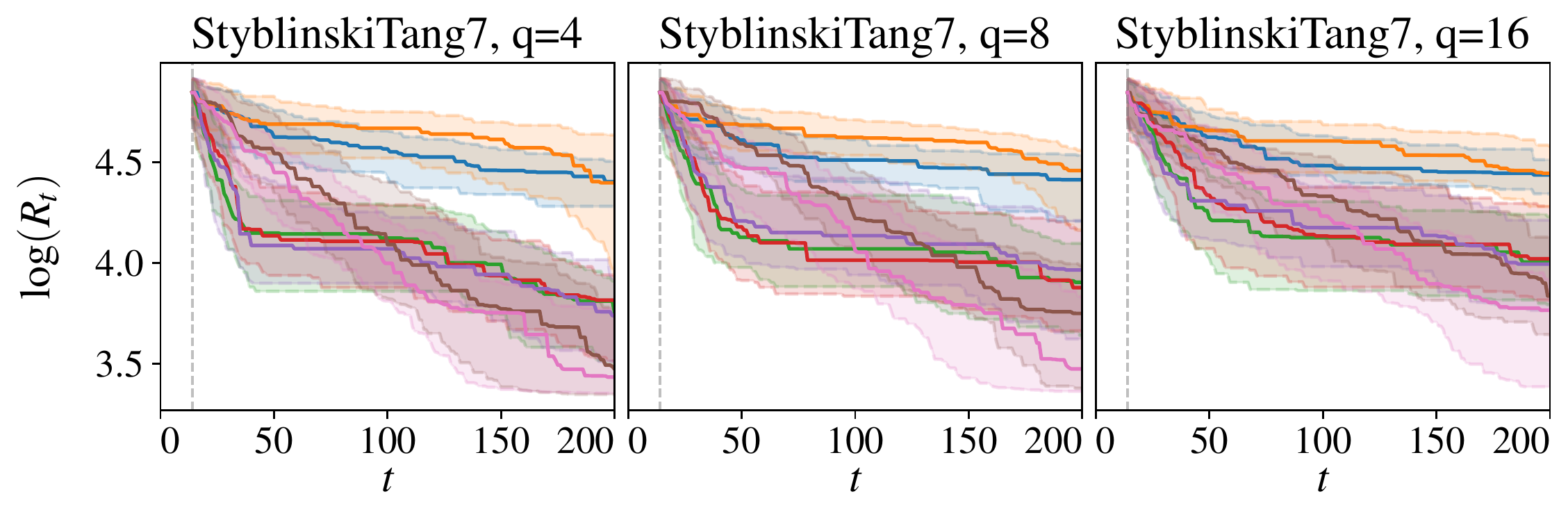}%
\includegraphics[width=0.5\linewidth, clip, trim={5 0 5 7}]{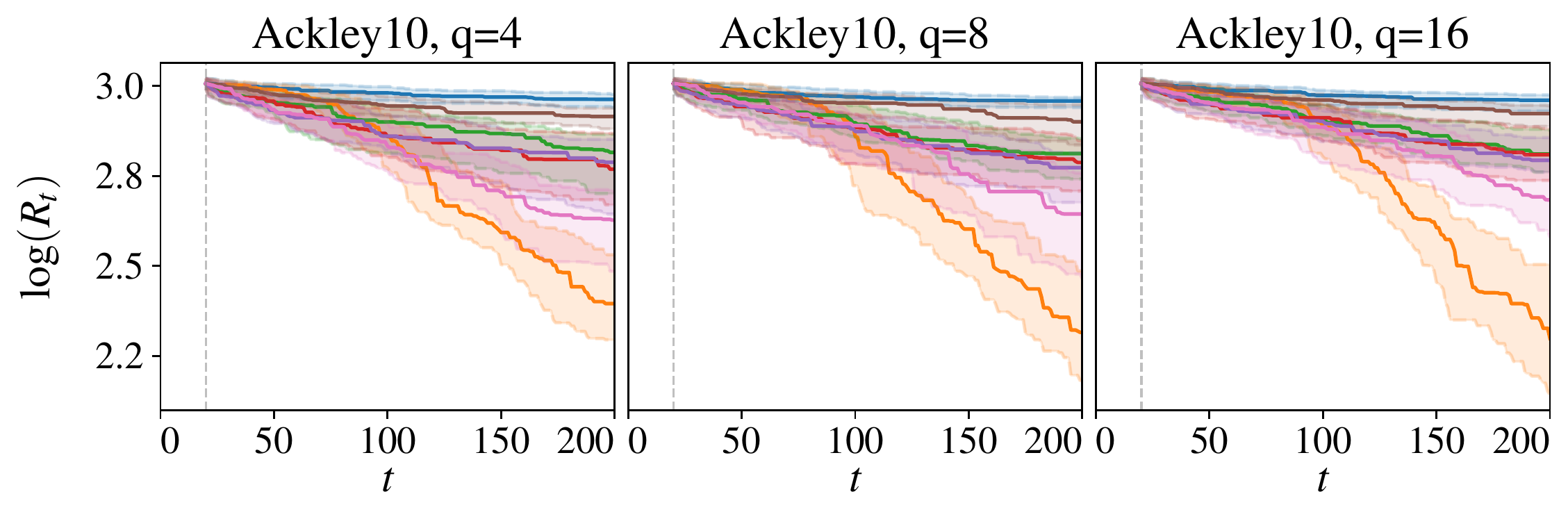}\\
\includegraphics[width=0.8\linewidth, clip, trim={10 15 10 13}]{figs/legend_twocol}%
\caption{Convergence results for the synthetic test problems.}
\label{fig:results:synthetic3}
\end{figure}

\begin{figure}[t] %
\centering
\includegraphics[width=0.5\linewidth, clip, trim={5 0 5 7}]{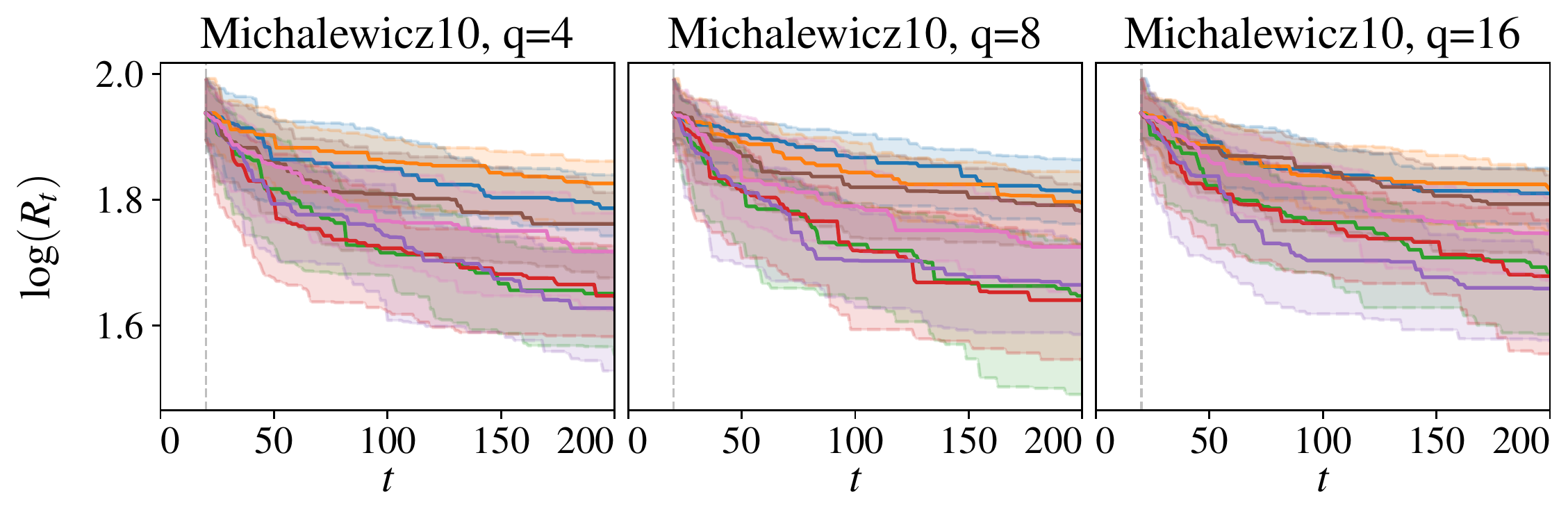}%
\includegraphics[width=0.5\linewidth, clip, trim={5 0 5 7}]{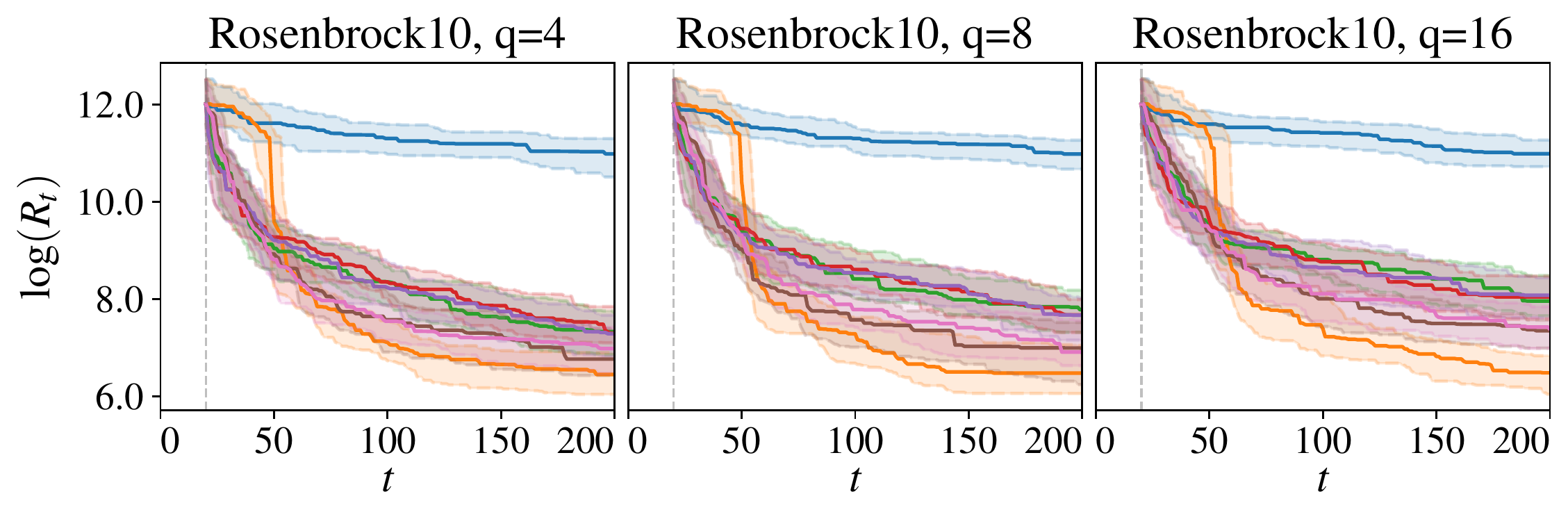}\\
\includegraphics[width=0.5\linewidth, clip, trim={5 0 5 7}]{figs/StyblinskiTang10}\\
\includegraphics[width=0.8\linewidth, clip, trim={10 15 10 13}]{figs/legend_twocol}%
\caption{Convergence results for the synthetic test problems.}
\label{fig:results:synthetic4}
\end{figure}

In this section we show all convergence plots and results tables for the 
synthetic function experiments. Figures~\ref{fig:results:synthetic1}, 
\ref{fig:results:synthetic2}, \ref{fig:results:synthetic3},
and~\ref{fig:results:synthetic4} show the convergence plots for 
each of the six methods on the fifteen benchmark problems for 
$q \in \{4,8,16\}$. Each plot shows the median log simple regret, with shading
representing the interquartile range over $51$ runs. 
Tables~\ref{tbl:synthetic_results_4}, \ref{tbl:synthetic_results_8} and 
\ref{tbl:synthetic_results_16} show the median log simple regret as well as 
the median absolute deviation from the median (MAD), a robust measure of
dispersion. The method with the best (lowest) median regret is shown in dark
grey, and those that are statistically equivalent to the best method according
to a one-sided, paired Wilcoxon signed-rank test with Holm-Bonferroni 
correction \citep{holm:test:1979} ($p \geq 0.05$) are shown in light grey.

\begin{table}[t]
\setlength{\tabcolsep}{2pt}
\sisetup{table-format=1.2e-1,table-number-alignment=center}
\caption{Tabulated results for $q = 4$ asynchronous workers, showing the
median log simple regret (\emph{left}) and median absolute deviation from
the median (MAD, \emph{right}) after 200 function evaluations across the 51 runs. 
The method with the lowest median performance is shown in dark grey, with 
those with statistically equivalent performance are shown in light grey.}
\resizebox{1\textwidth}{!}{%
\begin{tabular}{l Sz Sz Sz Sz Sz}
    \toprule
    \bfseries Method
    & \multicolumn{2}{c}{\bfseries Branin (2)} 
    & \multicolumn{2}{c}{\bfseries Eggholder (2)} 
    & \multicolumn{2}{c}{\bfseries GoldsteinPrice (2)} 
    & \multicolumn{2}{c}{\bfseries SixHumpCamel (2)} 
    & \multicolumn{2}{c}{\bfseries Hartmann3 (3)} \\ 
    & \multicolumn{1}{c}{Median} & \multicolumn{1}{c}{MAD}
    & \multicolumn{1}{c}{Median} & \multicolumn{1}{c}{MAD}
    & \multicolumn{1}{c}{Median} & \multicolumn{1}{c}{MAD}
    & \multicolumn{1}{c}{Median} & \multicolumn{1}{c}{MAD}
    & \multicolumn{1}{c}{Median} & \multicolumn{1}{c}{MAD}  \\ \midrule
    Random & 1.73e-01 & 2.02e-01 & 1.66e+02 & 7.34e+01 & 5.99e+00 & 6.22e+00 & 7.35e-02 & 6.39e-02 & 1.71e-01 & 1.07e-01 \\
    TS & 4.39e-03 & 5.65e-03 & \best 6.51e+01 & \best 2.06e+01 & 3.81e+00 & 5.60e+00 & 2.60e-04 & 3.84e-04 & 1.08e-02 & 1.57e-02 \\
    KB & 8.14e-05 & 1.16e-04 & 7.20e+01 & 9.85e+01 & \statsimilar 1.01e+00 & \statsimilar 1.13e+00 & 1.48e-04 & 1.43e-04 & \statsimilar 1.09e-04 & \statsimilar 1.29e-04 \\
    LP & 1.24e-04 & 1.54e-04 & 7.32e+01 & 9.70e+01 & 1.49e+00 & 1.42e+00 & 2.05e-04 & 2.62e-04 & \statsimilar 1.06e-04 & \statsimilar 1.42e-04 \\
    PLAyBOOK & 1.58e-04 & 1.96e-04 & 7.11e+01 & 1.31e+01 & \statsimilar 9.61e-01 & \statsimilar 1.04e+00 & 2.92e-04 & 2.97e-04 & \statsimilar 1.25e-04 & \statsimilar 1.50e-04 \\
    AEGiS-RS & 1.39e-04 & 1.96e-04 & \statsimilar 6.51e+01 & \statsimilar 2.78e+01 & \statsimilar 1.05e+00 & \statsimilar 1.48e+00 & \best 2.39e-06 & \best 2.79e-06 & 1.28e-02 & 1.61e-02 \\
    AEGiS & \best 5.99e-06 & \best 6.90e-06 & \statsimilar 6.52e+01 & \statsimilar 1.08e+01 & \best 6.99e-01 & \best 7.67e-01 & \statsimilar 2.93e-06 & \statsimilar 3.31e-06 & \best 5.29e-05 & \best 5.59e-05 \\
\bottomrule
\toprule
    \bfseries Method
    & \multicolumn{2}{c}{\bfseries Ackley5 (5)} 
    & \multicolumn{2}{c}{\bfseries Michalewicz5 (5)} 
    & \multicolumn{2}{c}{\bfseries StyblinskiTang5 (5)} 
    & \multicolumn{2}{c}{\bfseries Hartmann6 (6)} 
    & \multicolumn{2}{c}{\bfseries Rosenbrock7 (7)} \\ 
    & \multicolumn{1}{c}{Median} & \multicolumn{1}{c}{MAD}
    & \multicolumn{1}{c}{Median} & \multicolumn{1}{c}{MAD}
    & \multicolumn{1}{c}{Median} & \multicolumn{1}{c}{MAD}
    & \multicolumn{1}{c}{Median} & \multicolumn{1}{c}{MAD}
    & \multicolumn{1}{c}{Median} & \multicolumn{1}{c}{MAD}  \\ \midrule
    Random & 1.62e+01 & 1.71e+00 & 2.19e+00 & 2.81e-01 & 4.50e+01 & 1.00e+01 & 9.57e-01 & 3.60e-01 & 1.31e+04 & 9.64e+03 \\
    TS & 3.70e+00 & 8.13e-01 & 2.06e+00 & 5.05e-01 & \statsimilar 1.45e+01 & \statsimilar 2.05e+01 & \statsimilar 2.78e-03 & \statsimilar 3.25e-03 & \best 3.00e+02 & \best 2.54e+02 \\
    KB & 1.21e+01 & 5.72e+00 & \statsimilar 1.03e+00 & \statsimilar 8.01e-01 & \statsimilar 1.46e+01 & \statsimilar 4.55e+00 & 7.51e-03 & 1.04e-02 & 5.42e+02 & 3.00e+02 \\
    LP & 1.41e+01 & 4.13e+00 & \best 1.02e+00 & \best 5.61e-01 & \statsimilar 1.47e+01 & \statsimilar 5.64e+00 & 4.34e-03 & 5.40e-03 & 5.86e+02 & 3.31e+02 \\
    PLAyBOOK & 1.34e+01 & 4.81e+00 & \statsimilar 1.13e+00 & \statsimilar 6.83e-01 & \statsimilar 1.46e+01 & \statsimilar 1.24e+01 & 4.32e-03 & 5.37e-03 & 4.45e+02 & 3.00e+02 \\
    AEGiS-RS & 1.47e+01 & 3.95e+00 & 2.17e+00 & 3.35e-01 & \best 1.45e+01 & \best 1.21e+01 & 7.62e-03 & 1.03e-02 & 5.37e+02 & 3.14e+02 \\
    AEGiS & \best 2.81e+00 & \best 1.19e+00 & 1.54e+00 & 4.77e-01 & \statsimilar 1.51e+01 & \statsimilar 2.00e+01 & \best 2.28e-03 & \best 3.09e-03 & \statsimilar 3.64e+02 & \statsimilar 2.44e+02 \\
\bottomrule
\toprule
    \bfseries Method
    & \multicolumn{2}{c}{\bfseries StyblinskiTang7 (7)} 
    & \multicolumn{2}{c}{\bfseries Ackley10 (10)} 
    & \multicolumn{2}{c}{\bfseries Michalewicz10 (10)} 
    & \multicolumn{2}{c}{\bfseries Rosenbrock10 (10)} 
    & \multicolumn{2}{c}{\bfseries StyblinskiTang10 (10)} \\ 
    & \multicolumn{1}{c}{Median} & \multicolumn{1}{c}{MAD}
    & \multicolumn{1}{c}{Median} & \multicolumn{1}{c}{MAD}
    & \multicolumn{1}{c}{Median} & \multicolumn{1}{c}{MAD}
    & \multicolumn{1}{c}{Median} & \multicolumn{1}{c}{MAD}
    & \multicolumn{1}{c}{Median} & \multicolumn{1}{c}{MAD}  \\ \midrule
    Random & 8.18e+01 & 1.32e+01 & 1.93e+01 & 5.27e-01 & 5.97e+00 & 4.53e-01 & 5.91e+04 & 3.31e+04 & 1.44e+02 & 1.87e+01 \\
    TS & 8.14e+01 & 3.55e+01 & \best 1.10e+01 & \best 1.97e+00 & 6.21e+00 & 4.07e-01 & \best 6.34e+02 & \best 4.05e+02 & 1.65e+02 & 2.55e+01 \\
    KB & 4.27e+01 & 1.29e+01 & 1.67e+01 & 1.55e+00 & \statsimilar 5.18e+00 & \statsimilar 6.20e-01 & 1.53e+03 & 9.03e+02 & 8.93e+01 & 2.13e+01 \\
    LP & 4.54e+01 & 1.61e+01 & 1.59e+01 & 2.48e+00 & \statsimilar 5.20e+00 & \statsimilar 5.34e-01 & 1.45e+03 & 8.25e+02 & 8.62e+01 & 2.74e+01 \\
    PLAyBOOK & 4.20e+01 & 1.29e+01 & 1.62e+01 & 2.26e+00 & \best 5.08e+00 & \best 7.27e-01 & 1.46e+03 & 9.33e+02 & 8.58e+01 & 2.71e+01 \\
    AEGiS-RS & \statsimilar 3.22e+01 & \statsimilar 1.60e+01 & 1.84e+01 & 7.77e-01 & 5.82e+00 & 6.00e-01 & 8.69e+02 & 5.11e+02 & \statsimilar 6.85e+01 & \statsimilar 2.54e+01 \\
    AEGiS & \best 3.10e+01 & \best 1.79e+01 & 1.38e+01 & 2.30e+00 & 5.57e+00 & 7.16e-01 & 1.08e+03 & 7.27e+02 & \best 5.96e+01 & \best 2.41e+01 \\
\bottomrule
\end{tabular}
}
\label{tbl:synthetic_results_4}
\end{table}

\clearpage

\begin{table}[t]
\setlength{\tabcolsep}{2pt}
\sisetup{table-format=1.2e-1,table-number-alignment=center}
\caption{Tabulated results for $q = 8$ asynchronous workers, showing the
median log simple regret (\emph{left}) and median absolute deviation from
the median (MAD, \emph{right}) after 200 function evaluations across the 51 runs. 
The method with the lowest median performance is shown in dark grey, with 
those with statistically equivalent performance are shown in light grey.}
\resizebox{1\textwidth}{!}{%
\begin{tabular}{l Sz Sz Sz Sz Sz}
    \toprule
    \bfseries Method
    & \multicolumn{2}{c}{\bfseries Branin (2)} 
    & \multicolumn{2}{c}{\bfseries Eggholder (2)} 
    & \multicolumn{2}{c}{\bfseries GoldsteinPrice (2)} 
    & \multicolumn{2}{c}{\bfseries SixHumpCamel (2)} 
    & \multicolumn{2}{c}{\bfseries Hartmann3 (3)} \\ 
    & \multicolumn{1}{c}{Median} & \multicolumn{1}{c}{MAD}
    & \multicolumn{1}{c}{Median} & \multicolumn{1}{c}{MAD}
    & \multicolumn{1}{c}{Median} & \multicolumn{1}{c}{MAD}
    & \multicolumn{1}{c}{Median} & \multicolumn{1}{c}{MAD}
    & \multicolumn{1}{c}{Median} & \multicolumn{1}{c}{MAD}  \\ \midrule
    Random & 1.81e-01 & 2.21e-01 & 1.50e+02 & 7.39e+01 & 5.61e+00 & 6.48e+00 & 1.09e-01 & 8.72e-02 & 1.26e-01 & 8.14e-02 \\
    TS & 2.91e-03 & 3.73e-03 & \statsimilar 6.51e+01 & \statsimilar 9.41e+00 & \statsimilar 1.76e+00 & \statsimilar 2.60e+00 & 6.73e-05 & 9.86e-05 & 7.89e-03 & 1.07e-02 \\
    KB & 8.55e-05 & 1.14e-04 & 7.39e+01 & 6.63e+01 & 3.57e+00 & 3.41e+00 & 3.75e-04 & 4.68e-04 & 1.46e-04 & 1.69e-04 \\
    LP & 1.66e-04 & 1.76e-04 & 7.37e+01 & 8.40e+01 & 3.18e+00 & 3.19e+00 & 5.13e-04 & 5.47e-04 & 1.26e-04 & 1.59e-04 \\
    PLAyBOOK & 1.10e-04 & 1.27e-04 & 7.48e+01 & 7.94e+01 & \statsimilar 2.97e+00 & \statsimilar 3.67e+00 & 3.66e-04 & 4.38e-04 & 2.73e-04 & 3.28e-04 \\
    AEGiS-RS & 1.09e-04 & 1.19e-04 & 6.64e+01 & 4.31e+01 & \statsimilar 9.66e-01 & \statsimilar 1.39e+00 & \best 2.92e-06 & \best 3.27e-06 & 8.17e-03 & 1.16e-02 \\
    AEGiS & \best 5.32e-06 & \best 6.51e-06 & \best 6.51e+01 & \best 1.35e+01 & \best 8.10e-01 & \best 8.82e-01 & \statsimilar 2.98e-06 & \statsimilar 3.83e-06 & \best 9.99e-05 & \best 1.03e-04 \\
\bottomrule
\toprule
    \bfseries Method
    & \multicolumn{2}{c}{\bfseries Ackley5 (5)} 
    & \multicolumn{2}{c}{\bfseries Michalewicz5 (5)} 
    & \multicolumn{2}{c}{\bfseries StyblinskiTang5 (5)} 
    & \multicolumn{2}{c}{\bfseries Hartmann6 (6)} 
    & \multicolumn{2}{c}{\bfseries Rosenbrock7 (7)} \\ 
    & \multicolumn{1}{c}{Median} & \multicolumn{1}{c}{MAD}
    & \multicolumn{1}{c}{Median} & \multicolumn{1}{c}{MAD}
    & \multicolumn{1}{c}{Median} & \multicolumn{1}{c}{MAD}
    & \multicolumn{1}{c}{Median} & \multicolumn{1}{c}{MAD}
    & \multicolumn{1}{c}{Median} & \multicolumn{1}{c}{MAD}  \\ \midrule
    Random & 1.59e+01 & 2.07e+00 & 2.30e+00 & 2.86e-01 & 4.29e+01 & 1.22e+01 & 1.05e+00 & 3.09e-01 & 1.33e+04 & 7.34e+03 \\
    TS & \statsimilar 3.53e+00 & \statsimilar 6.96e-01 & 2.14e+00 & 4.66e-01 & \statsimilar 1.48e+01 & \statsimilar 1.68e+01 & \statsimilar 4.17e-03 & \statsimilar 4.99e-03 & \best 3.42e+02 & \best 2.79e+02 \\
    KB & 1.31e+01 & 5.66e+00 & \best 1.12e+00 & \best 6.19e-01 & 1.74e+01 & 1.68e+01 & 1.01e-02 & 1.37e-02 & 8.14e+02 & 4.87e+02 \\
    LP & 1.37e+01 & 3.02e+00 & \statsimilar 1.23e+00 & \statsimilar 4.53e-01 & 1.74e+01 & 1.20e+01 & 6.42e-03 & 8.18e-03 & 9.32e+02 & 5.76e+02 \\
    PLAyBOOK & 1.38e+01 & 4.38e+00 & \statsimilar 1.22e+00 & \statsimilar 4.05e-01 & 1.77e+01 & 1.73e+01 & 1.25e-02 & 1.78e-02 & 7.10e+02 & 4.25e+02 \\
    AEGiS-RS & 1.44e+01 & 3.48e+00 & 2.11e+00 & 2.98e-01 & \best 1.46e+01 & \best 1.98e+01 & 7.11e-03 & 8.85e-03 & 8.65e+02 & 5.77e+02 \\
    AEGiS & \best 3.39e+00 & \best 8.01e-01 & 1.53e+00 & 5.62e-01 & \statsimilar 1.53e+01 & \statsimilar 1.89e+01 & \best 2.67e-03 & \best 3.16e-03 & \statsimilar 5.04e+02 & \statsimilar 3.54e+02 \\
\bottomrule
\toprule
    \bfseries Method
    & \multicolumn{2}{c}{\bfseries StyblinskiTang7 (7)} 
    & \multicolumn{2}{c}{\bfseries Ackley10 (10)} 
    & \multicolumn{2}{c}{\bfseries Michalewicz10 (10)} 
    & \multicolumn{2}{c}{\bfseries Rosenbrock10 (10)} 
    & \multicolumn{2}{c}{\bfseries StyblinskiTang10 (10)} \\ 
    & \multicolumn{1}{c}{Median} & \multicolumn{1}{c}{MAD}
    & \multicolumn{1}{c}{Median} & \multicolumn{1}{c}{MAD}
    & \multicolumn{1}{c}{Median} & \multicolumn{1}{c}{MAD}
    & \multicolumn{1}{c}{Median} & \multicolumn{1}{c}{MAD}
    & \multicolumn{1}{c}{Median} & \multicolumn{1}{c}{MAD}  \\ \midrule
    Random & 8.26e+01 & 1.67e+01 & 1.93e+01 & 3.95e-01 & 6.13e+00 & 4.68e-01 & 5.88e+04 & 2.73e+04 & 1.43e+02 & 1.82e+01 \\
    TS & 8.65e+01 & 2.52e+01 & \best 1.01e+01 & \best 2.46e+00 & 6.02e+00 & 4.71e-01 & \best 6.52e+02 & \best 4.64e+02 & 1.65e+02 & 2.38e+01 \\
    KB & 4.96e+01 & 1.70e+01 & 1.66e+01 & 1.28e+00 & \statsimilar 5.19e+00 & \statsimilar 8.12e-01 & 2.40e+03 & 1.65e+03 & 9.73e+01 & 2.26e+01 \\
    LP & 4.83e+01 & 1.88e+01 & 1.62e+01 & 1.75e+00 & \best 5.16e+00 & \best 7.35e-01 & 2.16e+03 & 1.14e+03 & 8.94e+01 & 2.89e+01 \\
    PLAyBOOK & 5.27e+01 & 2.19e+01 & 1.60e+01 & 1.66e+00 & \statsimilar 5.29e+00 & \statsimilar 5.77e-01 & 2.14e+03 & 1.30e+03 & 8.64e+01 & 1.95e+01 \\
    AEGiS-RS & \statsimilar 4.25e+01 & \statsimilar 1.91e+01 & 1.82e+01 & 1.38e+00 & 5.94e+00 & 4.83e-01 & 1.09e+03 & 8.64e+02 & \best 6.50e+01 & \best 2.25e+01 \\
    AEGiS & \best 3.23e+01 & \best 1.69e+01 & 1.41e+01 & 2.83e+00 & 5.61e+00 & 5.60e-01 & 9.97e+02 & 5.96e+02 & \statsimilar 6.75e+01 & \statsimilar 2.31e+01 \\
\bottomrule
\end{tabular}
}
\label{tbl:synthetic_results_8}
\end{table}

\clearpage

\begin{table}[t]
\setlength{\tabcolsep}{2pt}
\sisetup{table-format=1.2e-1,table-number-alignment=center}
\caption{Tabulated results for $q = 16$ asynchronous workers, showing the
median log simple regret (\emph{left}) and median absolute deviation from
the median (MAD, \emph{right}) after 200 function evaluations across the 51 runs. 
The method with the lowest median performance is shown in dark grey, with 
those with statistically equivalent performance are shown in light grey.}
\resizebox{1\textwidth}{!}{%
\begin{tabular}{l Sz Sz Sz Sz Sz}
    \toprule
    \bfseries Method
    & \multicolumn{2}{c}{\bfseries Branin (2)} 
    & \multicolumn{2}{c}{\bfseries Eggholder (2)} 
    & \multicolumn{2}{c}{\bfseries GoldsteinPrice (2)} 
    & \multicolumn{2}{c}{\bfseries SixHumpCamel (2)} 
    & \multicolumn{2}{c}{\bfseries Hartmann3 (3)} \\ 
    & \multicolumn{1}{c}{Median} & \multicolumn{1}{c}{MAD}
    & \multicolumn{1}{c}{Median} & \multicolumn{1}{c}{MAD}
    & \multicolumn{1}{c}{Median} & \multicolumn{1}{c}{MAD}
    & \multicolumn{1}{c}{Median} & \multicolumn{1}{c}{MAD}
    & \multicolumn{1}{c}{Median} & \multicolumn{1}{c}{MAD}  \\ \midrule
    Random & 1.45e-01 & 1.47e-01 & 1.61e+02 & 1.05e+02 & 7.34e+00 & 8.20e+00 & 7.09e-02 & 7.83e-02 & 1.19e-01 & 1.01e-01 \\
    TS & 1.47e-03 & 2.11e-03 & \best 6.51e+01 & \best 4.59e+01 & \statsimilar 1.56e+00 & \statsimilar 2.29e+00 & 2.86e-05 & 3.62e-05 & 6.53e-03 & 9.08e-03 \\
    KB & 3.53e-04 & 3.72e-04 & 1.04e+02 & 8.39e+01 & \statsimilar 3.62e+00 & \statsimilar 3.45e+00 & 2.14e-03 & 3.07e-03 & 5.25e-04 & 6.34e-04 \\
    LP & 3.99e-04 & 4.98e-04 & 9.27e+01 & 8.75e+01 & 6.52e+00 & 6.35e+00 & 4.97e-03 & 6.41e-03 & 4.96e-04 & 6.01e-04 \\
    PLAyBOOK & 3.18e-04 & 4.11e-04 & 8.36e+01 & 8.29e+01 & 8.66e+00 & 9.65e+00 & 4.02e-03 & 5.46e-03 & 5.15e-04 & 6.17e-04 \\
    AEGiS-RS & 5.47e-05 & 7.30e-05 & \statsimilar 7.07e+01 & \statsimilar 2.74e+01 & \statsimilar 1.77e+00 & \statsimilar 2.32e+00 & \best 3.85e-06 & \best 4.28e-06 & 4.33e-03 & 6.27e-03 \\
    AEGiS & \best 1.28e-05 & \best 1.80e-05 & \statsimilar 6.53e+01 & \statsimilar 1.27e+01 & \best 1.56e+00 & \best 1.84e+00 & \statsimilar 5.82e-06 & \statsimilar 5.32e-06 & \best 1.93e-04 & \best 2.80e-04 \\
\bottomrule
\toprule
    \bfseries Method
    & \multicolumn{2}{c}{\bfseries Ackley5 (5)} 
    & \multicolumn{2}{c}{\bfseries Michalewicz5 (5)} 
    & \multicolumn{2}{c}{\bfseries StyblinskiTang5 (5)} 
    & \multicolumn{2}{c}{\bfseries Hartmann6 (6)} 
    & \multicolumn{2}{c}{\bfseries Rosenbrock7 (7)} \\ 
    & \multicolumn{1}{c}{Median} & \multicolumn{1}{c}{MAD}
    & \multicolumn{1}{c}{Median} & \multicolumn{1}{c}{MAD}
    & \multicolumn{1}{c}{Median} & \multicolumn{1}{c}{MAD}
    & \multicolumn{1}{c}{Median} & \multicolumn{1}{c}{MAD}
    & \multicolumn{1}{c}{Median} & \multicolumn{1}{c}{MAD}  \\ \midrule
    Random & 1.66e+01 & 1.22e+00 & 2.30e+00 & 3.03e-01 & 4.67e+01 & 9.10e+00 & 9.78e-01 & 4.73e-01 & 1.58e+04 & 1.03e+04 \\
    TS & \statsimilar 3.75e+00 & \statsimilar 7.74e-01 & 2.23e+00 & 3.20e-01 & \statsimilar 1.60e+01 & \statsimilar 1.29e+01 & \statsimilar 3.79e-03 & \statsimilar 4.91e-03 & \best 3.84e+02 & \best 2.63e+02 \\
    KB & 1.49e+01 & 3.57e+00 & \statsimilar 1.37e+00 & \statsimilar 6.89e-01 & 2.93e+01 & 1.65e+01 & 1.14e-02 & 1.53e-02 & 1.49e+03 & 1.18e+03 \\
    LP & 1.49e+01 & 4.20e+00 & \statsimilar 1.31e+00 & \statsimilar 5.50e-01 & 2.69e+01 & 1.25e+01 & 1.14e-02 & 1.48e-02 & 1.25e+03 & 9.30e+02 \\
    PLAyBOOK & 1.48e+01 & 3.31e+00 & \best 1.29e+00 & \best 5.21e-01 & 2.17e+01 & 1.43e+01 & 1.40e-02 & 1.93e-02 & 1.18e+03 & 8.95e+02 \\
    AEGiS-RS & 1.13e+01 & 6.40e+00 & 2.10e+00 & 4.70e-01 & \statsimilar 1.78e+01 & \statsimilar 1.42e+01 & 1.33e-02 & 1.70e-02 & 7.87e+02 & 5.00e+02 \\
    AEGiS & \best 3.39e+00 & \best 1.12e+00 & 1.62e+00 & 5.10e-01 & \best 1.56e+01 & \best 1.09e+01 & \best 3.21e-03 & \best 3.82e-03 & 7.38e+02 & 5.13e+02 \\
\bottomrule
\toprule
    \bfseries Method
    & \multicolumn{2}{c}{\bfseries StyblinskiTang7 (7)} 
    & \multicolumn{2}{c}{\bfseries Ackley10 (10)} 
    & \multicolumn{2}{c}{\bfseries Michalewicz10 (10)} 
    & \multicolumn{2}{c}{\bfseries Rosenbrock10 (10)} 
    & \multicolumn{2}{c}{\bfseries StyblinskiTang10 (10)} \\ 
    & \multicolumn{1}{c}{Median} & \multicolumn{1}{c}{MAD}
    & \multicolumn{1}{c}{Median} & \multicolumn{1}{c}{MAD}
    & \multicolumn{1}{c}{Median} & \multicolumn{1}{c}{MAD}
    & \multicolumn{1}{c}{Median} & \multicolumn{1}{c}{MAD}
    & \multicolumn{1}{c}{Median} & \multicolumn{1}{c}{MAD}  \\ \midrule
    Random & 8.46e+01 & 1.09e+01 & 1.93e+01 & 5.99e-01 & 6.11e+00 & 3.89e-01 & 5.93e+04 & 2.66e+04 & 1.51e+02 & 1.79e+01 \\
    TS & 8.53e+01 & 2.04e+01 & \best 9.96e+00 & \best 2.36e+00 & 6.15e+00 & 4.44e-01 & \best 6.55e+02 & \best 3.56e+02 & 1.61e+02 & 2.42e+01 \\
    KB & 5.48e+01 & 1.99e+01 & 1.66e+01 & 1.54e+00 & \statsimilar 5.39e+00 & \statsimilar 5.95e-01 & 2.87e+03 & 1.44e+03 & 8.68e+01 & 2.69e+01 \\
    LP & 5.58e+01 & 1.96e+01 & 1.66e+01 & 1.90e+00 & \statsimilar 5.36e+00 & \statsimilar 8.89e-01 & 3.13e+03 & 1.86e+03 & 9.27e+01 & 2.58e+01 \\
    PLAyBOOK & 5.43e+01 & 1.69e+01 & 1.63e+01 & 8.28e-01 & \best 5.26e+00 & \best 5.96e-01 & 3.23e+03 & 2.04e+03 & 9.86e+01 & 3.53e+01 \\
    AEGiS-RS & 4.63e+01 & 1.70e+01 & 1.86e+01 & 6.95e-01 & 6.01e+00 & 5.74e-01 & 1.56e+03 & 6.77e+02 & \statsimilar 7.84e+01 & \statsimilar 3.30e+01 \\
    AEGiS & \best 4.32e+01 & \best 1.80e+01 & 1.46e+01 & 2.22e+00 & 5.74e+00 & 5.74e-01 & 1.60e+03 & 1.01e+03 & \best 7.20e+01 & \best 2.83e+01 \\
\bottomrule
\end{tabular}
}
\label{tbl:synthetic_results_16}
\end{table}

\clearpage

\clearpage

\section{Instance-based Problems}
\label{sec:instanceres}
In this section we give further details of the hyperparameter optimisation
benchmark problems and the robot pushing problems, as well as providing the
average rank score plots for each problem.

\subsection{Hyperparameter Optimisation Benchmark}
\begin{figure}[t]
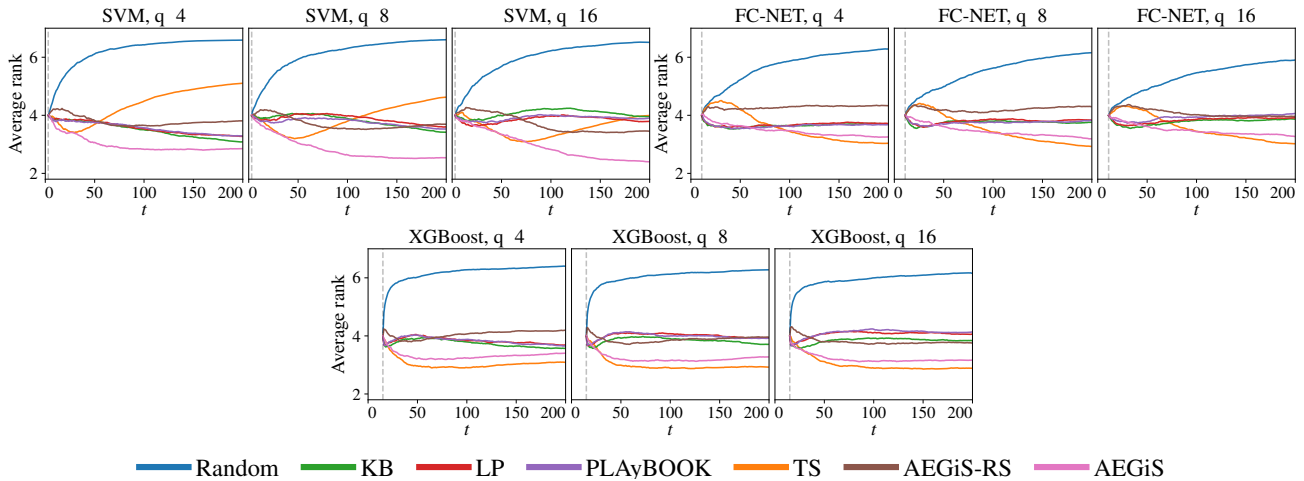

\centering
\includegraphics[width=0.5\linewidth, clip, trim={5 0 5 7}]{figs/svm}%
\includegraphics[width=0.5\linewidth, clip, trim={5 0 5 7}]{figs/fc-net}\\
\includegraphics[width=0.5\linewidth, clip, trim={5 0 5 7}]{figs/xgboost}\\
\includegraphics[width=0.8\linewidth, clip, trim={10 15 10 13}]{figs/legend_twocol}%
\caption{Average rank scores for the SVM, FC-NET and XGBoost 
hyperparameter optimisation problems.}
\label{fig:results:svm_fcnet_xgboost}
\end{figure}

The SVM problem was built using a set of SVM
classification models trained on 16 OpenML tasks, with $2$ input parameters
corresponding to the SVM's hyperparameters. The FC-NET problem was built using
a set of feed-forward neural networks trained on the same OpenML tasks, with 
$6$ input parameters corresponding to the network hyperparameters. XGBoost was
built using a set of XGBoost regression models trained on $11$ UCI datasets and
has $8$ input parameters, corresponding to the XGBoost hyperparameters. 
Table~\ref{tbl:profetparams} shows the hyperparameters optimised in the three
Profet benchmark problems \citep{klein:profet:2019}, their search spaces, and 
whether or not they were log scaled. Like the other optimisation runs performed
in this work, the inputs are rescaled to reside in $[0, 1]$. 
Figure~\ref{fig:results:svm_fcnet_xgboost} shows the average rank score plots
for the three functions.

\begin{table}[t]
\centering%
\caption{Parameters tuned in the Profet benchmark.}
\label{tbl:profetparams}
\begin{tabular}{l l c c}
\toprule
    \bfseries Benchmark & Parameter Name & Range & Log scaled \\
\midrule
    SVM & $C$      & $[e^{-10}, \, e^{10}]$ & Yes \\
        & $\gamma$ & $[e^{-10}, \, e^{10}]$ & Yes \\
\midrule
    FC-NET & Learning rate  & $[10^{-6}, \, 10^{-1}]$ & Yes \\
           & Batch size     & $[2^3, \, 2^7]$         & Yes \\
           & Layer 1: Units & $[2^4, \, 2^9]$         & Yes \\
           & Layer 2: Units & $[2^4, \, 2^9]$         & Yes \\
           & Layer 1: dropout rate & $[0, \, 0.99]$   & -   \\
           & Layer 2: dropout rate & $[0, \, 0.99]$   & -   \\
\midrule
    XGBoost & Learning rate        & $[10^{-6}, \, 10^{-1}]$ & Yes \\
            & $\gamma$             & $[0, 2]$                & -   \\
            & L1 regularisation    & $[10^-5, \, 10^3]$      & Yes \\
            & L2 regularisation    & $[10^-5, \, 10^3]$      & Yes \\
            & Number of estimators & $[10, 500]$             & -   \\
            & Subsampling          & $[0.1, 1]$              & -   \\
            & Maximum depth        & $[1, 15]$               & -   \\
            & Minimum child weight & $[0, 20]$               & -   \\
\bottomrule
\end{tabular}
\end{table}

\subsection{Robot Pushing Problems}
Here, we optimised the control parameters for two active learning robot pushing
problems. In the first problem (push4), a robot hand is given the task of 
pushing an object towards an unknown target location. Once the robot has
finished pushing the object, it receives feedback in the form of the
object-target distance. The adjustable parameters in this problem are the
robot's starting coordinates on the $2d$ plain, the orientation of its hand and
how long it pushes for. Therefore, this problem can be casts as a minimisation
in which we optimise the four parameters in order to minimise the object-target
distance. The object's initial location in push4 is always the centre of the
domain, and for each problem instance the target location is randomly
generated. Note that these instances are shared between methods.

The second problem (push8) is similar to the first except there are two robots
moving in the same arena, both having to push their own objects to their
respective targets. The final object-target distance from both robots are
summed to give an objective function to minimise, resulting in an
$8$-dimensional problem. Problem instances for push8 were generated such that
the minimum distance between the targets was sufficient that each of the
objects could be placed on the targets without overlapping. However, this does
not mean that each problem instance can be successfully optimised because the
targets may be positioned such that the robots block each others path 
\emph{en route} to the target location. 

\begin{figure}[t]
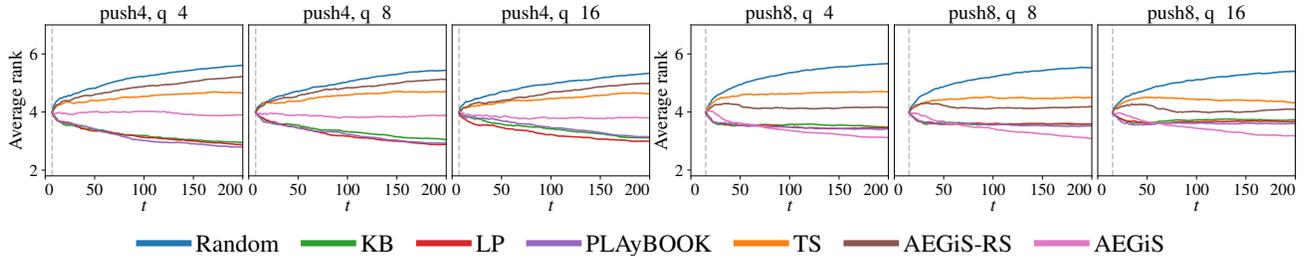

\centering
\includegraphics[width=0.5\linewidth, clip, trim={5 0 5 7}]{figs/push4}%
\includegraphics[width=0.5\linewidth, clip, trim={5 0 5 7}]{figs/push8}\\
\includegraphics[width=0.8\linewidth, clip, trim={10 15 10 13}]{figs/legend_twocol}%
\caption{Average rank scores for the push4 and push8 problems.}
\label{fig:results:push4_push8}
\end{figure}

Figure~\ref{fig:results:push4_push8} shows the average rank score plots for
push4 and push8.

\section{Ablation Study}
\label{sec:ablation}
\begin{figure}[H]
\centering
\includegraphics[width=0.5\linewidth, clip, trim={5 0 5 7}]{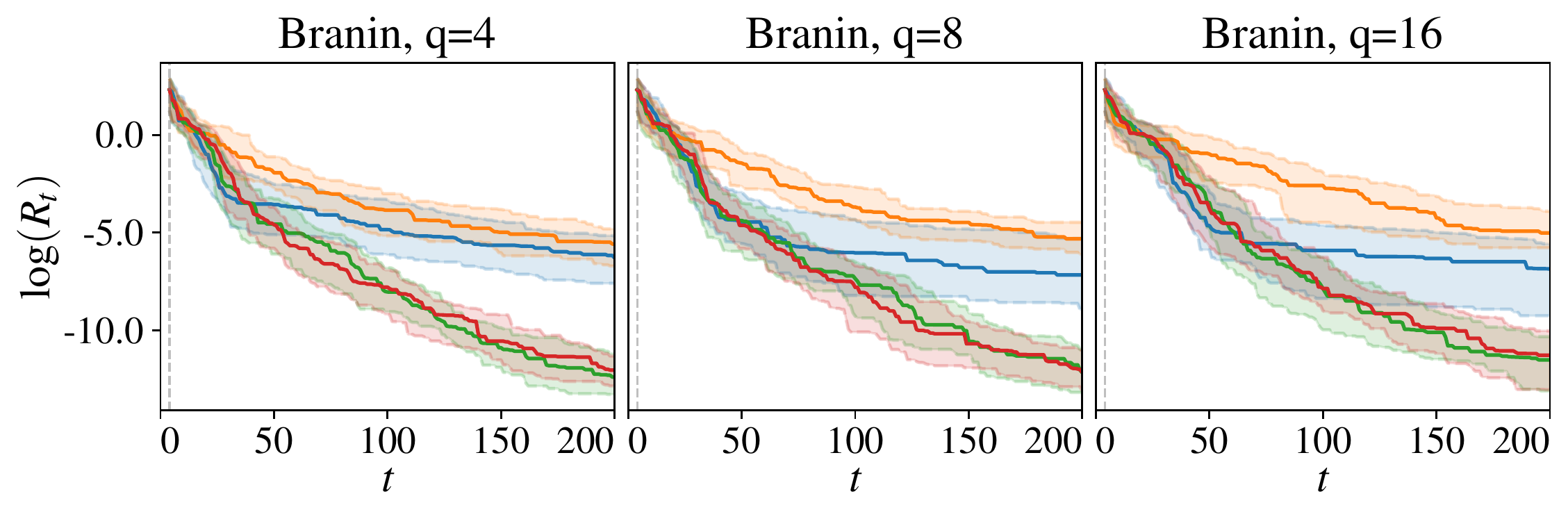}%
\includegraphics[width=0.5\linewidth, clip, trim={5 0 5 7}]{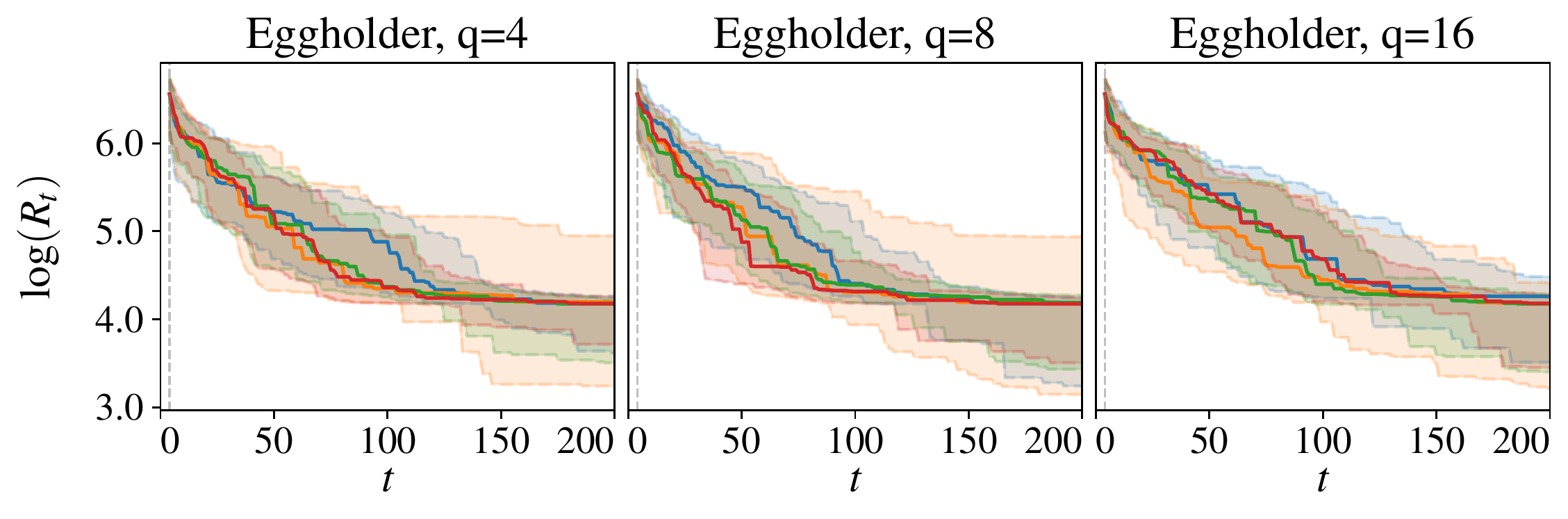}\\
\includegraphics[width=0.5\linewidth, clip, trim={5 0 5 7}]{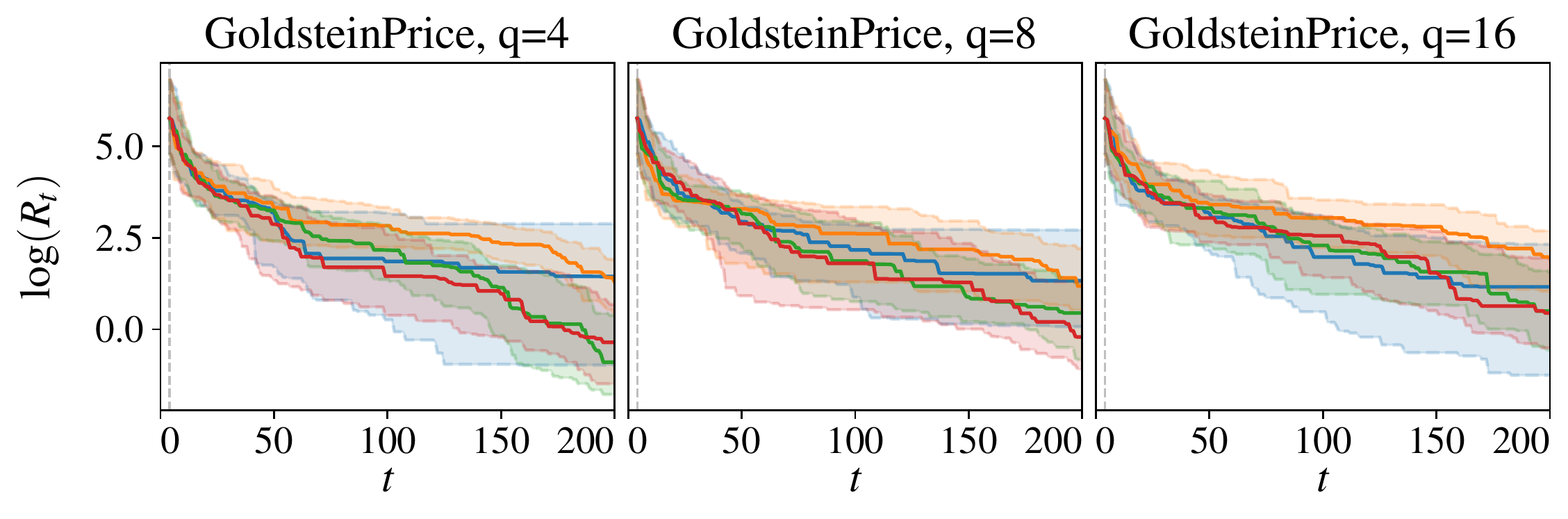}%
\includegraphics[width=0.5\linewidth, clip, trim={5 0 5 7}]{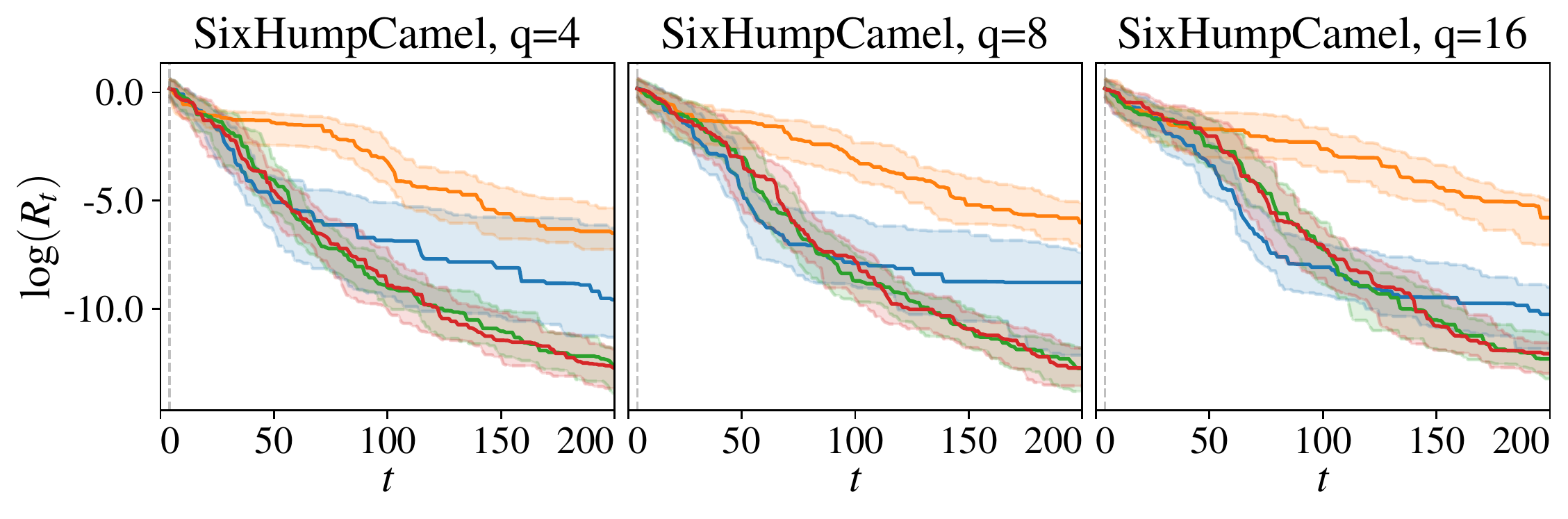}\\
\includegraphics[width=0.5\linewidth, clip, trim={5 0 5 7}]{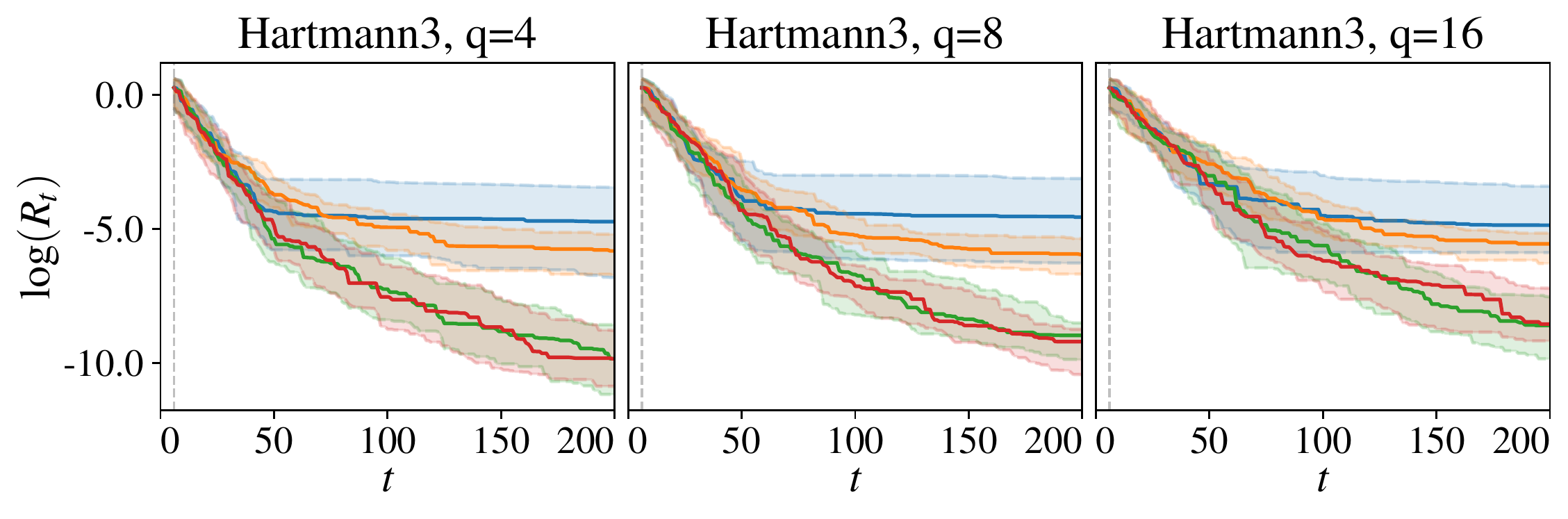}%
\includegraphics[width=0.5\linewidth, clip, trim={5 0 5 7}]{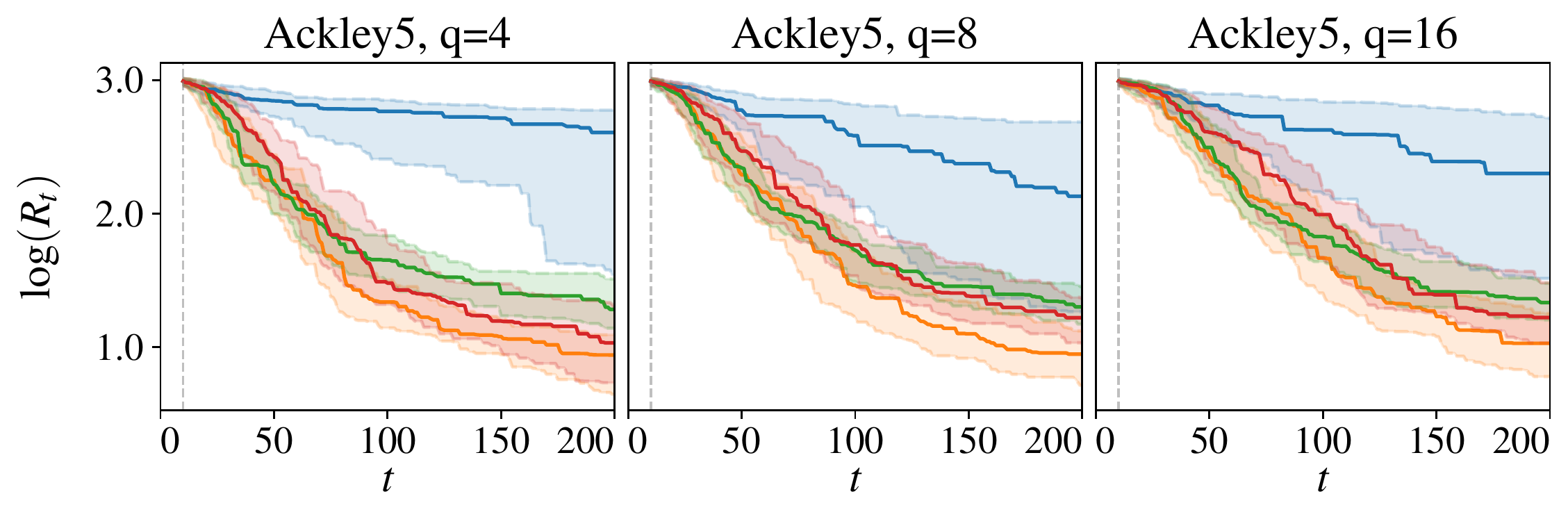}\\
\includegraphics[width=0.5\linewidth, clip, trim={5 0 5 7}]{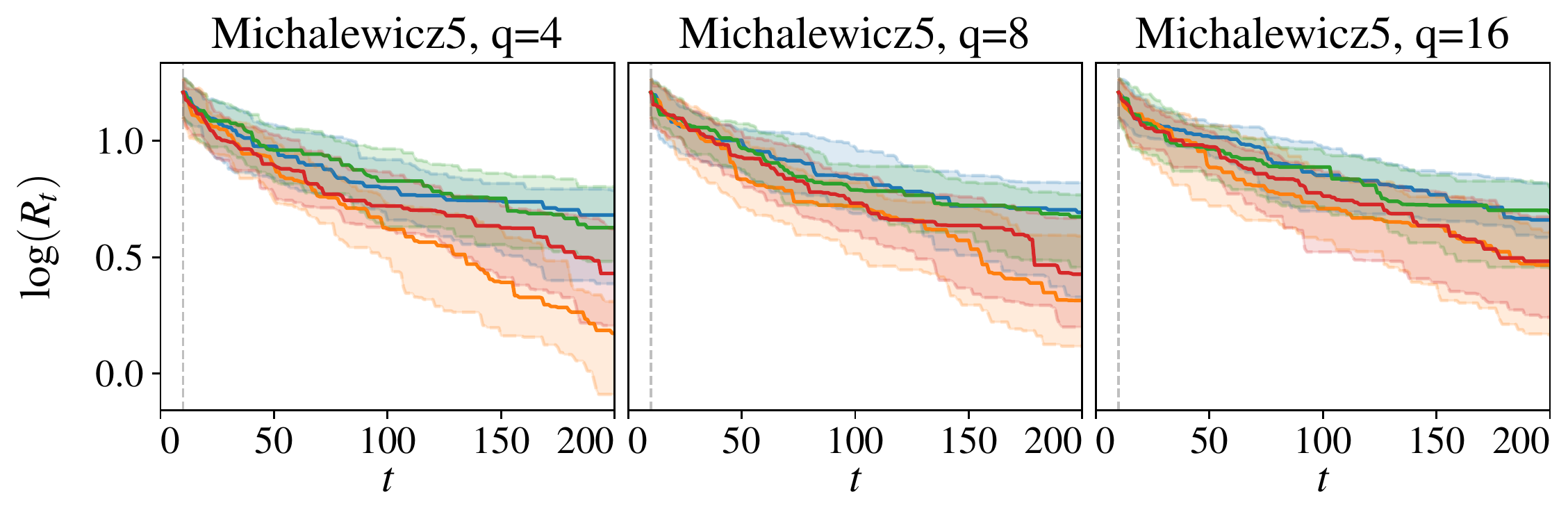}%
\includegraphics[width=0.5\linewidth, clip, trim={5 0 5 7}]{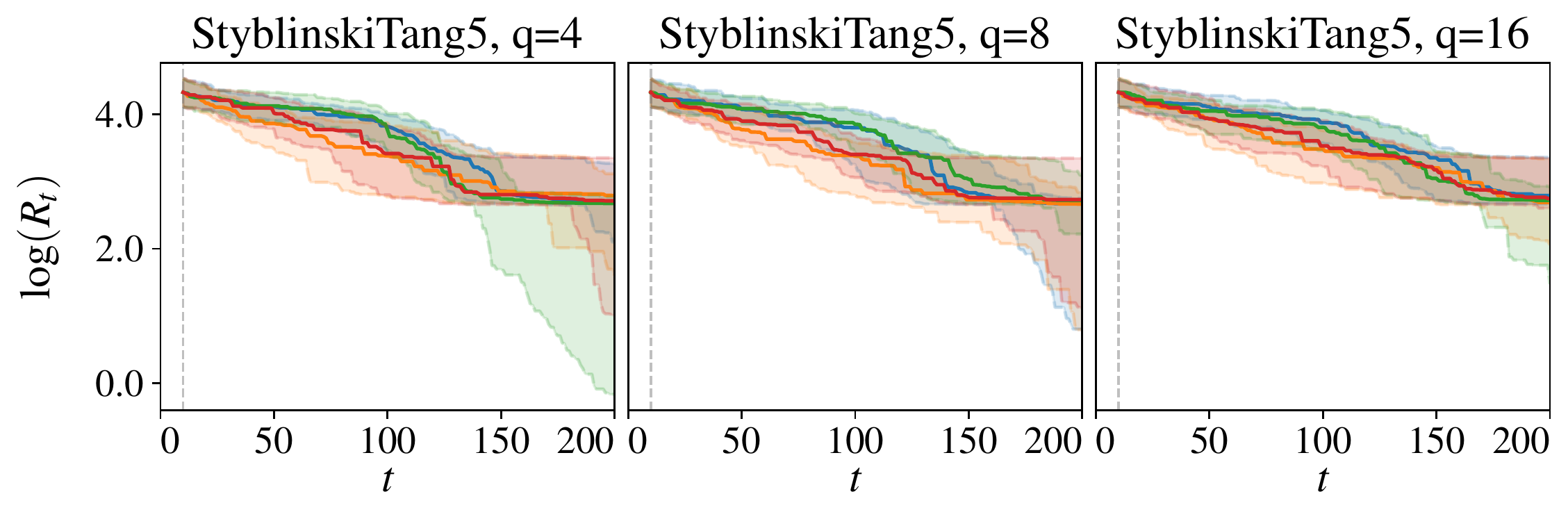}\\
\includegraphics[width=0.5\linewidth, clip, trim={5 0 5 7}]{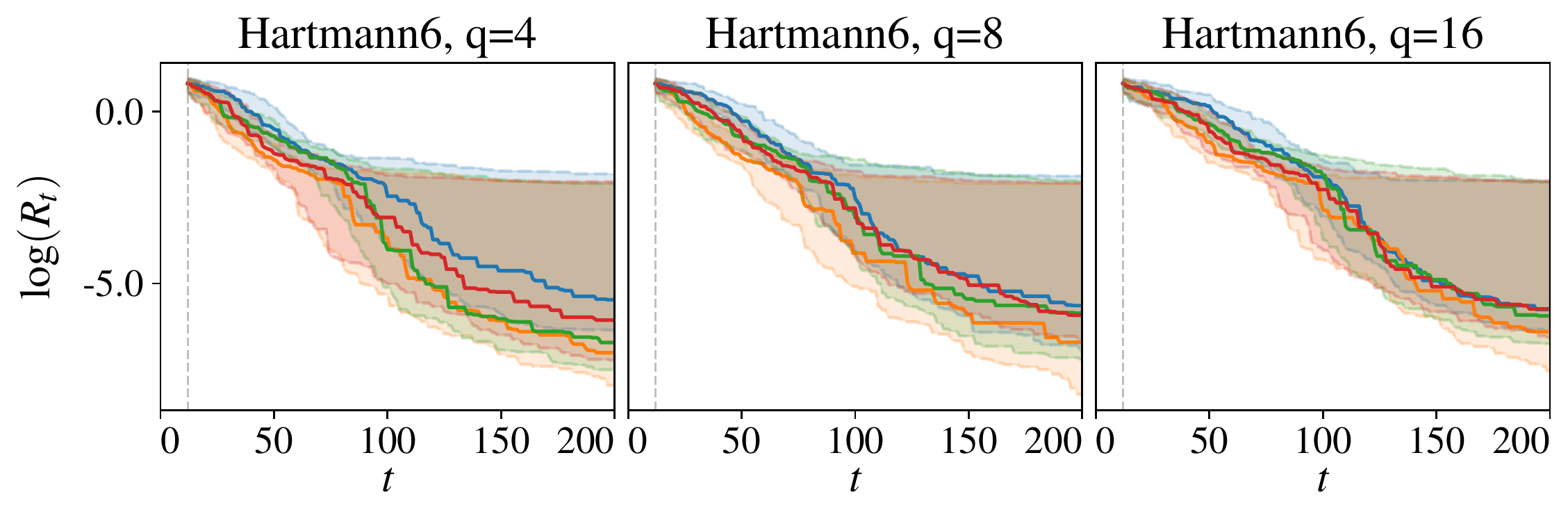}%
\includegraphics[width=0.5\linewidth, clip, trim={5 0 5 7}]{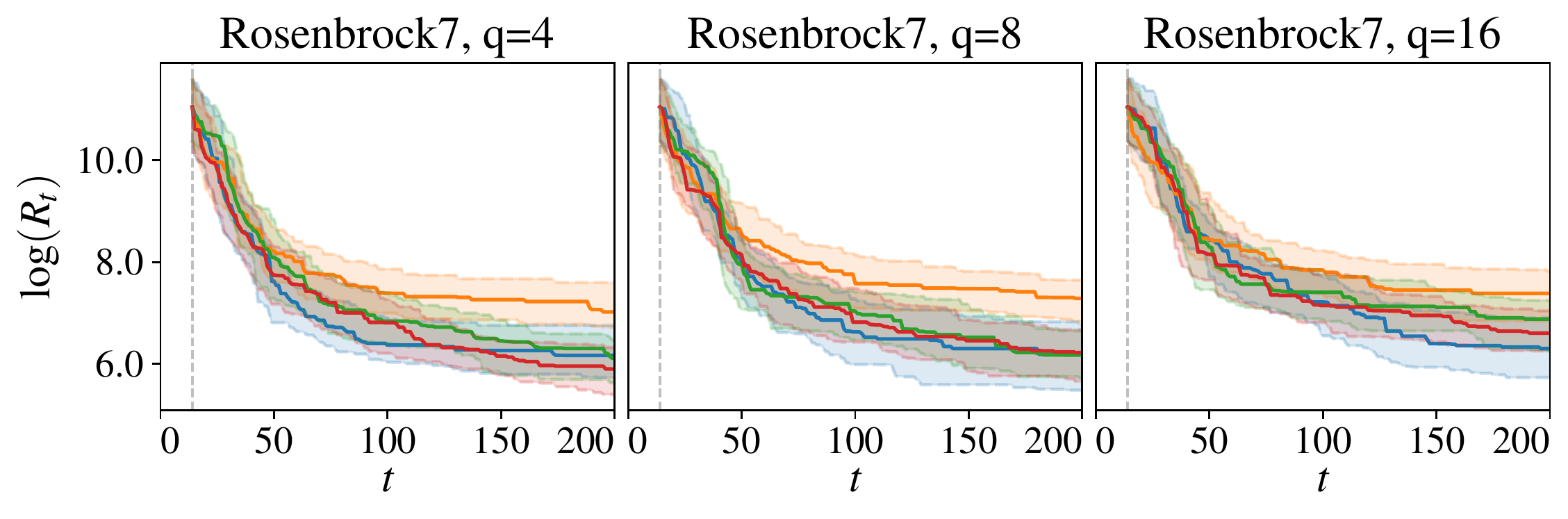}\\
\includegraphics[width=0.5\linewidth, clip, trim={5 0 5 7}]{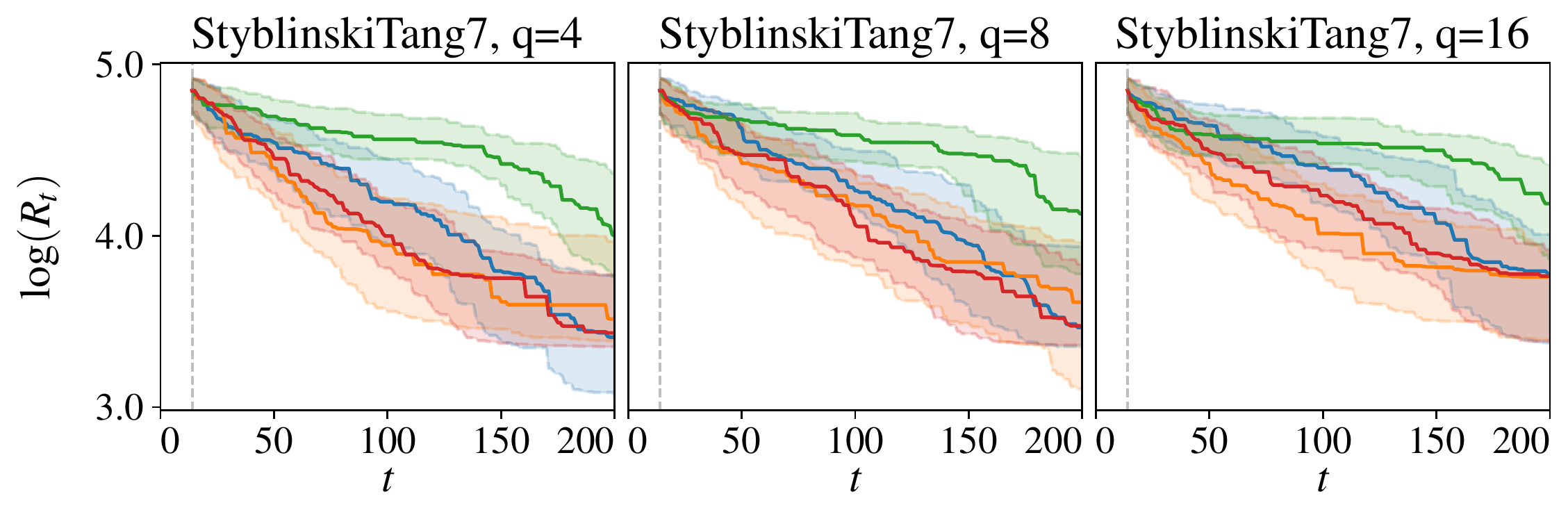}%
\includegraphics[width=0.5\linewidth, clip, trim={5 0 5 7}]{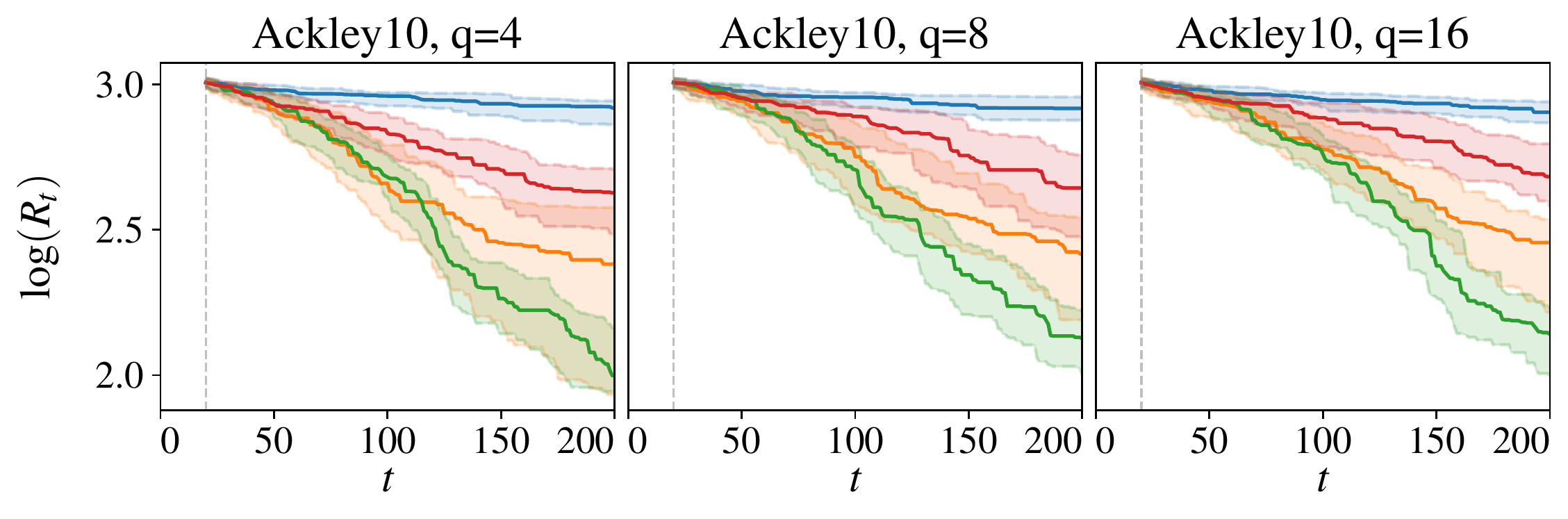}\\
\includegraphics[width=0.5\linewidth, clip, trim={5 0 5 7}]{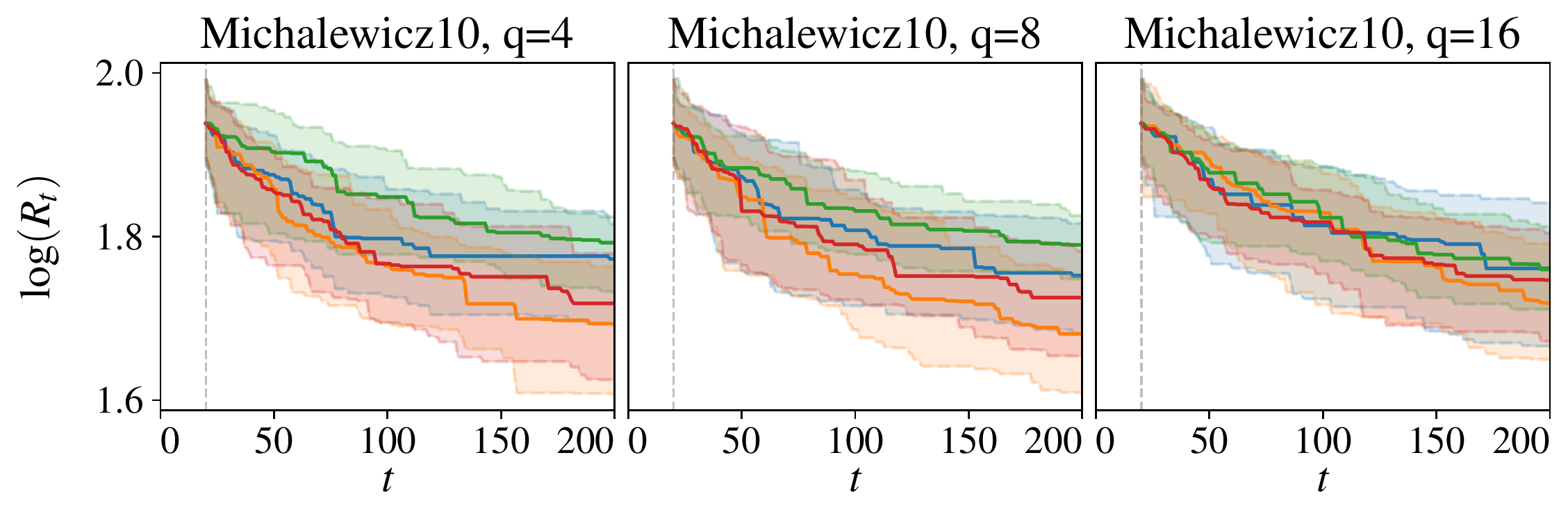}%
\includegraphics[width=0.5\linewidth, clip, trim={5 0 5 7}]{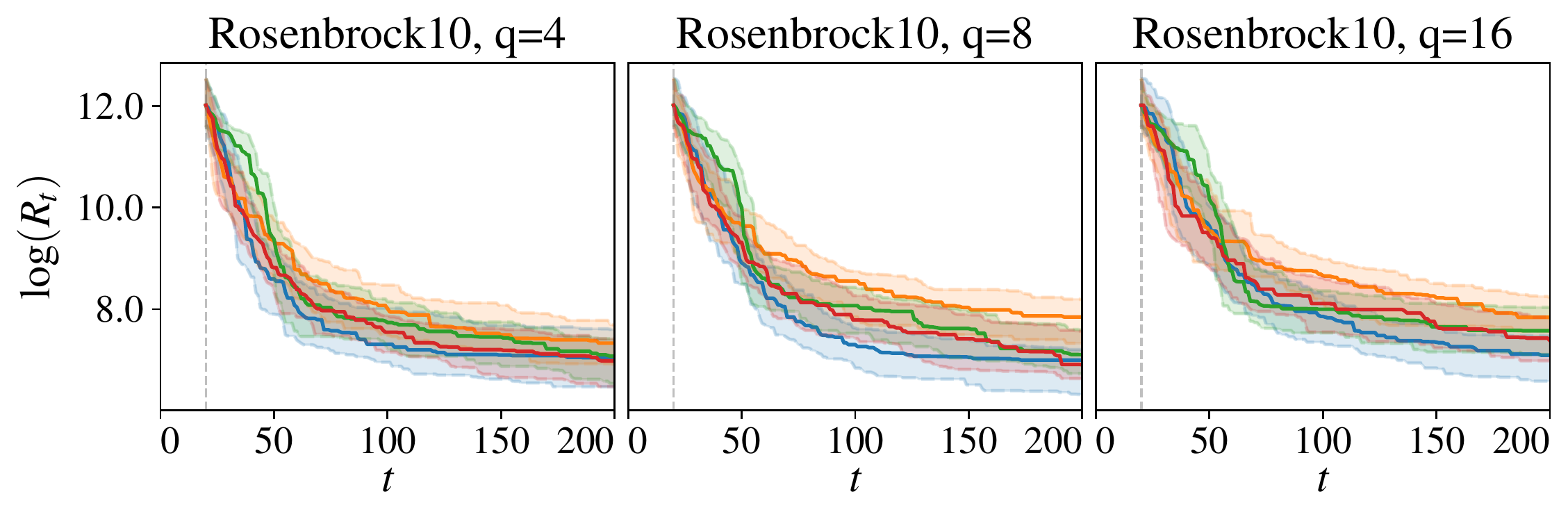}\\
\includegraphics[width=0.5\linewidth, clip, trim={5 0 5 7}]{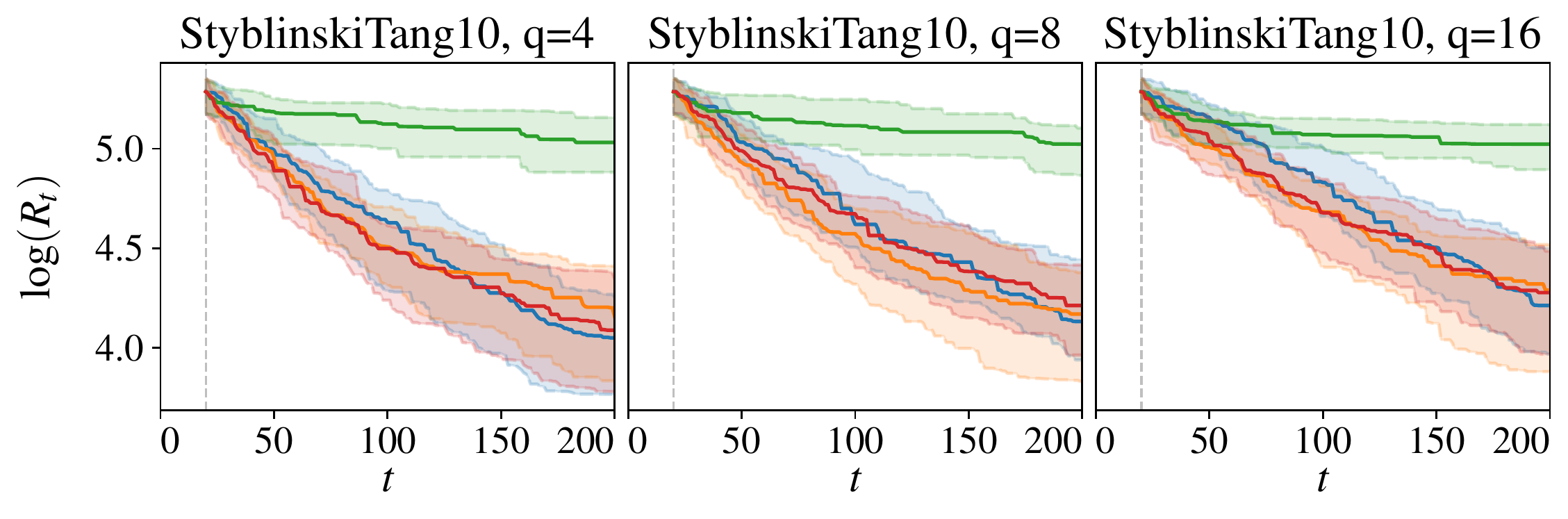}\\
\includegraphics[width=0.4\linewidth, clip, trim={10 15 10 13}]{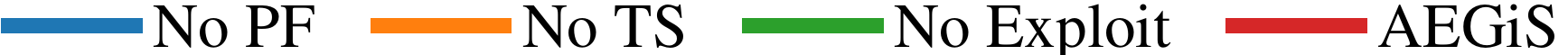}%
\caption{Convergence results for the ablation study.}
\label{fig:results:ablation1}
\end{figure}

In this section we show all convergence plots and results tables for the 
ablation study on AEGiS using the synthetic functions.
Figure~\ref{fig:results:ablation1} shows the convergence 
plots for each of the six methods on the fifteen benchmark problems for
$q \in \{4,8,16\}$. Each plot shows the median log simple regret, with shading 
representing the interquartile range over $51$ runs. 
Tables~\ref{tbl:ablation_results_4}, \ref{tbl:ablation_results_8} and 
\ref{tbl:ablation_results_16} show the median log simple regret as well as the
median absolute deviation from the median (MAD), a robust measure of 
dispersion. The method with the best (lowest) median regret is shown in dark
grey, and those that are statistically equivalent to the best method according
to a one-sided, paired Wilcoxon signed-rank test with Holm-Bonferroni 
correction \citep{holm:test:1979} ($p \geq 0.05$) are shown in light grey.

\begin{table}[H]
\setlength{\tabcolsep}{2pt}
\sisetup{table-format=1.2e-1,table-number-alignment=center}
\caption{Tabulated results for $q = 4$ asynchronous workers, showing the
median log simple regret (\emph{left}) and median absolute deviation from
the median (MAD, \emph{right}) after 200 function evaluations across the 51 runs. 
The method with the lowest median performance is shown in dark grey, with 
those with statistically equivalent performance are shown in light grey.}
\resizebox{1\textwidth}{!}{%
\begin{tabular}{l Sz Sz Sz Sz Sz}
    \toprule
    \bfseries Method
    & \multicolumn{2}{c}{\bfseries Branin (2)} 
    & \multicolumn{2}{c}{\bfseries Eggholder (2)} 
    & \multicolumn{2}{c}{\bfseries GoldsteinPrice (2)} 
    & \multicolumn{2}{c}{\bfseries SixHumpCamel (2)} 
    & \multicolumn{2}{c}{\bfseries Hartmann3 (3)} \\ 
    & \multicolumn{1}{c}{Median} & \multicolumn{1}{c}{MAD}
    & \multicolumn{1}{c}{Median} & \multicolumn{1}{c}{MAD}
    & \multicolumn{1}{c}{Median} & \multicolumn{1}{c}{MAD}
    & \multicolumn{1}{c}{Median} & \multicolumn{1}{c}{MAD}
    & \multicolumn{1}{c}{Median} & \multicolumn{1}{c}{MAD}  \\ \midrule
    No PF & 1.87e-03 & 2.58e-03 & \statsimilar 6.51e+01 & \statsimilar 1.80e+01 & 4.24e+00 & 6.26e+00 & 7.00e-05 & 1.03e-04 & 8.84e-03 & 1.22e-02 \\
    No TS & 3.78e-03 & 4.51e-03 & 6.63e+01 & 8.79e+01 & 3.64e+00 & 3.73e+00 & 1.50e-03 & 1.77e-03 & 2.97e-03 & 2.78e-03 \\
    No Exploit & \best 4.20e-06 & \best 4.73e-06 & \best 6.51e+01 & \best 1.74e+01 & \best 4.08e-01 & \best 5.43e-01 & \statsimilar 3.22e-06 & \statsimilar 4.12e-06 & \statsimilar 5.30e-05 & \statsimilar 6.79e-05 \\
    AEGiS & \statsimilar 5.99e-06 & \statsimilar 6.90e-06 & \statsimilar 6.52e+01 & \statsimilar 1.08e+01 & \statsimilar 6.99e-01 & \statsimilar 7.67e-01 & \best 2.93e-06 & \best 3.31e-06 & \best 5.29e-05 & \best 5.59e-05 \\
\bottomrule
\toprule
    \bfseries Method
    & \multicolumn{2}{c}{\bfseries Ackley5 (5)} 
    & \multicolumn{2}{c}{\bfseries Michalewicz5 (5)} 
    & \multicolumn{2}{c}{\bfseries StyblinskiTang5 (5)} 
    & \multicolumn{2}{c}{\bfseries Hartmann6 (6)} 
    & \multicolumn{2}{c}{\bfseries Rosenbrock7 (7)} \\ 
    & \multicolumn{1}{c}{Median} & \multicolumn{1}{c}{MAD}
    & \multicolumn{1}{c}{Median} & \multicolumn{1}{c}{MAD}
    & \multicolumn{1}{c}{Median} & \multicolumn{1}{c}{MAD}
    & \multicolumn{1}{c}{Median} & \multicolumn{1}{c}{MAD}
    & \multicolumn{1}{c}{Median} & \multicolumn{1}{c}{MAD}  \\ \midrule
    No PF & 1.36e+01 & 4.78e+00 & 1.98e+00 & 4.16e-01 & \statsimilar 1.47e+01 & \statsimilar 1.47e+01 & 4.20e-03 & 5.19e-03 & 4.76e+02 & 3.30e+02 \\
    No TS & \best 2.56e+00 & \best 7.91e-01 & \best 1.19e+00 & \best 3.18e-01 & \statsimilar 1.63e+01 & \statsimilar 1.13e+01 & \best 8.95e-04 & \best 1.13e-03 & 1.12e+03 & 7.51e+02 \\
    No Exploit & 3.61e+00 & 9.86e-01 & 1.87e+00 & 4.42e-01 & \best 1.45e+01 & \best 1.73e+01 & \statsimilar 1.21e-03 & \statsimilar 1.50e-03 & 4.50e+02 & 2.59e+02 \\
    AEGiS & 2.81e+00 & 1.19e+00 & 1.54e+00 & 4.77e-01 & \statsimilar 1.51e+01 & \statsimilar 2.00e+01 & \statsimilar 2.28e-03 & \statsimilar 3.09e-03 & \best 3.64e+02 & \best 2.44e+02 \\
\bottomrule
\toprule
    \bfseries Method
    & \multicolumn{2}{c}{\bfseries StyblinskiTang7 (7)} 
    & \multicolumn{2}{c}{\bfseries Ackley10 (10)} 
    & \multicolumn{2}{c}{\bfseries Michalewicz10 (10)} 
    & \multicolumn{2}{c}{\bfseries Rosenbrock10 (10)} 
    & \multicolumn{2}{c}{\bfseries StyblinskiTang10 (10)} \\ 
    & \multicolumn{1}{c}{Median} & \multicolumn{1}{c}{MAD}
    & \multicolumn{1}{c}{Median} & \multicolumn{1}{c}{MAD}
    & \multicolumn{1}{c}{Median} & \multicolumn{1}{c}{MAD}
    & \multicolumn{1}{c}{Median} & \multicolumn{1}{c}{MAD}
    & \multicolumn{1}{c}{Median} & \multicolumn{1}{c}{MAD}  \\ \midrule
    No PF & \best 3.02e+01 & \best 1.95e+01 & 1.85e+01 & 1.09e+00 & 5.89e+00 & 5.40e-01 & \statsimilar 1.15e+03 & \statsimilar 9.62e+02 & \best 5.74e+01 & \best 2.10e+01 \\
    No TS & \statsimilar 3.36e+01 & \statsimilar 2.25e+01 & 1.08e+01 & 3.93e+00 & \best 5.43e+00 & \best 6.51e-01 & 1.53e+03 & 8.32e+02 & 6.39e+01 & 2.64e+01 \\
    No Exploit & 5.46e+01 & 2.71e+01 & \best 7.38e+00 & \best 1.39e+00 & 6.01e+00 & 4.58e-01 & \statsimilar 1.18e+03 & \statsimilar 7.14e+02 & 1.53e+02 & 3.35e+01 \\
    AEGiS & \statsimilar 3.10e+01 & \statsimilar 1.79e+01 & 1.38e+01 & 2.30e+00 & \statsimilar 5.57e+00 & \statsimilar 7.16e-01 & \best 1.08e+03 & \best 7.27e+02 & \statsimilar 5.96e+01 & \statsimilar 2.41e+01 \\
\bottomrule
\end{tabular}
}
\label{tbl:ablation_results_4}
\end{table}

\begin{table}[H]
\setlength{\tabcolsep}{2pt}
\sisetup{table-format=1.2e-1,table-number-alignment=center}
\caption{Tabulated results for $q = 8$ asynchronous workers, showing the
median log simple regret (\emph{left}) and median absolute deviation from
the median (MAD, \emph{right}) after 200 function evaluations across the 51 runs. 
The method with the lowest median performance is shown in dark grey, with 
those with statistically equivalent performance are shown in light grey.}
\resizebox{1\textwidth}{!}{%
\begin{tabular}{l Sz Sz Sz Sz Sz}
    \toprule
    \bfseries Method
    & \multicolumn{2}{c}{\bfseries Branin (2)} 
    & \multicolumn{2}{c}{\bfseries Eggholder (2)} 
    & \multicolumn{2}{c}{\bfseries GoldsteinPrice (2)} 
    & \multicolumn{2}{c}{\bfseries SixHumpCamel (2)} 
    & \multicolumn{2}{c}{\bfseries Hartmann3 (3)} \\ 
    & \multicolumn{1}{c}{Median} & \multicolumn{1}{c}{MAD}
    & \multicolumn{1}{c}{Median} & \multicolumn{1}{c}{MAD}
    & \multicolumn{1}{c}{Median} & \multicolumn{1}{c}{MAD}
    & \multicolumn{1}{c}{Median} & \multicolumn{1}{c}{MAD}
    & \multicolumn{1}{c}{Median} & \multicolumn{1}{c}{MAD}  \\ \midrule
    No PF & 7.73e-04 & 1.14e-03 & \statsimilar 6.51e+01 & \statsimilar 3.39e+01 & 3.78e+00 & 5.46e+00 & 1.55e-04 & 2.26e-04 & 1.05e-02 & 1.44e-02 \\
    No TS & 4.90e-03 & 5.69e-03 & \statsimilar 6.54e+01 & \statsimilar 7.27e+01 & 3.26e+00 & 3.61e+00 & 2.44e-03 & 2.86e-03 & 2.60e-03 & 2.39e-03 \\
    No Exploit & \best 5.28e-06 & \best 6.77e-06 & \statsimilar 6.60e+01 & \statsimilar 2.61e+01 & \statsimilar 1.56e+00 & \statsimilar 1.95e+00 & \best 2.94e-06 & \best 3.49e-06 & \statsimilar 1.26e-04 & \statsimilar 1.13e-04 \\
    AEGiS & \statsimilar 5.32e-06 & \statsimilar 6.51e-06 & \best 6.51e+01 & \best 1.35e+01 & \best 8.10e-01 & \best 8.82e-01 & \statsimilar 2.98e-06 & \statsimilar 3.83e-06 & \best 9.99e-05 & \best 1.03e-04 \\
\bottomrule
\toprule
    \bfseries Method
    & \multicolumn{2}{c}{\bfseries Ackley5 (5)} 
    & \multicolumn{2}{c}{\bfseries Michalewicz5 (5)} 
    & \multicolumn{2}{c}{\bfseries StyblinskiTang5 (5)} 
    & \multicolumn{2}{c}{\bfseries Hartmann6 (6)} 
    & \multicolumn{2}{c}{\bfseries Rosenbrock7 (7)} \\ 
    & \multicolumn{1}{c}{Median} & \multicolumn{1}{c}{MAD}
    & \multicolumn{1}{c}{Median} & \multicolumn{1}{c}{MAD}
    & \multicolumn{1}{c}{Median} & \multicolumn{1}{c}{MAD}
    & \multicolumn{1}{c}{Median} & \multicolumn{1}{c}{MAD}
    & \multicolumn{1}{c}{Median} & \multicolumn{1}{c}{MAD}  \\ \midrule
    No PF & 8.41e+00 & 8.03e+00 & 2.00e+00 & 4.95e-01 & \statsimilar 1.44e+01 & \statsimilar 3.17e+00 & 3.54e-03 & 4.78e-03 & \statsimilar 4.85e+02 & \statsimilar 4.39e+02 \\
    No TS & \best 2.58e+00 & \best 7.86e-01 & \best 1.37e+00 & \best 4.58e-01 & \best 1.44e+01 & \best 1.23e+01 & \best 1.22e-03 & \best 1.60e-03 & 1.46e+03 & 9.07e+02 \\
    No Exploit & 3.68e+00 & 7.93e-01 & 1.96e+00 & 3.87e-01 & \statsimilar 1.51e+01 & \statsimilar 1.31e+01 & 2.83e-03 & 3.96e-03 & \best 4.79e+02 & \best 3.09e+02 \\
    AEGiS & 3.39e+00 & 8.01e-01 & \statsimilar 1.53e+00 & \statsimilar 5.62e-01 & \statsimilar 1.53e+01 & \statsimilar 1.89e+01 & 2.67e-03 & 3.16e-03 & \statsimilar 5.04e+02 & \statsimilar 3.54e+02 \\
\bottomrule
\toprule
    \bfseries Method
    & \multicolumn{2}{c}{\bfseries StyblinskiTang7 (7)} 
    & \multicolumn{2}{c}{\bfseries Ackley10 (10)} 
    & \multicolumn{2}{c}{\bfseries Michalewicz10 (10)} 
    & \multicolumn{2}{c}{\bfseries Rosenbrock10 (10)} 
    & \multicolumn{2}{c}{\bfseries StyblinskiTang10 (10)} \\ 
    & \multicolumn{1}{c}{Median} & \multicolumn{1}{c}{MAD}
    & \multicolumn{1}{c}{Median} & \multicolumn{1}{c}{MAD}
    & \multicolumn{1}{c}{Median} & \multicolumn{1}{c}{MAD}
    & \multicolumn{1}{c}{Median} & \multicolumn{1}{c}{MAD}
    & \multicolumn{1}{c}{Median} & \multicolumn{1}{c}{MAD}  \\ \midrule
    No PF & \best 3.20e+01 & \best 1.80e+01 & 1.85e+01 & 1.09e+00 & 5.77e+00 & 5.89e-01 & \statsimilar 1.09e+03 & \statsimilar 7.33e+02 & \best 6.22e+01 & \best 2.28e+01 \\
    No TS & \statsimilar 3.70e+01 & \statsimilar 2.21e+01 & 1.12e+01 & 2.31e+00 & \best 5.37e+00 & \best 5.73e-01 & 2.55e+03 & 1.58e+03 & \statsimilar 6.46e+01 & \statsimilar 2.72e+01 \\
    No Exploit & 6.21e+01 & 3.51e+01 & \best 8.39e+00 & \best 1.34e+00 & 5.99e+00 & 3.30e-01 & \statsimilar 1.22e+03 & \statsimilar 6.31e+02 & 1.52e+02 & 2.54e+01 \\
    AEGiS & \statsimilar 3.23e+01 & \statsimilar 1.69e+01 & 1.41e+01 & 2.83e+00 & \statsimilar 5.61e+00 & \statsimilar 5.60e-01 & \best 9.97e+02 & \best 5.96e+02 & \statsimilar 6.75e+01 & \statsimilar 2.31e+01 \\
\bottomrule
\end{tabular}
}
\label{tbl:ablation_results_8}
\end{table}

\begin{table}[H]
\setlength{\tabcolsep}{2pt}
\sisetup{table-format=1.2e-1,table-number-alignment=center}
\caption{Tabulated results for $q = 16$ asynchronous workers, showing the
median log simple regret (\emph{left}) and median absolute deviation from
the median (MAD, \emph{right}) after 200 function evaluations across the 51 runs. 
The method with the lowest median performance is shown in dark grey, with 
those with statistically equivalent performance are shown in light grey.}
\resizebox{1\textwidth}{!}{%
\begin{tabular}{l Sz Sz Sz Sz Sz}
    \toprule
    \bfseries Method
    & \multicolumn{2}{c}{\bfseries Branin (2)} 
    & \multicolumn{2}{c}{\bfseries Eggholder (2)} 
    & \multicolumn{2}{c}{\bfseries GoldsteinPrice (2)} 
    & \multicolumn{2}{c}{\bfseries SixHumpCamel (2)} 
    & \multicolumn{2}{c}{\bfseries Hartmann3 (3)} \\ 
    & \multicolumn{1}{c}{Median} & \multicolumn{1}{c}{MAD}
    & \multicolumn{1}{c}{Median} & \multicolumn{1}{c}{MAD}
    & \multicolumn{1}{c}{Median} & \multicolumn{1}{c}{MAD}
    & \multicolumn{1}{c}{Median} & \multicolumn{1}{c}{MAD}
    & \multicolumn{1}{c}{Median} & \multicolumn{1}{c}{MAD}  \\ \midrule
    No PF & 1.06e-03 & 1.55e-03 & \statsimilar 7.07e+01 & \statsimilar 4.58e+01 & \statsimilar 3.19e+00 & \statsimilar 4.63e+00 & 3.56e-05 & 4.69e-05 & 7.75e-03 & 1.13e-02 \\
    No TS & 6.67e-03 & 7.63e-03 & \statsimilar 6.51e+01 & \statsimilar 5.92e+01 & 7.21e+00 & 8.09e+00 & 3.09e-03 & 4.11e-03 & 3.86e-03 & 2.95e-03 \\
    No Exploit & \best 1.01e-05 & \best 1.36e-05 & \best 6.51e+01 & \best 3.15e+01 & \statsimilar 1.65e+00 & \statsimilar 2.05e+00 & \best 4.54e-06 & \best 5.45e-06 & \best 1.83e-04 & \best 2.41e-04 \\
    AEGiS & \statsimilar 1.28e-05 & \statsimilar 1.80e-05 & \statsimilar 6.53e+01 & \statsimilar 1.27e+01 & \best 1.56e+00 & \best 1.84e+00 & \statsimilar 5.82e-06 & \statsimilar 5.32e-06 & \statsimilar 1.93e-04 & \statsimilar 2.80e-04 \\
\bottomrule
\toprule
    \bfseries Method
    & \multicolumn{2}{c}{\bfseries Ackley5 (5)} 
    & \multicolumn{2}{c}{\bfseries Michalewicz5 (5)} 
    & \multicolumn{2}{c}{\bfseries StyblinskiTang5 (5)} 
    & \multicolumn{2}{c}{\bfseries Hartmann6 (6)} 
    & \multicolumn{2}{c}{\bfseries Rosenbrock7 (7)} \\ 
    & \multicolumn{1}{c}{Median} & \multicolumn{1}{c}{MAD}
    & \multicolumn{1}{c}{Median} & \multicolumn{1}{c}{MAD}
    & \multicolumn{1}{c}{Median} & \multicolumn{1}{c}{MAD}
    & \multicolumn{1}{c}{Median} & \multicolumn{1}{c}{MAD}
    & \multicolumn{1}{c}{Median} & \multicolumn{1}{c}{MAD}  \\ \midrule
    No PF & 9.98e+00 & 8.05e+00 & 1.93e+00 & 2.98e-01 & \statsimilar 1.63e+01 & \statsimilar 1.82e+01 & \statsimilar 3.16e-03 & \statsimilar 3.50e-03 & \best 5.44e+02 & \best 4.00e+02 \\
    No TS & \best 2.80e+00 & \best 9.83e-01 & \best 1.59e+00 & \best 4.78e-01 & \best 1.47e+01 & \best 1.23e+01 & \best 1.66e-03 & \best 1.90e-03 & 1.61e+03 & 1.13e+03 \\
    No Exploit & 3.80e+00 & 7.76e-01 & 1.99e+00 & 4.63e-01 & \statsimilar 1.51e+01 & \statsimilar 6.02e+00 & \statsimilar 2.62e-03 & \statsimilar 3.46e-03 & 9.71e+02 & 6.76e+02 \\
    AEGiS & 3.39e+00 & 1.12e+00 & \statsimilar 1.62e+00 & \statsimilar 5.10e-01 & \statsimilar 1.56e+01 & \statsimilar 1.09e+01 & 3.21e-03 & 3.82e-03 & \statsimilar 7.38e+02 & \statsimilar 5.13e+02 \\
\bottomrule
\toprule
    \bfseries Method
    & \multicolumn{2}{c}{\bfseries StyblinskiTang7 (7)} 
    & \multicolumn{2}{c}{\bfseries Ackley10 (10)} 
    & \multicolumn{2}{c}{\bfseries Michalewicz10 (10)} 
    & \multicolumn{2}{c}{\bfseries Rosenbrock10 (10)} 
    & \multicolumn{2}{c}{\bfseries StyblinskiTang10 (10)} \\ 
    & \multicolumn{1}{c}{Median} & \multicolumn{1}{c}{MAD}
    & \multicolumn{1}{c}{Median} & \multicolumn{1}{c}{MAD}
    & \multicolumn{1}{c}{Median} & \multicolumn{1}{c}{MAD}
    & \multicolumn{1}{c}{Median} & \multicolumn{1}{c}{MAD}
    & \multicolumn{1}{c}{Median} & \multicolumn{1}{c}{MAD}  \\ \midrule
    No PF & \statsimilar 4.35e+01 & \statsimilar 2.08e+01 & 1.83e+01 & 9.93e-01 & \statsimilar 5.82e+00 & \statsimilar 7.68e-01 & \best 1.20e+03 & \best 8.13e+02 & \best 6.75e+01 & \best 2.87e+01 \\
    No TS & \best 4.29e+01 & \best 1.89e+01 & 1.17e+01 & 2.85e+00 & \best 5.58e+00 & \best 5.76e-01 & 2.53e+03 & 1.45e+03 & \statsimilar 7.29e+01 & \statsimilar 3.20e+01 \\
    No Exploit & 6.58e+01 & 2.60e+01 & \best 8.50e+00 & \best 1.54e+00 & 5.81e+00 & 4.38e-01 & 1.95e+03 & 1.27e+03 & 1.52e+02 & 2.42e+01 \\
    AEGiS & \statsimilar 4.32e+01 & \statsimilar 1.80e+01 & 1.46e+01 & 2.22e+00 & \statsimilar 5.74e+00 & \statsimilar 5.74e-01 & 1.60e+03 & 1.01e+03 & \statsimilar 7.20e+01 & \statsimilar 2.83e+01 \\
\bottomrule
\end{tabular}
}
\label{tbl:ablation_results_16}
\end{table}

\section{Selecting \eps}
\label{sec:settingeps}
\begin{figure}[H] %
\centering
\includegraphics[width=0.8\linewidth, clip, trim={5 5 5 7}]{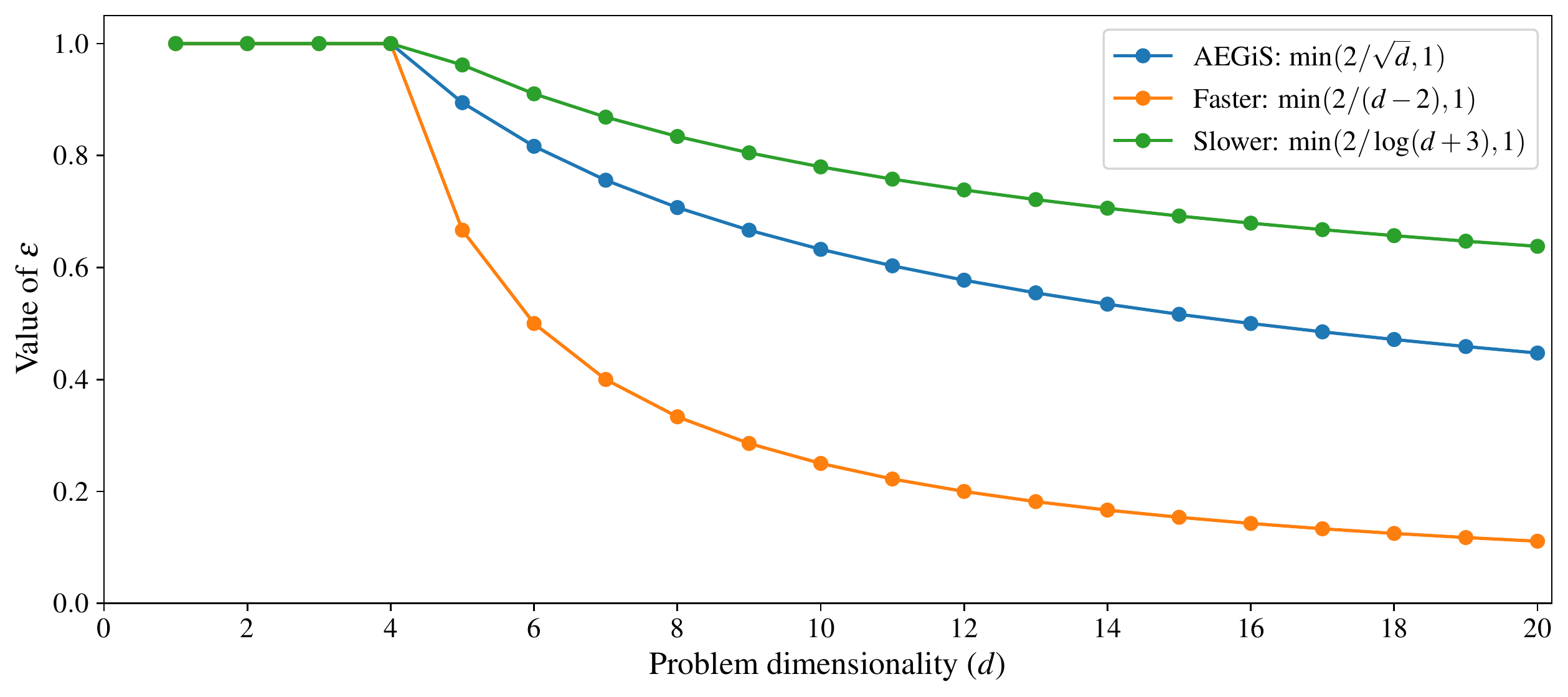}%
\caption{Epsilon decay rate for the three evaluated heuristics. Here,
\emph{faster} refers to a higher rate of decay and thus more exploitation,
whereas \emph{slower} refers to a lower rate of decay and therefore more
exploration.}
\label{fig:results:epsdecay}
\end{figure}

\begin{figure}[H]
\centering
\includegraphics[width=0.5\linewidth, clip, trim={5 0 5 7}]{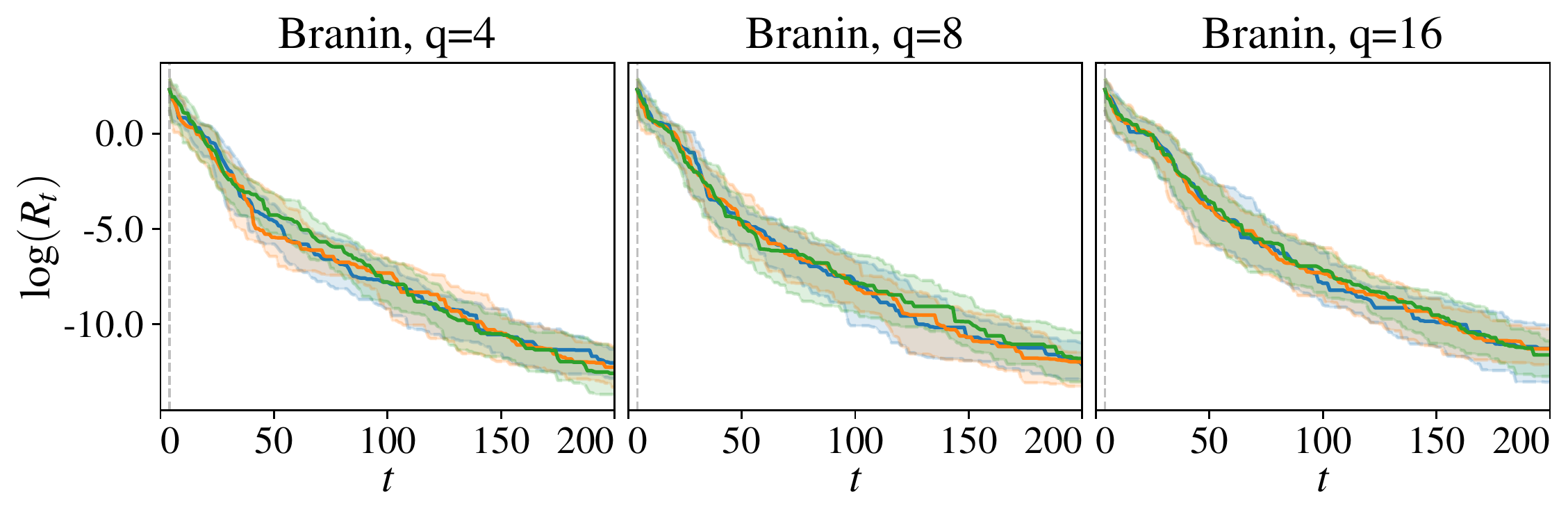}%
\includegraphics[width=0.5\linewidth, clip, trim={5 0 5 7}]{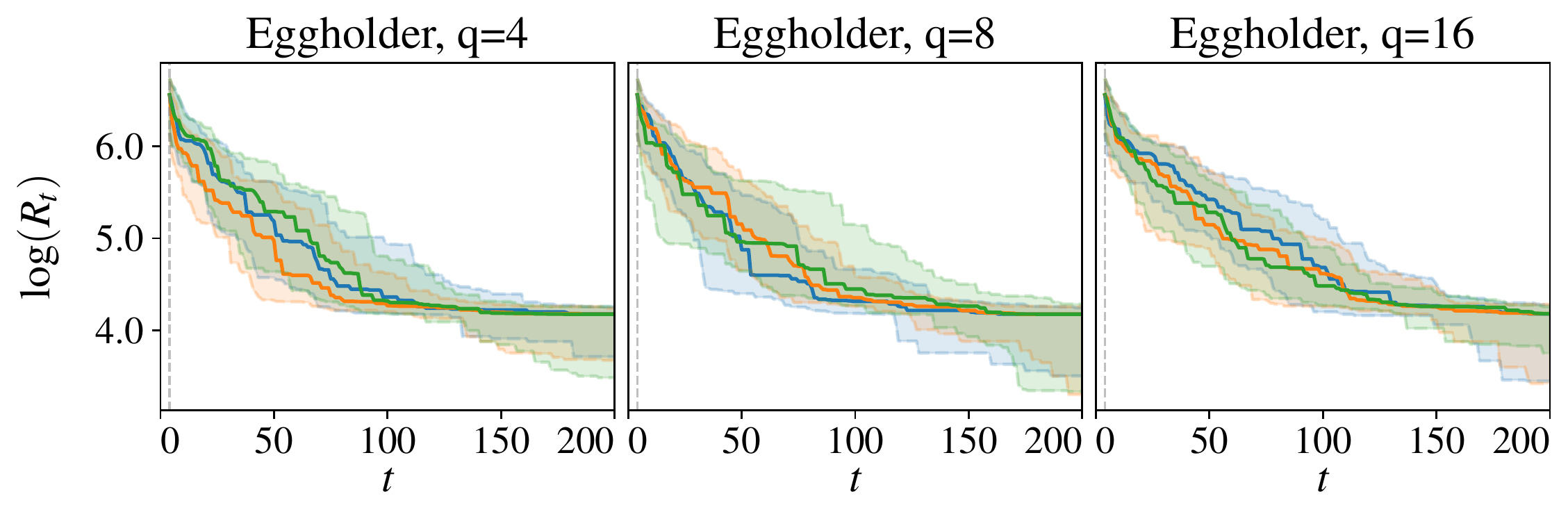}\\
\includegraphics[width=0.5\linewidth, clip, trim={5 0 5 7}]{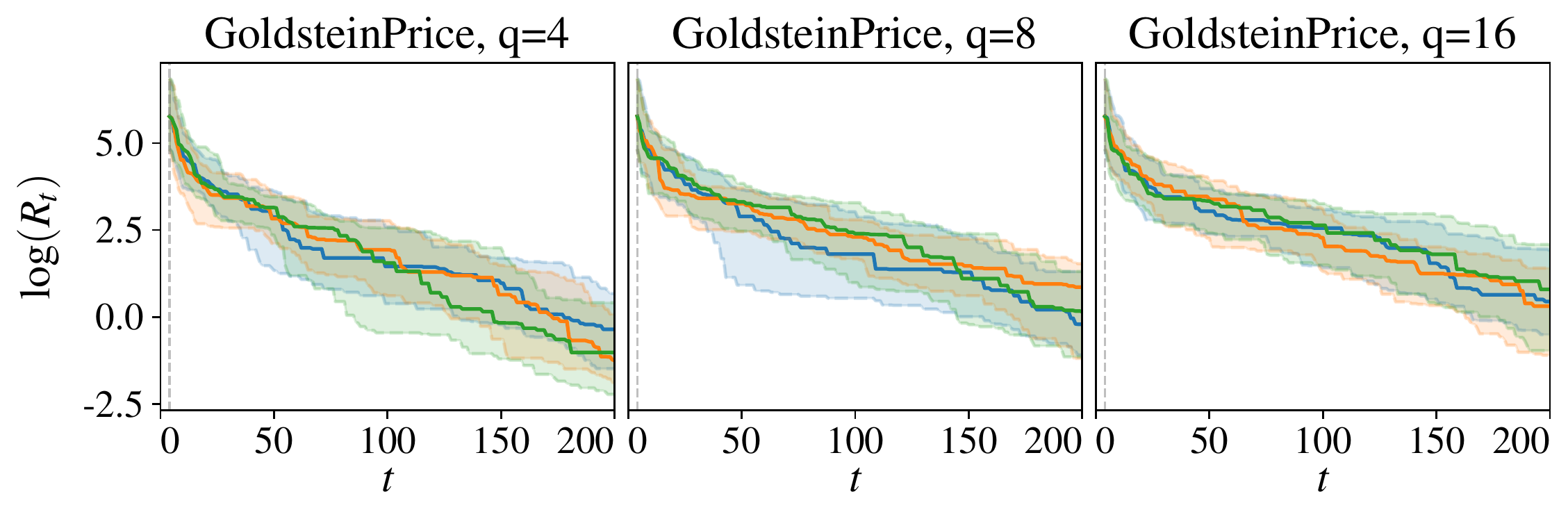}%
\includegraphics[width=0.5\linewidth, clip, trim={5 0 5 7}]{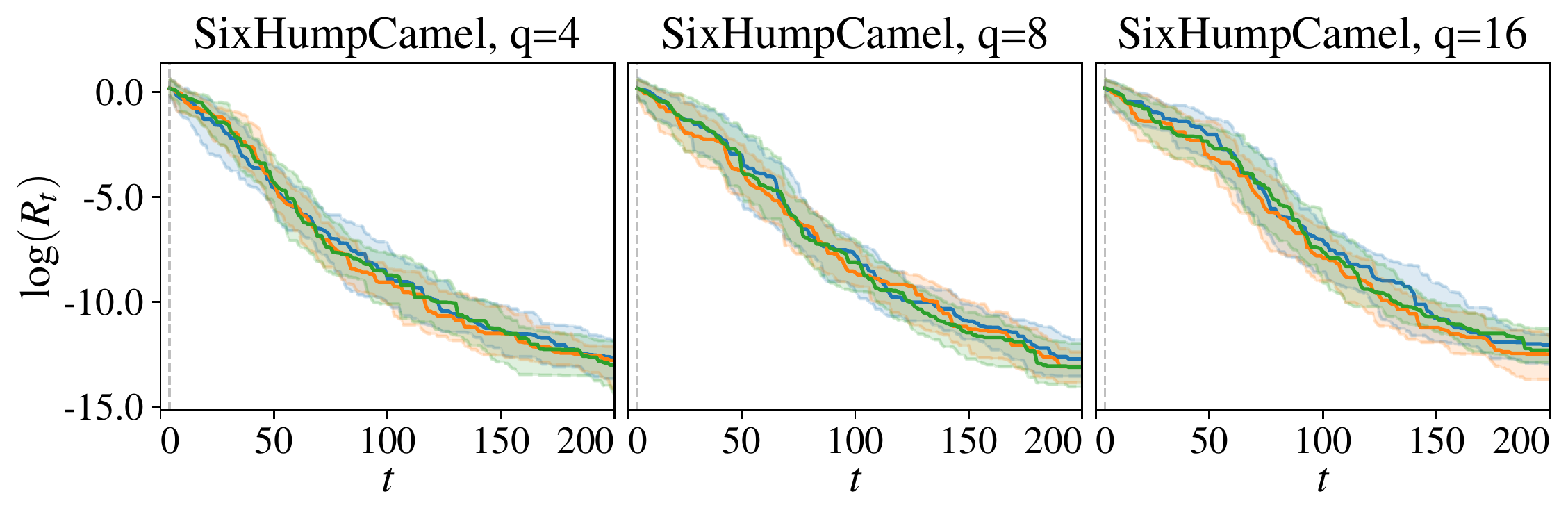}\\
\includegraphics[width=0.5\linewidth, clip, trim={5 0 5 7}]{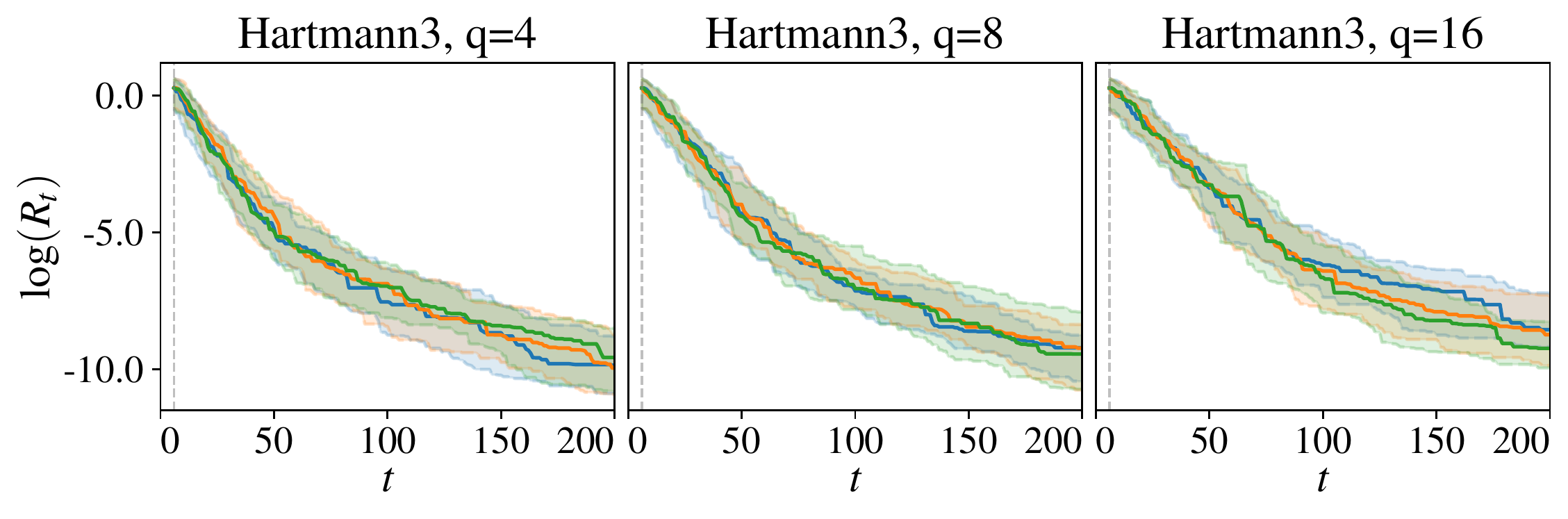}%
\includegraphics[width=0.5\linewidth, clip, trim={5 0 5 7}]{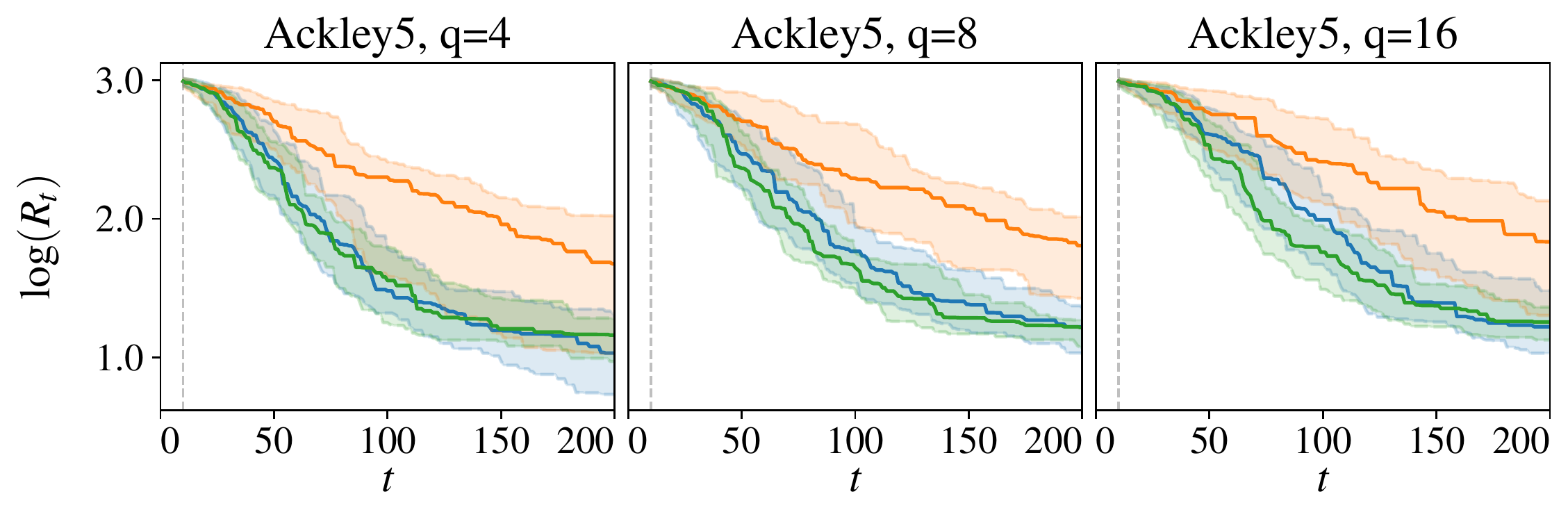}\\
\includegraphics[width=0.5\linewidth, clip, trim={5 0 5 7}]{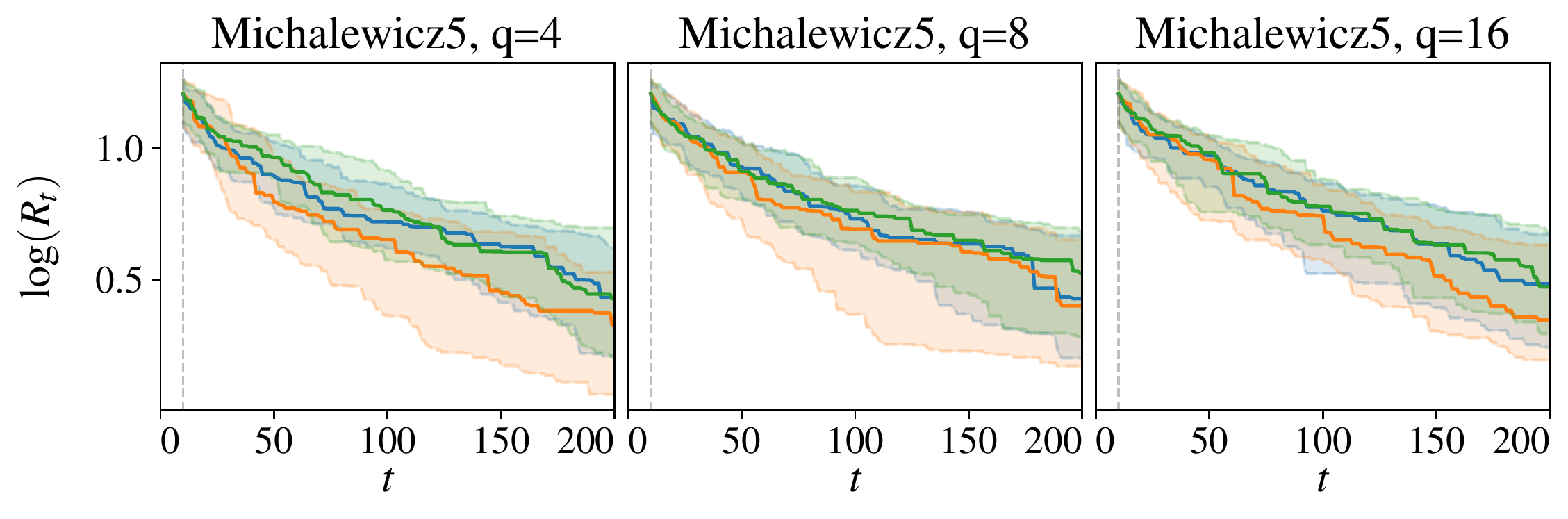}%
\includegraphics[width=0.5\linewidth, clip, trim={5 0 5 7}]{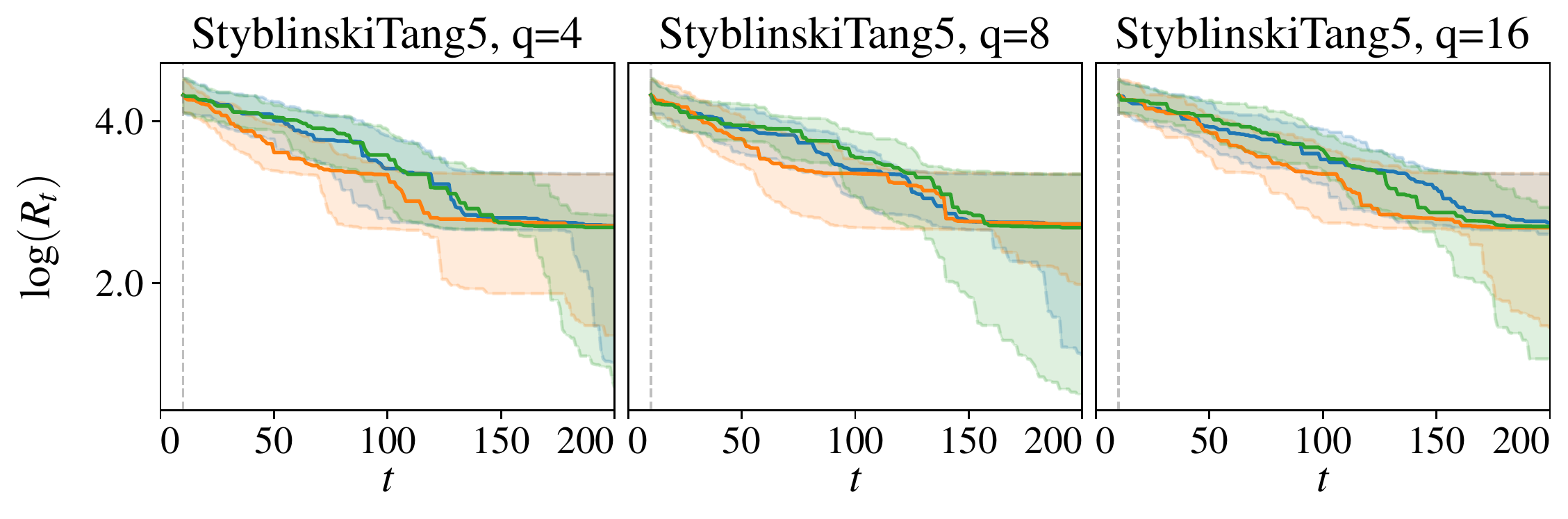}\\
\includegraphics[width=0.5\linewidth, clip, trim={5 0 5 7}]{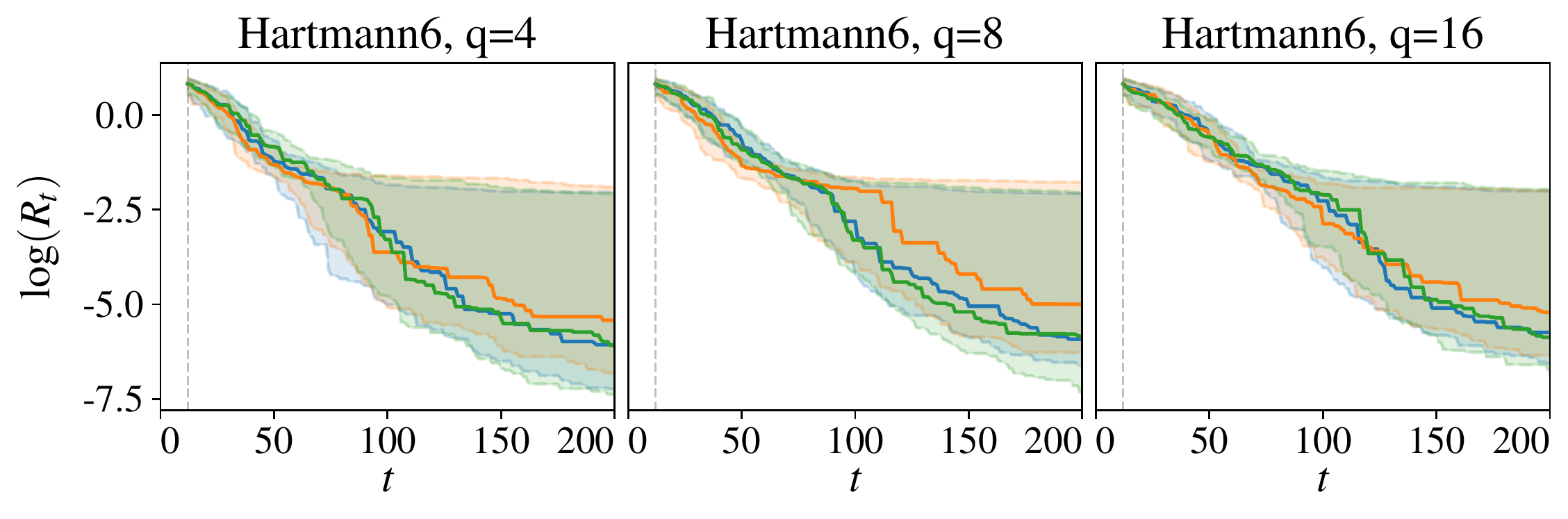}%
\includegraphics[width=0.5\linewidth, clip, trim={5 0 5 7}]{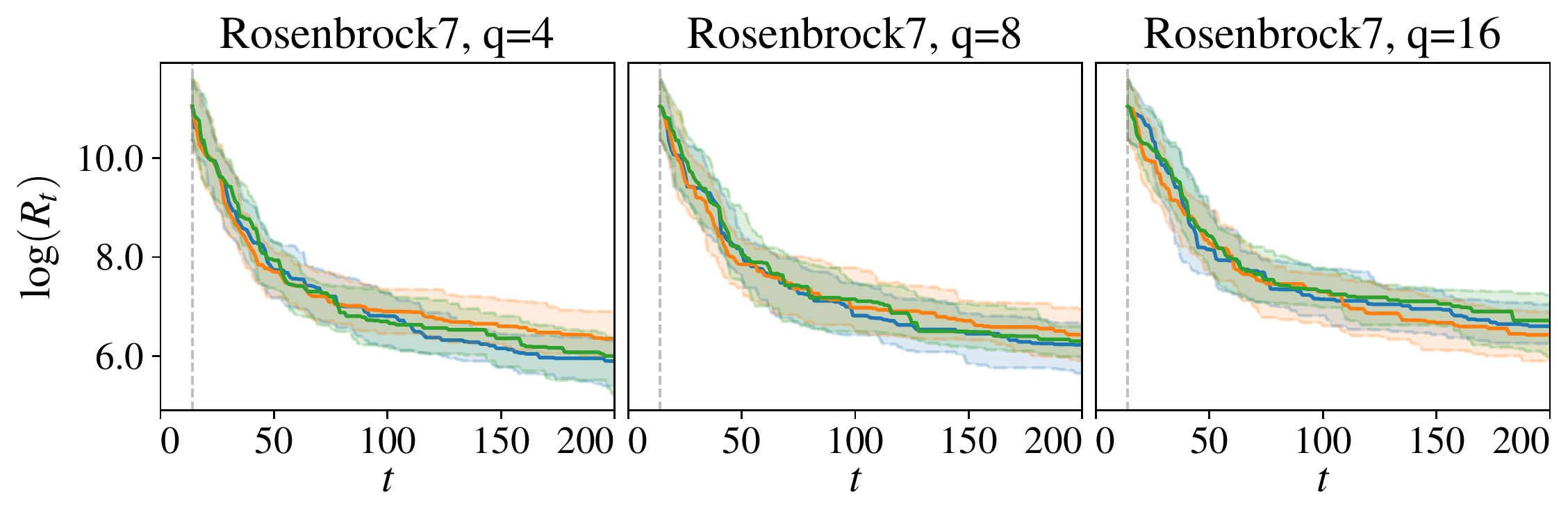}\\
\includegraphics[width=0.5\linewidth, clip, trim={5 0 5 7}]{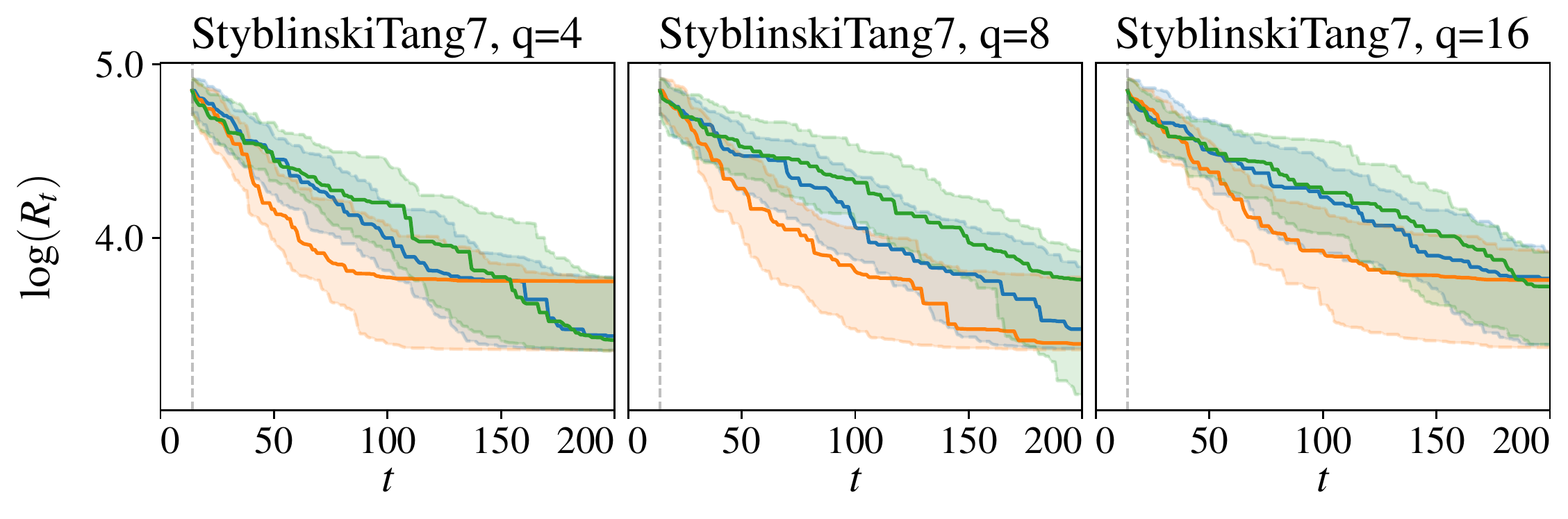}%
\includegraphics[width=0.5\linewidth, clip, trim={5 0 5 7}]{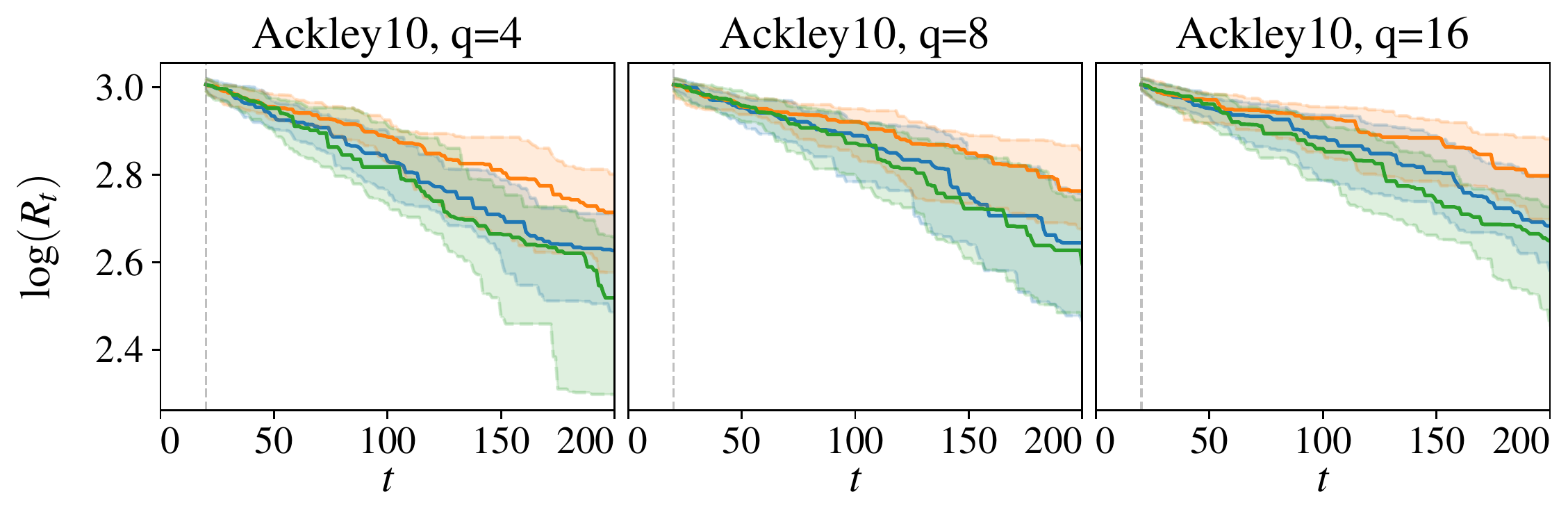}\\
\includegraphics[width=0.5\linewidth, clip, trim={5 0 5 7}]{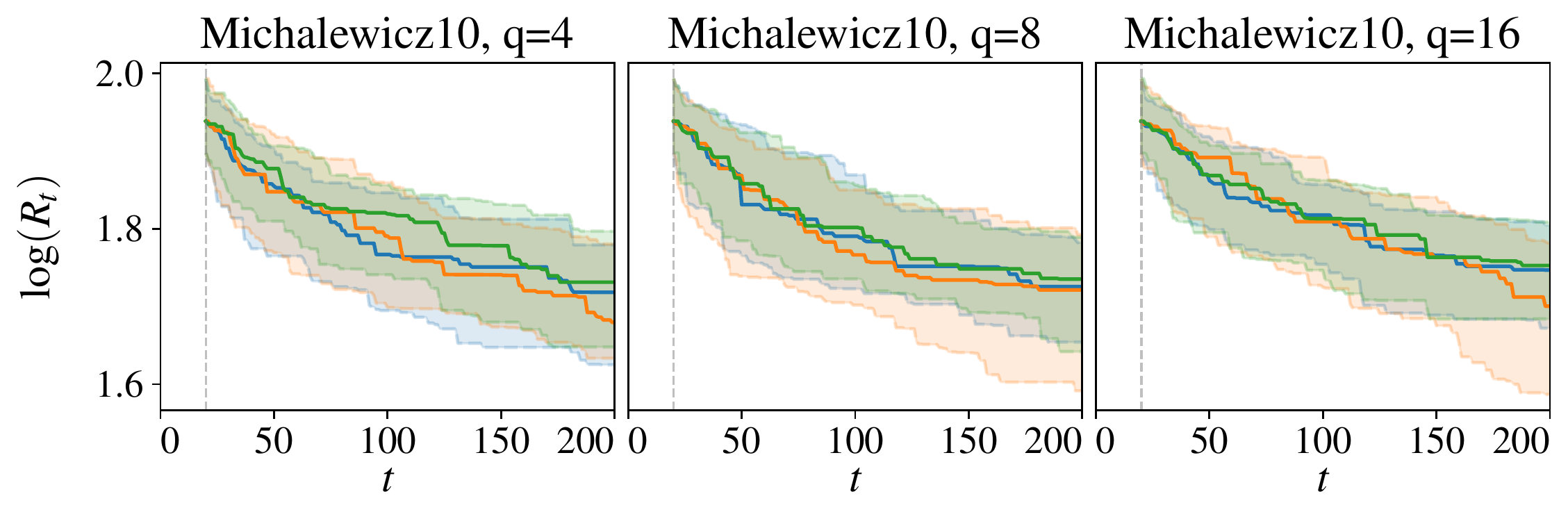}%
\includegraphics[width=0.5\linewidth, clip, trim={5 0 5 7}]{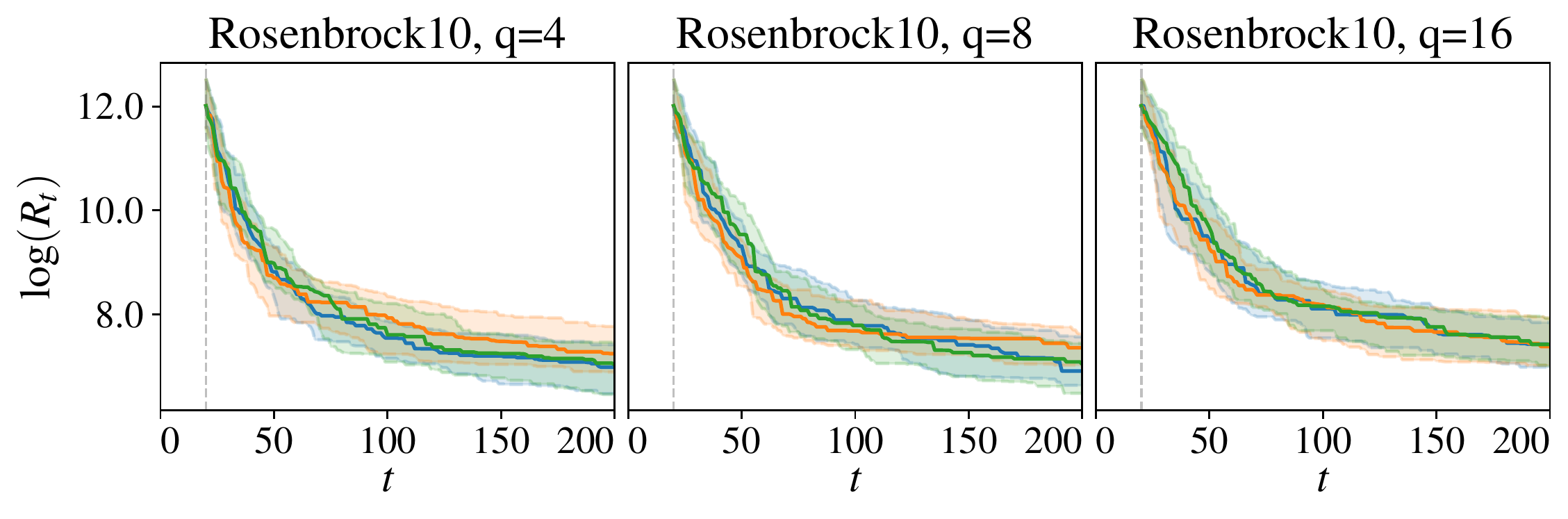}\\
\includegraphics[width=0.5\linewidth, clip, trim={5 0 5 7}]{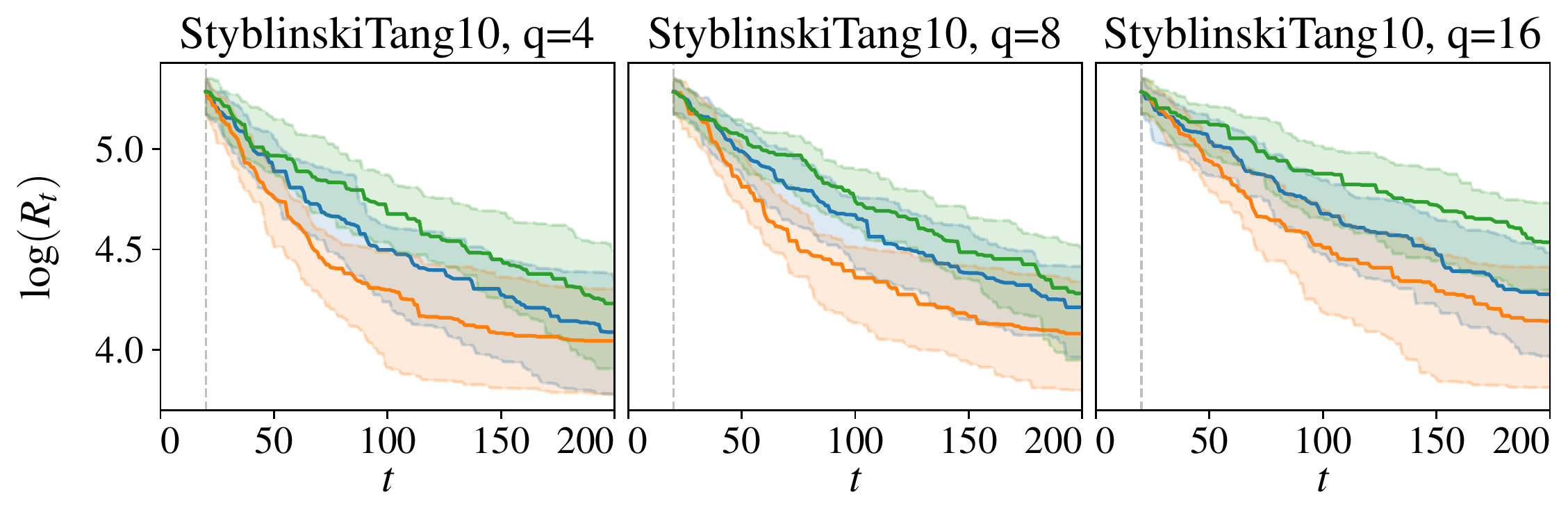}\\
\includegraphics[width=0.4\linewidth, clip, trim={10 15 10 13}]{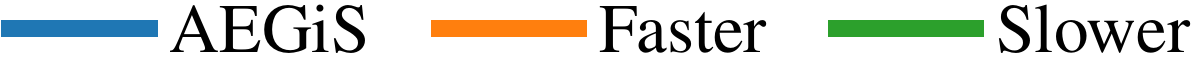}%
\caption{Convergence results for the $\epsilon$ value experiments.}
\label{fig:results:epsbracket}
\end{figure}

In this section we show all convergence plots and results table for choosing
a suitable value of $\epsilon$ ($\epsilon_T = \epsilon_P = \epsilon/2$).
Here, AEGiS corresponds to the value used 
throughout the rest of the paper, $\epsilon = \min(2/\sqrt{d},1)$. 
The labels \emph{faster} and \emph{slower} correspond to using values of 
$\epsilon = \min(2 / (d - 2), 1)$ and $\epsilon = \min(2 / \log(d + 3), 1)$
respectively, where \emph{faster} and \emph{slower} refer to the increased and
decrease rate of decay of $\epsilon$ with respect to the problem dimensionality
$d$. Therefore, \emph{faster} exploits more than AEGiS, and \emph{slower}
explores more than AEGiS for a given problem dimensionality.
Figure~\ref{fig:results:epsdecay} shows the three $\epsilon$ decay curves.
Figure~\ref{fig:results:epsbracket} shows the convergence plots for AEGiS with
the three $\epsilon$ decay rates evaluated on the $15$ synthetic test functions
for $q \in \{4, 8, 16\}$. Tables~\ref{tbl:epsbracket1}, \ref{tbl:epsbracket2},
and~\ref{tbl:epsbracket3} show the median log simple regret as well as the 
median absolute deviation from the median (MAD), a robust measure of 
dispersion.

\begin{table}[H]
\setlength{\tabcolsep}{2pt}
\sisetup{table-format=1.2e-1,table-number-alignment=center}
\caption{Tabulated results for $q = 4$ asynchronous workers, showing the
median log simple regret (\emph{left}) and median absolute deviation from
the median (MAD, \emph{right}) after 200 function evaluations across the 51 runs. 
The method with the lowest median performance is shown in dark grey, with 
those with statistically equivalent performance are shown in light grey.}
\resizebox{1\textwidth}{!}{%
\begin{tabular}{l Sz Sz Sz Sz Sz}
\toprule
    \bfseries Method
    & \multicolumn{2}{c}{\bfseries Branin (2)} 
    & \multicolumn{2}{c}{\bfseries Eggholder (2)} 
    & \multicolumn{2}{c}{\bfseries GoldsteinPrice (2)} 
    & \multicolumn{2}{c}{\bfseries SixHumpCamel (2)} 
    & \multicolumn{2}{c}{\bfseries Hartmann3 (3)} \\ 
    & \multicolumn{1}{c}{Median} & \multicolumn{1}{c}{MAD}
    & \multicolumn{1}{c}{Median} & \multicolumn{1}{c}{MAD}
    & \multicolumn{1}{c}{Median} & \multicolumn{1}{c}{MAD}
    & \multicolumn{1}{c}{Median} & \multicolumn{1}{c}{MAD}
    & \multicolumn{1}{c}{Median} & \multicolumn{1}{c}{MAD}  \\ \midrule
    AEGiS & \statsimilar 5.99e-06 & \statsimilar 6.90e-06 & \statsimilar 6.52e+01 & \statsimilar 1.08e+01 & \statsimilar 6.99e-01 & \statsimilar 7.67e-01 & \statsimilar 2.93e-06 & \statsimilar 3.31e-06 & \statsimilar 5.29e-05 & \statsimilar 5.59e-05 \\
    Faster & \statsimilar 4.81e-06 & \statsimilar 5.56e-06 & \statsimilar 6.51e+01 & \statsimilar 1.10e+01 & \best 2.84e-01 & \best 3.57e-01 & \statsimilar 2.83e-06 & \statsimilar 3.17e-06 & \best 4.76e-05 & \best 5.89e-05 \\
    Slower & \best 3.47e-06 & \best 4.18e-06 & \best 6.51e+01 & \best 1.61e+01 & \statsimilar 3.60e-01 & \statsimilar 4.71e-01 & \best 2.25e-06 & \best 2.87e-06 & \statsimilar 7.01e-05 & \statsimilar 8.80e-05 \\
\bottomrule
\toprule
    \bfseries Method
    & \multicolumn{2}{c}{\bfseries Ackley5 (5)} 
    & \multicolumn{2}{c}{\bfseries Michalewicz5 (5)} 
    & \multicolumn{2}{c}{\bfseries StyblinskiTang5 (5)} 
    & \multicolumn{2}{c}{\bfseries Hartmann6 (6)} 
    & \multicolumn{2}{c}{\bfseries Rosenbrock7 (7)} \\ 
    & \multicolumn{1}{c}{Median} & \multicolumn{1}{c}{MAD}
    & \multicolumn{1}{c}{Median} & \multicolumn{1}{c}{MAD}
    & \multicolumn{1}{c}{Median} & \multicolumn{1}{c}{MAD}
    & \multicolumn{1}{c}{Median} & \multicolumn{1}{c}{MAD}
    & \multicolumn{1}{c}{Median} & \multicolumn{1}{c}{MAD}  \\ \midrule
    AEGiS & \best 2.81e+00 & \best 1.19e+00 & \statsimilar 1.54e+00 & \statsimilar 4.77e-01 & \statsimilar 1.51e+01 & \statsimilar 2.00e+01 & \statsimilar 2.28e-03 & \statsimilar 3.09e-03 & \best 3.64e+02 & \best 2.44e+02 \\
    Faster & 5.34e+00 & 3.53e+00 & \best 1.38e+00 & \best 4.77e-01 & \statsimilar 1.50e+01 & \statsimilar 1.99e+01 & 4.41e-03 & 6.05e-03 & 5.72e+02 & 3.43e+02 \\
    Slower & \statsimilar 3.20e+00 & \statsimilar 6.71e-01 & 1.53e+00 & 6.13e-01 & \best 1.47e+01 & \best 1.32e+01 & \best 2.28e-03 & \best 3.07e-03 & \statsimilar 4.04e+02 & \statsimilar 3.08e+02 \\
\bottomrule
\toprule
    \bfseries Method
    & \multicolumn{2}{c}{\bfseries StyblinskiTang7 (7)} 
    & \multicolumn{2}{c}{\bfseries Ackley10 (10)} 
    & \multicolumn{2}{c}{\bfseries Michalewicz10 (10)} 
    & \multicolumn{2}{c}{\bfseries Rosenbrock10 (10)} 
    & \multicolumn{2}{c}{\bfseries StyblinskiTang10 (10)} \\ 
    & \multicolumn{1}{c}{Median} & \multicolumn{1}{c}{MAD}
    & \multicolumn{1}{c}{Median} & \multicolumn{1}{c}{MAD}
    & \multicolumn{1}{c}{Median} & \multicolumn{1}{c}{MAD}
    & \multicolumn{1}{c}{Median} & \multicolumn{1}{c}{MAD}
    & \multicolumn{1}{c}{Median} & \multicolumn{1}{c}{MAD}  \\ \midrule
    AEGiS & \statsimilar 3.10e+01 & \statsimilar 1.79e+01 & \statsimilar 1.38e+01 & \statsimilar 2.30e+00 & \statsimilar 5.57e+00 & \statsimilar 7.16e-01 & \best 1.08e+03 & \best 7.27e+02 & \statsimilar 5.96e+01 & \statsimilar 2.41e+01 \\
    Faster & \statsimilar 4.25e+01 & \statsimilar 2.04e+01 & 1.51e+01 & 2.52e+00 & \best 5.36e+00 & \best 7.35e-01 & 1.40e+03 & 9.75e+02 & \best 5.70e+01 & \best 2.25e+01 \\
    Slower & \best 3.03e+01 & \best 1.93e+01 & \best 1.24e+01 & \best 3.37e+00 & \statsimilar 5.65e+00 & \statsimilar 6.11e-01 & \statsimilar 1.13e+03 & \statsimilar 8.73e+02 & 6.87e+01 & 3.00e+01 \\
\bottomrule
\end{tabular}
}
\label{tbl:epsbracket1}
\end{table}

\begin{table}[H]
\setlength{\tabcolsep}{2pt}
\sisetup{table-format=1.2e-1,table-number-alignment=center}
\caption{Tabulated results for $q = 8$ asynchronous workers, showing the
median log simple regret (\emph{left}) and median absolute deviation from
the median (MAD, \emph{right}) after 200 function evaluations across the 51 runs. 
The method with the lowest median performance is shown in dark grey, with 
those with statistically equivalent performance are shown in light grey.}
\resizebox{1\textwidth}{!}{%
\begin{tabular}{l Sz Sz Sz Sz Sz}
    \toprule
    \bfseries Method
    & \multicolumn{2}{c}{\bfseries Branin (2)} 
    & \multicolumn{2}{c}{\bfseries Eggholder (2)} 
    & \multicolumn{2}{c}{\bfseries GoldsteinPrice (2)} 
    & \multicolumn{2}{c}{\bfseries SixHumpCamel (2)} 
    & \multicolumn{2}{c}{\bfseries Hartmann3 (3)} \\ 
    & \multicolumn{1}{c}{Median} & \multicolumn{1}{c}{MAD}
    & \multicolumn{1}{c}{Median} & \multicolumn{1}{c}{MAD}
    & \multicolumn{1}{c}{Median} & \multicolumn{1}{c}{MAD}
    & \multicolumn{1}{c}{Median} & \multicolumn{1}{c}{MAD}
    & \multicolumn{1}{c}{Median} & \multicolumn{1}{c}{MAD}  \\ \midrule
    AEGiS & \best 5.32e-06 & \best 6.51e-06 & \statsimilar 6.51e+01 & \statsimilar 1.35e+01 & \best 8.10e-01 & \best 8.82e-01 & \statsimilar 2.98e-06 & \statsimilar 3.83e-06 & \statsimilar 9.99e-05 & \statsimilar 1.03e-04 \\
    Faster & \statsimilar 5.96e-06 & \statsimilar 6.64e-06 & \statsimilar 6.51e+01 & \statsimilar 1.42e+01 & \statsimilar 2.36e+00 & \statsimilar 3.10e+00 & \statsimilar 2.09e-06 & \statsimilar 2.07e-06 & \statsimilar 9.89e-05 & \statsimilar 1.16e-04 \\
    Slower & \statsimilar 7.57e-06 & \statsimilar 1.02e-05 & \best 6.51e+01 & \best 2.30e+01 & \statsimilar 1.16e+00 & \statsimilar 1.62e+00 & \best 2.02e-06 & \best 2.63e-06 & \best 7.94e-05 & \best 9.45e-05 \\
\bottomrule
\toprule
    \bfseries Method
    & \multicolumn{2}{c}{\bfseries Ackley5 (5)} 
    & \multicolumn{2}{c}{\bfseries Michalewicz5 (5)} 
    & \multicolumn{2}{c}{\bfseries StyblinskiTang5 (5)} 
    & \multicolumn{2}{c}{\bfseries Hartmann6 (6)} 
    & \multicolumn{2}{c}{\bfseries Rosenbrock7 (7)} \\ 
    & \multicolumn{1}{c}{Median} & \multicolumn{1}{c}{MAD}
    & \multicolumn{1}{c}{Median} & \multicolumn{1}{c}{MAD}
    & \multicolumn{1}{c}{Median} & \multicolumn{1}{c}{MAD}
    & \multicolumn{1}{c}{Median} & \multicolumn{1}{c}{MAD}
    & \multicolumn{1}{c}{Median} & \multicolumn{1}{c}{MAD}  \\ \midrule
    AEGiS & \statsimilar 3.39e+00 & \statsimilar 8.01e-01 & \statsimilar 1.53e+00 & \statsimilar 5.62e-01 & \statsimilar 1.53e+01 & \statsimilar 1.89e+01 & \best 2.67e-03 & \best 3.16e-03 & \best 5.04e+02 & \best 3.54e+02 \\
    Faster & 6.09e+00 & 2.54e+00 & \best 1.49e+00 & \best 5.13e-01 & \statsimilar 1.53e+01 & \statsimilar 1.94e+01 & 6.73e-03 & 9.57e-03 & \statsimilar 6.20e+02 & \statsimilar 4.57e+02 \\
    Slower & \best 3.37e+00 & \best 6.10e-01 & \statsimilar 1.69e+00 & \statsimilar 5.22e-01 & \best 1.47e+01 & \best 2.03e+01 & \statsimilar 2.91e-03 & \statsimilar 4.06e-03 & \statsimilar 5.45e+02 & \statsimilar 2.68e+02 \\
\bottomrule
\toprule
    \bfseries Method
    & \multicolumn{2}{c}{\bfseries StyblinskiTang7 (7)} 
    & \multicolumn{2}{c}{\bfseries Ackley10 (10)} 
    & \multicolumn{2}{c}{\bfseries Michalewicz10 (10)} 
    & \multicolumn{2}{c}{\bfseries Rosenbrock10 (10)} 
    & \multicolumn{2}{c}{\bfseries StyblinskiTang10 (10)} \\ 
    & \multicolumn{1}{c}{Median} & \multicolumn{1}{c}{MAD}
    & \multicolumn{1}{c}{Median} & \multicolumn{1}{c}{MAD}
    & \multicolumn{1}{c}{Median} & \multicolumn{1}{c}{MAD}
    & \multicolumn{1}{c}{Median} & \multicolumn{1}{c}{MAD}
    & \multicolumn{1}{c}{Median} & \multicolumn{1}{c}{MAD}  \\ \midrule
    AEGiS & \statsimilar 3.23e+01 & \statsimilar 1.69e+01 & \statsimilar 1.41e+01 & \statsimilar 2.83e+00 & \statsimilar 5.61e+00 & \statsimilar 5.60e-01 & \best 9.97e+02 & \best 5.96e+02 & \statsimilar 6.75e+01 & \statsimilar 2.31e+01 \\
    Faster & \best 2.96e+01 & \best 1.94e+01 & 1.58e+01 & 2.13e+00 & \best 5.59e+00 & \best 8.40e-01 & \statsimilar 1.57e+03 & \statsimilar 6.94e+02 & \best 5.92e+01 & \best 2.36e+01 \\
    Slower & \statsimilar 4.29e+01 & \statsimilar 1.85e+01 & \best 1.34e+01 & \best 2.84e+00 & \statsimilar 5.67e+00 & \statsimilar 4.79e-01 & \statsimilar 1.14e+03 & \statsimilar 8.00e+02 & 7.23e+01 & 2.85e+01 \\
\bottomrule
\end{tabular}
}
\label{tbl:epsbracket2}
\end{table}

\begin{table}[H]
\setlength{\tabcolsep}{2pt}
\sisetup{table-format=1.2e-1,table-number-alignment=center}
\caption{Tabulated results for $q = 16$ asynchronous workers, showing the
median log simple regret (\emph{left}) and median absolute deviation from
the median (MAD, \emph{right}) after 200 function evaluations across the 51 runs. 
The method with the lowest median performance is shown in dark grey, with 
those with statistically equivalent performance are shown in light grey.}
\resizebox{1\textwidth}{!}{%
\begin{tabular}{l Sz Sz Sz Sz Sz}
\toprule
    \bfseries Method
    & \multicolumn{2}{c}{\bfseries Branin (2)} 
    & \multicolumn{2}{c}{\bfseries Eggholder (2)} 
    & \multicolumn{2}{c}{\bfseries GoldsteinPrice (2)} 
    & \multicolumn{2}{c}{\bfseries SixHumpCamel (2)} 
    & \multicolumn{2}{c}{\bfseries Hartmann3 (3)} \\ 
    & \multicolumn{1}{c}{Median} & \multicolumn{1}{c}{MAD}
    & \multicolumn{1}{c}{Median} & \multicolumn{1}{c}{MAD}
    & \multicolumn{1}{c}{Median} & \multicolumn{1}{c}{MAD}
    & \multicolumn{1}{c}{Median} & \multicolumn{1}{c}{MAD}
    & \multicolumn{1}{c}{Median} & \multicolumn{1}{c}{MAD}  \\ \midrule
    AEGiS & 1.28e-05 & 1.80e-05 & \best 6.53e+01 & \best 1.27e+01 & \statsimilar 1.56e+00 & \statsimilar 1.84e+00 & \statsimilar 5.82e-06 & \statsimilar 5.32e-06 & 1.93e-04 & 2.80e-04 \\
    Faster & \statsimilar 1.26e-05 & \statsimilar 1.55e-05 & \statsimilar 6.54e+01 & \statsimilar 1.32e+01 & \best 1.36e+00 & \best 1.92e+00 & \best 3.71e-06 & \best 4.40e-06 & \statsimilar 1.61e-04 & \statsimilar 1.92e-04 \\
    Slower & \best 9.08e-06 & \best 9.72e-06 & \statsimilar 6.54e+01 & \statsimilar 1.73e+01 & \statsimilar 2.20e+00 & \statsimilar 3.04e+00 & \statsimilar 4.51e-06 & \statsimilar 5.24e-06 & \best 9.79e-05 & \best 1.17e-04 \\
\bottomrule
\toprule
    \bfseries Method
    & \multicolumn{2}{c}{\bfseries Ackley5 (5)} 
    & \multicolumn{2}{c}{\bfseries Michalewicz5 (5)} 
    & \multicolumn{2}{c}{\bfseries StyblinskiTang5 (5)} 
    & \multicolumn{2}{c}{\bfseries Hartmann6 (6)} 
    & \multicolumn{2}{c}{\bfseries Rosenbrock7 (7)} \\ 
    & \multicolumn{1}{c}{Median} & \multicolumn{1}{c}{MAD}
    & \multicolumn{1}{c}{Median} & \multicolumn{1}{c}{MAD}
    & \multicolumn{1}{c}{Median} & \multicolumn{1}{c}{MAD}
    & \multicolumn{1}{c}{Median} & \multicolumn{1}{c}{MAD}
    & \multicolumn{1}{c}{Median} & \multicolumn{1}{c}{MAD}  \\ \midrule
    AEGiS & \best 3.39e+00 & \best 1.12e+00 & \statsimilar 1.62e+00 & \statsimilar 5.10e-01 & \statsimilar 1.56e+01 & \statsimilar 1.09e+01 & \statsimilar 3.21e-03 & \statsimilar 3.82e-03 & \statsimilar 7.38e+02 & \statsimilar 5.13e+02 \\
    Faster & 6.26e+00 & 3.70e+00 & \best 1.41e+00 & \best 5.64e-01 & \best 1.46e+01 & \best 2.03e+01 & \statsimilar 5.44e-03 & \statsimilar 7.27e-03 & \best 6.18e+02 & \best 5.03e+02 \\
    Slower & \statsimilar 3.51e+00 & \statsimilar 6.31e-01 & \statsimilar 1.60e+00 & \statsimilar 4.95e-01 & \statsimilar 1.49e+01 & \statsimilar 1.74e+01 & \best 2.82e-03 & \best 3.14e-03 & \statsimilar 8.27e+02 & \statsimilar 7.18e+02 \\
\bottomrule
\toprule
    \bfseries Method
    & \multicolumn{2}{c}{\bfseries StyblinskiTang7 (7)} 
    & \multicolumn{2}{c}{\bfseries Ackley10 (10)} 
    & \multicolumn{2}{c}{\bfseries Michalewicz10 (10)} 
    & \multicolumn{2}{c}{\bfseries Rosenbrock10 (10)} 
    & \multicolumn{2}{c}{\bfseries StyblinskiTang10 (10)} \\ 
    & \multicolumn{1}{c}{Median} & \multicolumn{1}{c}{MAD}
    & \multicolumn{1}{c}{Median} & \multicolumn{1}{c}{MAD}
    & \multicolumn{1}{c}{Median} & \multicolumn{1}{c}{MAD}
    & \multicolumn{1}{c}{Median} & \multicolumn{1}{c}{MAD}
    & \multicolumn{1}{c}{Median} & \multicolumn{1}{c}{MAD}  \\ \midrule
    AEGiS & \statsimilar 4.32e+01 & \statsimilar 1.80e+01 & \statsimilar 1.46e+01 & \statsimilar 2.22e+00 & \statsimilar 5.74e+00 & \statsimilar 5.74e-01 & \best 1.60e+03 & \best 1.01e+03 & 7.20e+01 & 2.83e+01 \\
    Faster & \statsimilar 4.28e+01 & \statsimilar 2.04e+01 & 1.64e+01 & 2.21e+00 & \best 5.48e+00 & \best 7.44e-01 & \statsimilar 1.61e+03 & \statsimilar 1.12e+03 & \best 6.30e+01 & \best 2.76e+01 \\
    Slower & \best 4.12e+01 & \best 1.62e+01 & \best 1.41e+01 & \best 2.44e+00 & \statsimilar 5.77e+00 & \statsimilar 5.26e-01 & \statsimilar 1.67e+03 & \statsimilar 1.11e+03 & 9.33e+01 & 2.98e+01 \\
\bottomrule
\end{tabular}
}
\label{tbl:epsbracket3}
\end{table}

\section{Sequential Bayesian Optimisation (\texorpdfstring{$q=1$}{q = 1})}
\label{sec:sequential}
In this section we evaluate AEGiS in the sequential BO setting, \ie with $q =
1$ workers and compare it to the other asynchronous methods in 
Section~\ref{q1:eval}, carry out an ablation study in Section~\ref{q1:ablation}
and investigate faster and slower $\epsilon$ decay rates in
Section~\ref{q1:eps}.

\subsection{Evaluation}
\label{q1:eval}
\begin{figure}[H]
\centering
\includegraphics[width=1.\linewidth, clip, trim={5 0 5 7}]{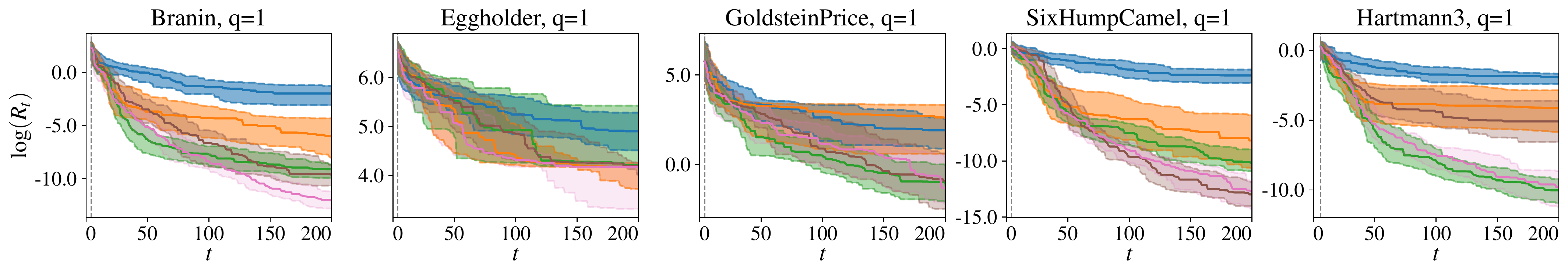}\\
\includegraphics[width=1.\linewidth, clip, trim={5 0 5 7}]{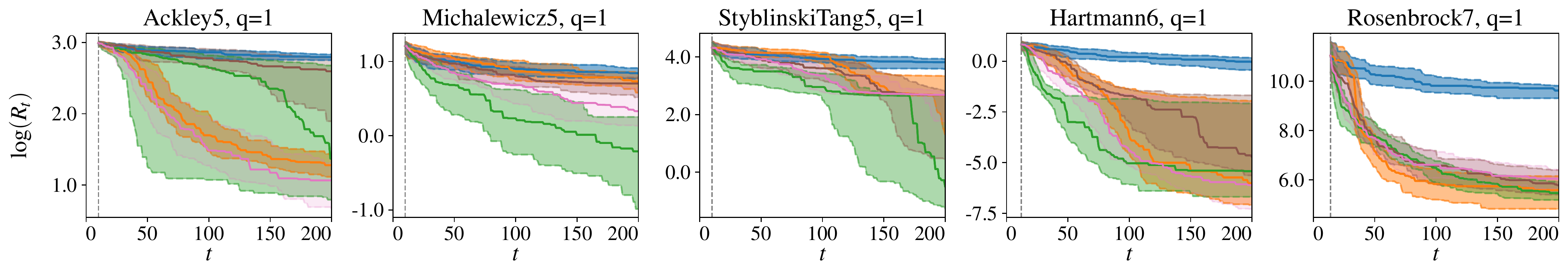}\\
\includegraphics[width=1.\linewidth, clip, trim={5 0 5 7}]{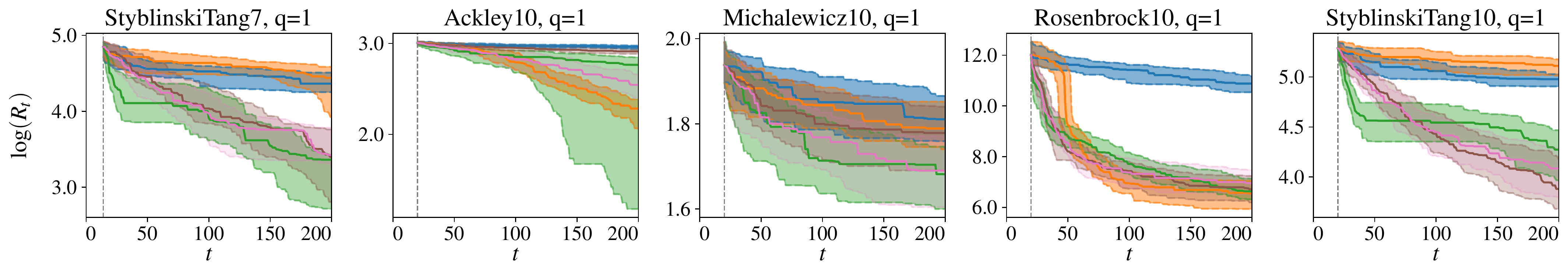}\\
\includegraphics[width=0.6\linewidth, clip, trim={10 15 10 13}]{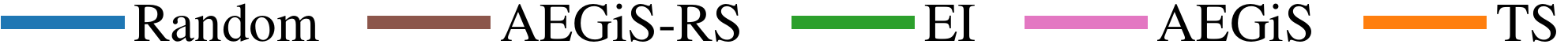}%
\caption{Convergence results for the sequential BO experiments.}
\label{fig:results:sequential:synth}
\end{figure}

\begin{figure}[H]
\centering
\includegraphics[width=0.8\linewidth, clip, trim={5 0 5 7}]{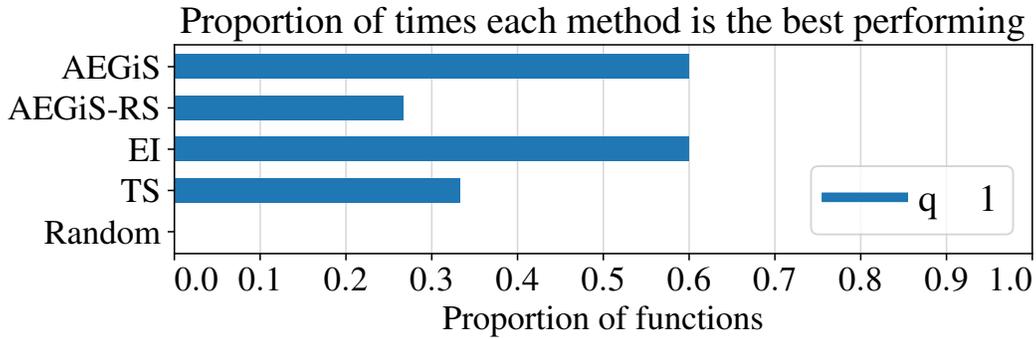}%
\caption{Synthetic function optimisation summary. Bar lengths correspond to the
proportion of times that a method is best or statistically equivalent to the 
best method across the 15 synthetic functions.}
\label{fig:results:sequential:equaltobest}
\end{figure}

We first compare compare AEGiS to the other asynchronous methods in the
sequential setting. Note that KB, LP and PLAyBOOK are equivalent because each
select the first location to be evaluated using EI. Therefore, we compare AEGiS
and AEGiS-RS to EI, TS and Latin hypercube sampling (\ie Random) on the $15$
synthetic benchmark functions. Figure~\ref{fig:results:sequential:synth} shows
the convergence plots for the benchmark functions, with solid lines showing the
median log simple regret and shading showing the interquartile range. 
Table~\ref{tbl:q1:synthetic}, show the median log simple regret as well as the
median absolute deviation from the median (MAD), a robust measure of 
dispersion. The method with the best (lowest) median regret is shown in dark 
grey, and those that are statistically equivalent to the best method according
to a one-sided, paired Wilcoxon signed-rank test with Holm-Bonferroni
correction \citep{holm:test:1979} ($p \geq 0.05$) are shown in light grey. 
Figure~\ref{fig:results:sequential:equaltobest} summarises the tabulated
results and shows the number of times each method is best or statistically
equal to the best performing method.

\begin{table}[H]
\setlength{\tabcolsep}{2pt}
\sisetup{table-format=1.2e-1,table-number-alignment=center}
\caption{Tabulated results for the sequential ($q=1$) BO optimisation runs, 
showing the median log simple regret (\emph{left}) and median absolute
deviation from the median (MAD, \emph{right}) after 200 function evaluations
across the 51 runs. The method with the lowest median performance is shown in
dark grey, with those with statistically equivalent performance are shown in
light grey.}
\resizebox{1\textwidth}{!}{%
\begin{tabular}{l Sz Sz Sz Sz Sz}
    \toprule
    \bfseries Method
    & \multicolumn{2}{c}{\bfseries Branin (2)} 
    & \multicolumn{2}{c}{\bfseries Eggholder (2)} 
    & \multicolumn{2}{c}{\bfseries GoldsteinPrice (2)} 
    & \multicolumn{2}{c}{\bfseries SixHumpCamel (2)} 
    & \multicolumn{2}{c}{\bfseries Hartmann3 (3)} \\ 
    & \multicolumn{1}{c}{Median} & \multicolumn{1}{c}{MAD}
    & \multicolumn{1}{c}{Median} & \multicolumn{1}{c}{MAD}
    & \multicolumn{1}{c}{Median} & \multicolumn{1}{c}{MAD}
    & \multicolumn{1}{c}{Median} & \multicolumn{1}{c}{MAD}
    & \multicolumn{1}{c}{Median} & \multicolumn{1}{c}{MAD}  \\ \midrule
    Random & 1.37e-01 & 1.41e-01 & 1.35e+02 & 7.08e+01 & 6.75e+00 & 7.57e+00 & 9.30e-02 & 7.41e-02 & 1.46e-01 & 7.08e-02 \\
    TS & 2.51e-03 & 3.58e-03 & \statsimilar 6.51e+01 & \statsimilar 1.13e+01 & 1.32e+01 & 1.91e+01 & 2.81e-04 & 4.14e-04 & 1.59e-02 & 2.15e-02 \\
    EI & 1.11e-04 & 9.39e-05 & 6.71e+01 & 9.25e+00 & \statsimilar 3.46e-01 & \statsimilar 4.25e-01 & 3.87e-05 & 4.74e-05 & \best 4.42e-05 & \best 4.74e-05 \\
    AEGiS-RS & 6.76e-05 & 9.41e-05 & 6.58e+01 & 7.47e+00 & \statsimilar 3.96e-01 & \statsimilar 5.23e-01 & \best 2.30e-06 & \best 2.64e-06 & 6.16e-03 & 8.59e-03 \\
    AEGiS & \best 5.69e-06 & \best 4.87e-06 & \best 6.51e+01 & \best 2.09e+01 & \best 2.71e-01 & \best 3.67e-01 & \statsimilar 3.08e-06 & \statsimilar 3.94e-06 & \statsimilar 5.62e-05 & \statsimilar 6.94e-05 \\
\bottomrule
\toprule
    \bfseries Method
    & \multicolumn{2}{c}{\bfseries Ackley5 (5)} 
    & \multicolumn{2}{c}{\bfseries Michalewicz5 (5)} 
    & \multicolumn{2}{c}{\bfseries StyblinskiTang5 (5)} 
    & \multicolumn{2}{c}{\bfseries Hartmann6 (6)} 
    & \multicolumn{2}{c}{\bfseries Rosenbrock7 (7)} \\ 
    & \multicolumn{1}{c}{Median} & \multicolumn{1}{c}{MAD}
    & \multicolumn{1}{c}{Median} & \multicolumn{1}{c}{MAD}
    & \multicolumn{1}{c}{Median} & \multicolumn{1}{c}{MAD}
    & \multicolumn{1}{c}{Median} & \multicolumn{1}{c}{MAD}
    & \multicolumn{1}{c}{Median} & \multicolumn{1}{c}{MAD}  \\ \midrule
    Random & 1.64e+01 & 1.01e+00 & 2.32e+00 & 2.49e-01 & 4.56e+01 & 1.32e+01 & 9.51e-01 & 4.03e-01 & 1.51e+04 & 4.90e+03 \\
    TS & 3.59e+00 & 8.21e-01 & 2.07e+00 & 4.32e-01 & 1.45e+01 & 1.94e+01 & \statsimilar 2.44e-03 & \statsimilar 2.89e-03 & \statsimilar 2.68e+02 & \statsimilar 2.35e+02 \\
    EI & 3.94e+00 & 4.60e+00 & \best 8.09e-01 & \best 6.85e-01 & \best 6.13e-01 & \best 7.75e-01 & \statsimilar 4.40e-03 & \statsimilar 5.64e-03 & \best 2.33e+02 & \best 1.19e+02 \\
    AEGiS-RS & 1.34e+01 & 6.18e+00 & 2.02e+00 & 3.77e-01 & 1.46e+01 & 1.09e+01 & 8.86e-03 & 1.16e-02 & 3.24e+02 & 2.10e+02 \\
    AEGiS & \best 2.90e+00 & \best 1.44e+00 & 1.39e+00 & 5.38e-01 & 1.46e+01 & 2.95e+00 & \best 2.21e-03 & \best 2.77e-03 & 4.20e+02 & 2.33e+02 \\
\bottomrule
\toprule
    \bfseries Method
    & \multicolumn{2}{c}{\bfseries StyblinskiTang7 (7)} 
    & \multicolumn{2}{c}{\bfseries Ackley10 (10)} 
    & \multicolumn{2}{c}{\bfseries Michalewicz10 (10)} 
    & \multicolumn{2}{c}{\bfseries Rosenbrock10 (10)} 
    & \multicolumn{2}{c}{\bfseries StyblinskiTang10 (10)} \\ 
    & \multicolumn{1}{c}{Median} & \multicolumn{1}{c}{MAD}
    & \multicolumn{1}{c}{Median} & \multicolumn{1}{c}{MAD}
    & \multicolumn{1}{c}{Median} & \multicolumn{1}{c}{MAD}
    & \multicolumn{1}{c}{Median} & \multicolumn{1}{c}{MAD}
    & \multicolumn{1}{c}{Median} & \multicolumn{1}{c}{MAD}  \\ \midrule
    Random & 7.85e+01 & 1.81e+01 & 1.93e+01 & 4.99e-01 & 6.11e+00 & 4.75e-01 & 5.32e+04 & 2.30e+04 & 1.45e+02 & 1.46e+01 \\
    TS & 8.48e+01 & 2.92e+01 & \best 9.82e+00 & \best 2.18e+00 & 5.98e+00 & 5.30e-01 & \best 6.98e+02 & \best 5.45e+02 & 1.66e+02 & 2.04e+01 \\
    EI & \best 2.89e+01 & \best 1.74e+01 & 1.59e+01 & 4.32e+00 & \best 5.37e+00 & \best 5.65e-01 & \statsimilar 7.59e+02 & \statsimilar 3.91e+02 & 7.18e+01 & 2.66e+01 \\
    AEGiS-RS & 3.00e+01 & 1.93e+01 & 1.84e+01 & 6.36e-01 & 5.92e+00 & 6.22e-01 & \statsimilar 8.23e+02 & \statsimilar 5.14e+02 & \best 4.85e+01 & \best 1.96e+01 \\
    AEGiS & 3.07e+01 & 1.89e+01 & 1.27e+01 & 2.38e+00 & \statsimilar 5.42e+00 & \statsimilar 6.94e-01 & 1.03e+03 & 7.26e+02 & \statsimilar 5.95e+01 & \statsimilar 2.10e+01 \\
\bottomrule
    \end{tabular}
}
\label{tbl:q1:synthetic}
\end{table}

\subsection{Ablation Study}
\label{q1:ablation}
\begin{figure}[H]
\centering
\includegraphics[width=1.\linewidth, clip, trim={5 0 5 7}]{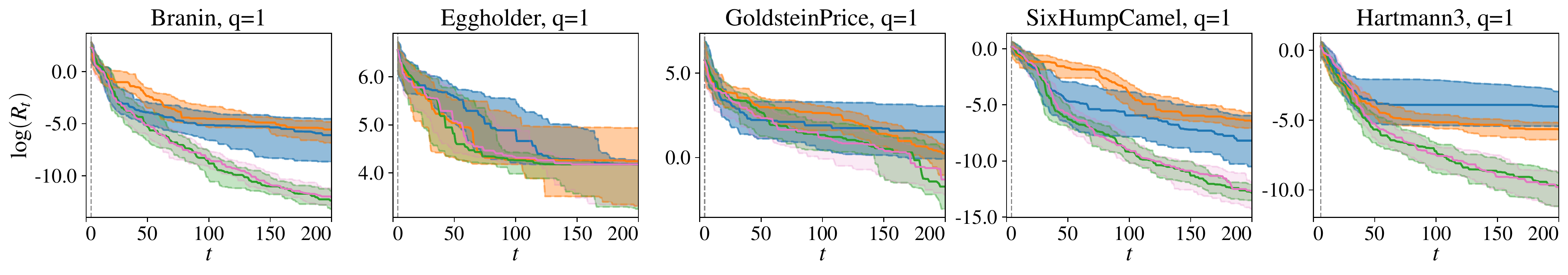}\\
\includegraphics[width=1.\linewidth, clip, trim={5 0 5 7}]{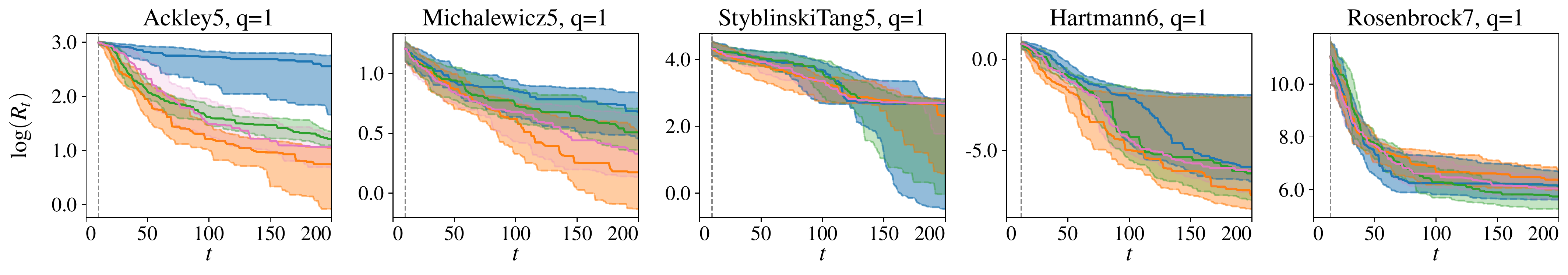}\\
\includegraphics[width=1.\linewidth, clip, trim={5 0 5 7}]{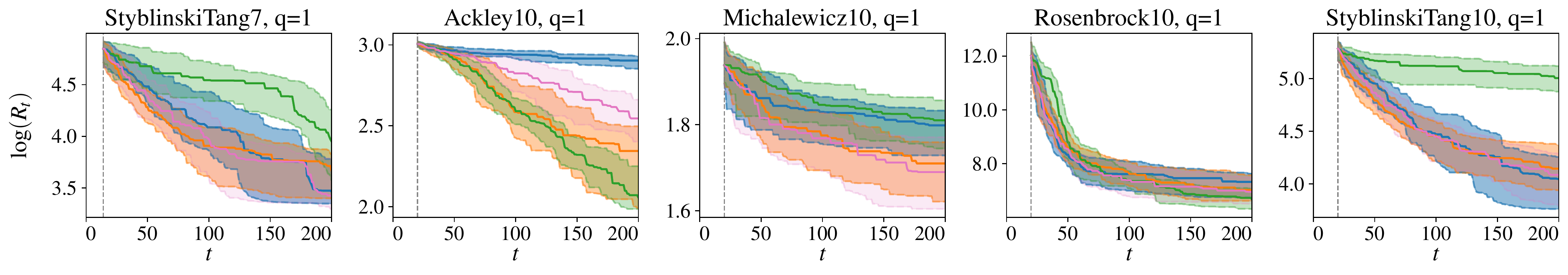}\\
\includegraphics[width=0.6\linewidth, clip, trim={10 15 10 13}]{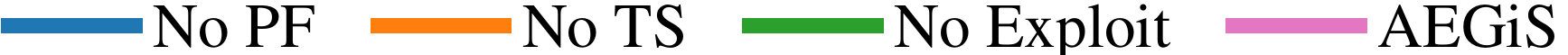}%
\caption{Convergence results for the ablation study experiments.}
\label{fig:results:sequential:ablation}
\end{figure}

\begin{figure}[H]
\centering
\includegraphics[width=0.8\linewidth, clip, trim={5 0 5 7}]{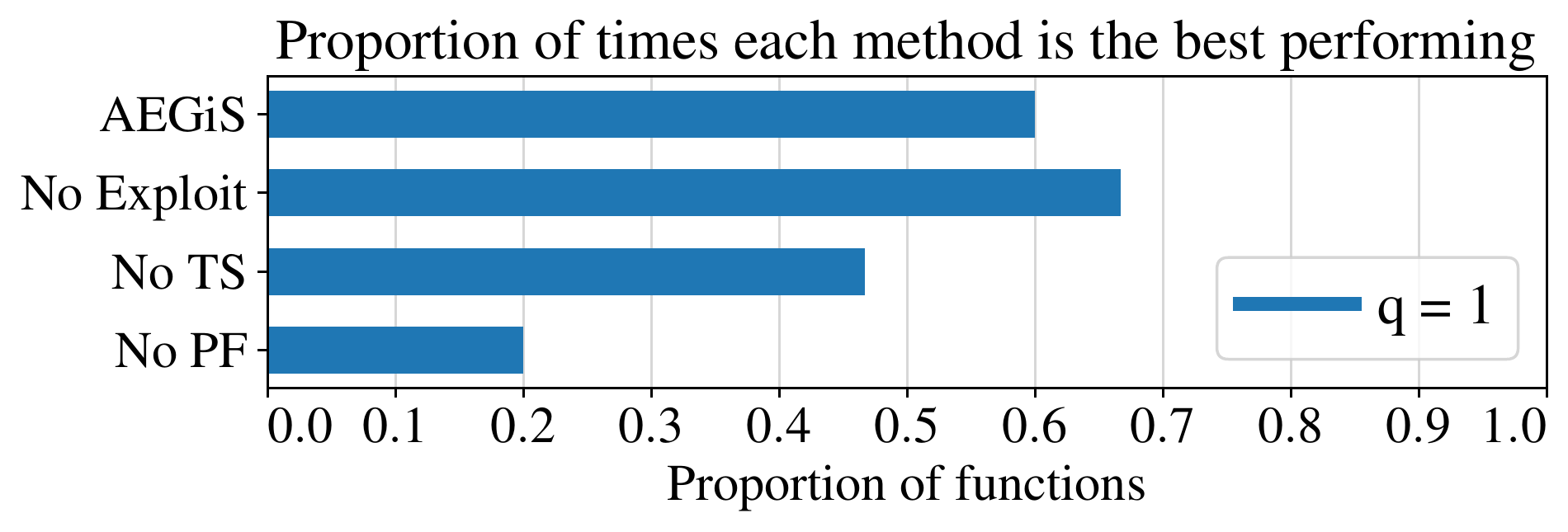}%
\caption{Ablation study summary. Bar lengths correspond to the proportion of
times that a method is best or statistically equivalent to the best method 
across the 15 synthetic functions.}
\label{fig:results:ablation:equaltobest}
\end{figure}

Next, we perform an ablation study as before in the main paper.
Figure~\ref{fig:results:sequential:ablation} shows
the convergence plots for the benchmark functions, with solid lines showing the
median log simple regret and shading showing the interquartile range. 
Table~\ref{tbl:q1:ablation}, show the median log simple regret as well as the
median absolute deviation from the median (MAD), a robust measure of 
dispersion. The method with the best (lowest) median regret is shown in dark 
grey, and those that are statistically equivalent to the best method according
to a one-sided, paired Wilcoxon signed-rank test with Holm-Bonferroni
correction \citep{holm:test:1979} ($p \geq 0.05$) are shown in light grey. 
Figure~\ref{fig:results:ablation:equaltobest} summarises the tabulated
results and shows the number of times each method is best or statistically
equal to the best performing method.

\begin{table}[t]
\setlength{\tabcolsep}{2pt}
\sisetup{table-format=1.2e-1,table-number-alignment=center}
\caption{Tabulated results for the ablation study of the sequential ($q=1$) BO
optimisation runs, showing the median log simple regret (\emph{left}) and
median absolute deviation from the median (MAD, \emph{right}) after 200 
function evaluations across the 51 runs. The method with the lowest median 
performance is shown in dark grey, with those with statistically equivalent
performance are shown in light grey.}
\resizebox{1\textwidth}{!}{%
\begin{tabular}{l Sz Sz Sz Sz Sz}
    \toprule
    \bfseries Method
    & \multicolumn{2}{c}{\bfseries Branin (2)} 
    & \multicolumn{2}{c}{\bfseries Eggholder (2)} 
    & \multicolumn{2}{c}{\bfseries GoldsteinPrice (2)} 
    & \multicolumn{2}{c}{\bfseries SixHumpCamel (2)} 
    & \multicolumn{2}{c}{\bfseries Hartmann3 (3)} \\ 
    & \multicolumn{1}{c}{Median} & \multicolumn{1}{c}{MAD}
    & \multicolumn{1}{c}{Median} & \multicolumn{1}{c}{MAD}
    & \multicolumn{1}{c}{Median} & \multicolumn{1}{c}{MAD}
    & \multicolumn{1}{c}{Median} & \multicolumn{1}{c}{MAD}
    & \multicolumn{1}{c}{Median} & \multicolumn{1}{c}{MAD}  \\ \midrule
    No PF & 2.24e-03 & 3.28e-03 & 6.51e+01 & 8.41e+00 & 4.54e+00 & 6.51e+00 & 2.73e-04 & 4.01e-04 & 1.75e-02 & 2.20e-02 \\
    No TS & 3.85e-03 & 4.64e-03 & 6.96e+01 & 8.70e+01 & \statsimilar 1.28e+00 & \statsimilar 1.36e+00 & 1.47e-03 & 1.41e-03 & 3.43e-03 & 2.87e-03 \\
    No Exploit & \best 4.20e-06 & \best 4.25e-06 & \best 6.51e+01 & \best 1.32e+01 & \best 1.79e-01 & \best 2.50e-01 & \best 2.59e-06 & \best 2.57e-06 & \statsimilar 6.05e-05 & \statsimilar 7.30e-05 \\
    AEGiS & \statsimilar 5.69e-06 & \statsimilar 4.87e-06 & \statsimilar 6.51e+01 & \statsimilar 2.09e+01 & \statsimilar 2.71e-01 & \statsimilar 3.67e-01 & \statsimilar 3.08e-06 & \statsimilar 3.94e-06 & \best 5.62e-05 & \best 6.94e-05 \\
\bottomrule
\toprule
    \bfseries Method
    & \multicolumn{2}{c}{\bfseries Ackley5 (5)} 
    & \multicolumn{2}{c}{\bfseries Michalewicz5 (5)} 
    & \multicolumn{2}{c}{\bfseries StyblinskiTang5 (5)} 
    & \multicolumn{2}{c}{\bfseries Hartmann6 (6)} 
    & \multicolumn{2}{c}{\bfseries Rosenbrock7 (7)} \\ 
    & \multicolumn{1}{c}{Median} & \multicolumn{1}{c}{MAD}
    & \multicolumn{1}{c}{Median} & \multicolumn{1}{c}{MAD}
    & \multicolumn{1}{c}{Median} & \multicolumn{1}{c}{MAD}
    & \multicolumn{1}{c}{Median} & \multicolumn{1}{c}{MAD}
    & \multicolumn{1}{c}{Median} & \multicolumn{1}{c}{MAD}  \\ \midrule
    No PF & 1.29e+01 & 5.68e+00 & 1.98e+00 & 5.14e-01 & \statsimilar 1.43e+01 & \statsimilar 2.00e+01 & 2.78e-03 & 3.38e-03 & 4.73e+02 & 3.39e+02 \\
    No TS & \best 2.09e+00 & \best 1.24e+00 & \best 1.19e+00 & \best 5.80e-01 & \best 1.02e+01 & \best 1.05e+01 & \best 5.86e-04 & \best 6.70e-04 & 5.93e+02 & 3.03e+02 \\
    No Exploit & 3.35e+00 & 7.60e-01 & 1.66e+00 & 4.23e-01 & \statsimilar 1.46e+01 & \statsimilar 1.03e+01 & \statsimilar 1.93e-03 & \statsimilar 2.66e-03 & \best 3.12e+02 & \best 1.88e+02 \\
    AEGiS & 2.90e+00 & 1.44e+00 & 1.39e+00 & 5.38e-01 & \statsimilar 1.46e+01 & \statsimilar 2.95e+00 & 2.21e-03 & 2.77e-03 & \statsimilar 4.20e+02 & \statsimilar 2.33e+02 \\
\bottomrule
\toprule
    \bfseries Method
    & \multicolumn{2}{c}{\bfseries StyblinskiTang7 (7)} 
    & \multicolumn{2}{c}{\bfseries Ackley10 (10)} 
    & \multicolumn{2}{c}{\bfseries Michalewicz10 (10)} 
    & \multicolumn{2}{c}{\bfseries Rosenbrock10 (10)} 
    & \multicolumn{2}{c}{\bfseries StyblinskiTang10 (10)} \\ 
    & \multicolumn{1}{c}{Median} & \multicolumn{1}{c}{MAD}
    & \multicolumn{1}{c}{Median} & \multicolumn{1}{c}{MAD}
    & \multicolumn{1}{c}{Median} & \multicolumn{1}{c}{MAD}
    & \multicolumn{1}{c}{Median} & \multicolumn{1}{c}{MAD}
    & \multicolumn{1}{c}{Median} & \multicolumn{1}{c}{MAD}  \\ \midrule
    No PF & \statsimilar 3.20e+01 & \statsimilar 1.67e+01 & 1.82e+01 & 1.00e+00 & 6.04e+00 & 4.16e-01 & 1.51e+03 & 9.78e+02 & \best 5.73e+01 & \best 2.04e+01 \\
    No TS & \statsimilar 3.92e+01 & \statsimilar 1.32e+01 & 1.04e+01 & 3.56e+00 & \statsimilar 5.53e+00 & \statsimilar 5.38e-01 & 1.12e+03 & 7.36e+02 & 6.31e+01 & 2.16e+01 \\
    No Exploit & 5.19e+01 & 2.51e+01 & \best 7.82e+00 & \best 1.52e+00 & 6.11e+00 & 5.01e-01 & \best 7.89e+02 & \best 4.23e+02 & 1.49e+02 & 2.73e+01 \\
    AEGiS & \best 3.07e+01 & \best 1.89e+01 & 1.27e+01 & 2.38e+00 & \best 5.42e+00 & \best 6.94e-01 & 1.03e+03 & 7.26e+02 & 5.95e+01 & 2.10e+01 \\
\bottomrule
    \end{tabular}
}
\label{tbl:q1:ablation}
\end{table}

\subsection{Setting \eps}
\label{q1:eps}
\begin{figure}[H]
\centering
\includegraphics[width=1.\linewidth, clip, trim={5 0 5 7}]{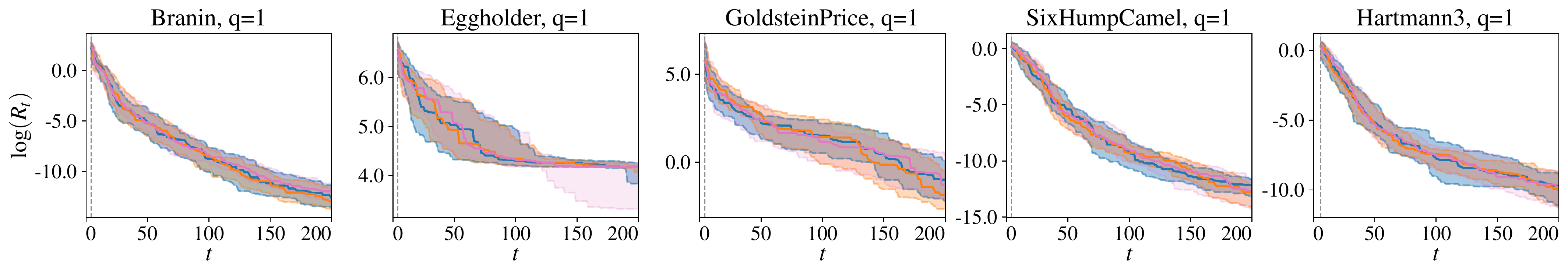}\\
\includegraphics[width=1.\linewidth, clip, trim={5 0 5 7}]{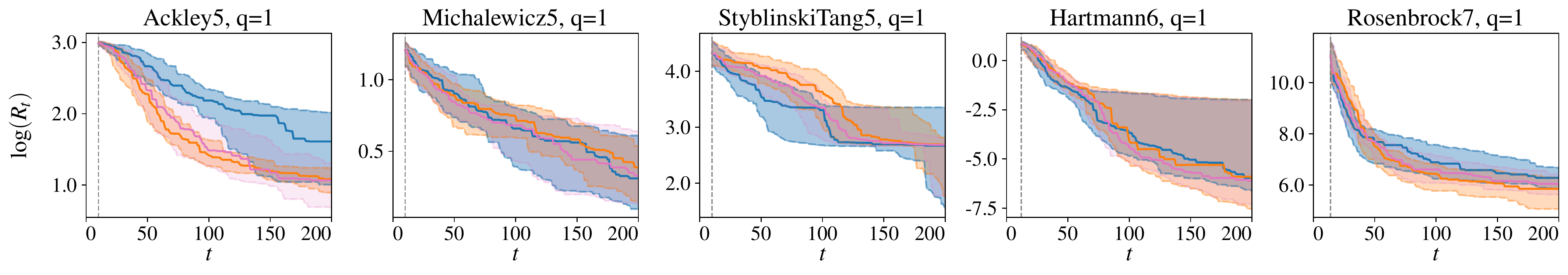}\\
\includegraphics[width=1.\linewidth, clip, trim={5 0 5 7}]{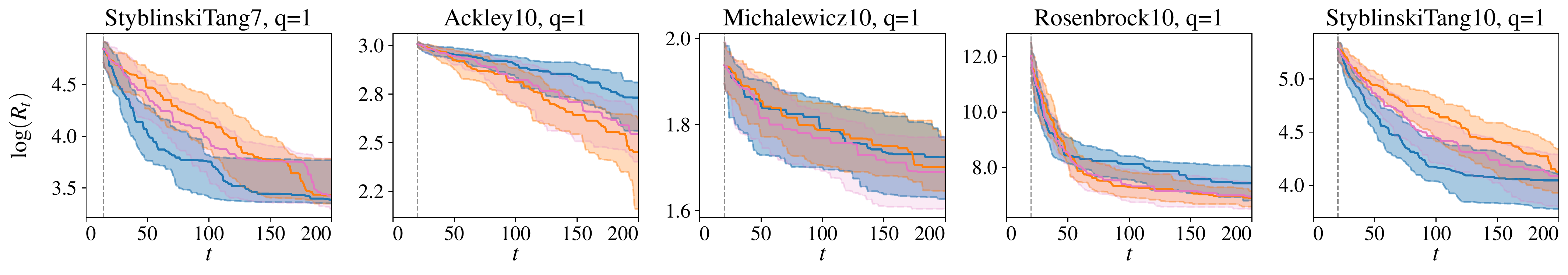}\\
\includegraphics[width=0.5\linewidth, clip, trim={10 15 10 13}]{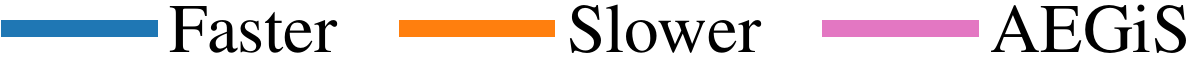}%
\caption{Convergence results for the setting $\epsilon$ experiments.}
\label{fig:results:sequential:epsbracket}
\end{figure}

\begin{figure}[H]
\centering
\includegraphics[width=0.8\linewidth, clip, trim={5 0 5 7}]{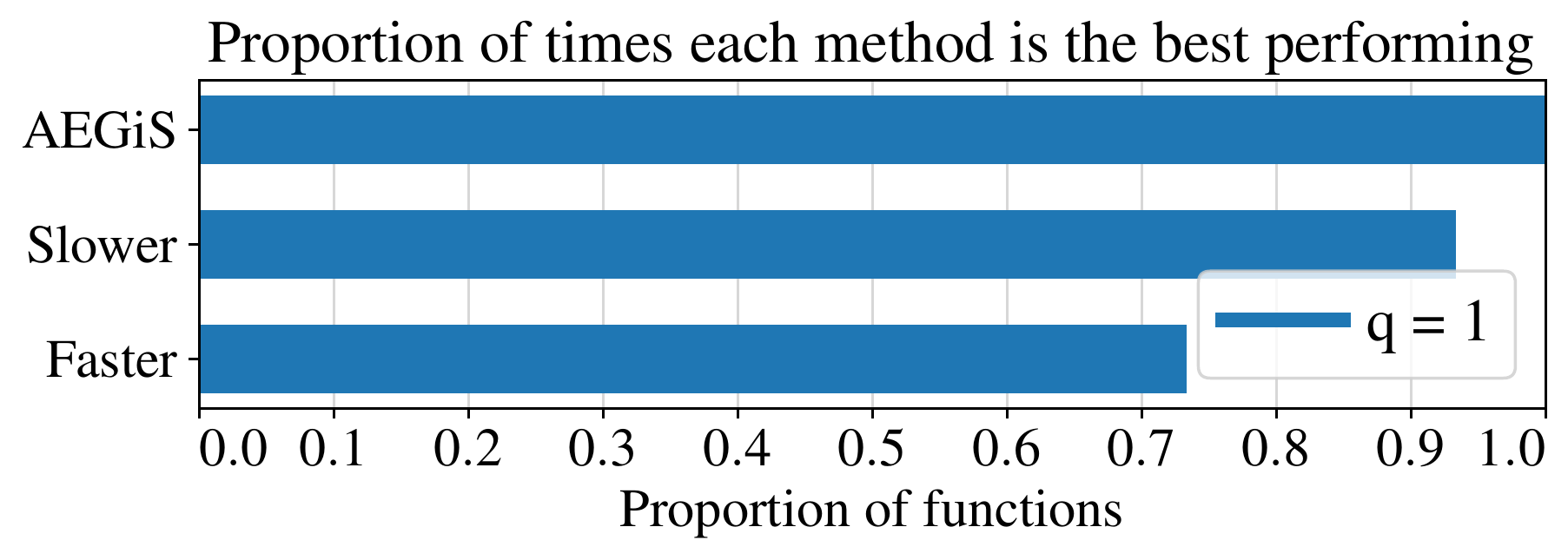}%
\caption{Setting $\epsilon$ summary. Bar lengths correspond to the proportion
of times that a method is best or statistically equivalent to the best method
across the 15 synthetic functions.}
\label{fig:results:epsbracket:equaltobest}
\end{figure}

Lastly, we compare different rates of $\epsilon$ decay ($\epsilon_T =
\epsilon_P = \epsilon/2$). Here, \emph{faster}
corresponds to a quicker rate of decay and thus an increase in exploitation,
and \emph{slower} corresponds to a reduced rate, leading to less exploitation.
Figure~\ref{fig:results:sequential:epsbracket} shows
the convergence plots for the benchmark functions, with solid lines showing the
median log simple regret and shading showing the interquartile range. 
Table~\ref{tbl:q1:epsbracket}, show the median log simple regret as well as the
median absolute deviation from the median (MAD), a robust measure of 
dispersion. The method with the best (lowest) median regret is shown in dark 
grey, and those that are statistically equivalent to the best method according
to a one-sided, paired Wilcoxon signed-rank test with Holm-Bonferroni
correction \citep{holm:test:1979} ($p \geq 0.05$) are shown in light grey. 
Figure~\ref{fig:results:epsbracket:equaltobest} summarises the tabulated
results and shows the number of times each method is best or statistically
equal to the best performing method.

\begin{table}[H]
\setlength{\tabcolsep}{2pt}
\sisetup{table-format=1.2e-1,table-number-alignment=center}
\caption{Tabulated results for the setting $\epsilon$ experiments in the
sequential setting. The table shows the median log simple regret (\emph{left})
and median absolute deviation from the median (MAD, \emph{right}) after 200
function evaluations across the 51 runs. The method with the lowest median
performance is shown in dark grey, with those with statistically equivalent
performance are shown in light grey.}
\resizebox{1\textwidth}{!}{%
\begin{tabular}{l Sz Sz Sz Sz Sz}
    \toprule
    \bfseries Method
    & \multicolumn{2}{c}{\bfseries Branin (2)} 
    & \multicolumn{2}{c}{\bfseries Eggholder (2)} 
    & \multicolumn{2}{c}{\bfseries GoldsteinPrice (2)} 
    & \multicolumn{2}{c}{\bfseries SixHumpCamel (2)} 
    & \multicolumn{2}{c}{\bfseries Hartmann3 (3)} \\ 
    & \multicolumn{1}{c}{Median} & \multicolumn{1}{c}{MAD}
    & \multicolumn{1}{c}{Median} & \multicolumn{1}{c}{MAD}
    & \multicolumn{1}{c}{Median} & \multicolumn{1}{c}{MAD}
    & \multicolumn{1}{c}{Median} & \multicolumn{1}{c}{MAD}
    & \multicolumn{1}{c}{Median} & \multicolumn{1}{c}{MAD}  \\ \midrule
    Faster & \statsimilar 3.16e-06 & \statsimilar 4.21e-06 & \statsimilar 6.51e+01 & \statsimilar 8.39e+00 & \statsimilar 3.68e-01 & \statsimilar 4.81e-01 & \statsimilar 5.28e-06 & \statsimilar 5.65e-06 & \statsimilar 6.15e-05 & \statsimilar 7.50e-05 \\
    Slower & \best 2.23e-06 & \best 2.69e-06 & \statsimilar 6.51e+01 & \statsimilar 4.03e+00 & \best 1.55e-01 & \best 1.84e-01 & \best 2.75e-06 & \best 3.13e-06 & \best 4.44e-05 & \best 6.19e-05 \\
    AEGiS & \statsimilar 5.69e-06 & \statsimilar 4.87e-06 & \best 6.51e+01 & \best 2.09e+01 & \statsimilar 2.71e-01 & \statsimilar 3.67e-01 & \statsimilar 3.08e-06 & \statsimilar 3.94e-06 & \statsimilar 5.62e-05 & \statsimilar 6.94e-05 \\
\bottomrule
\toprule
    \bfseries Method
    & \multicolumn{2}{c}{\bfseries Ackley5 (5)} 
    & \multicolumn{2}{c}{\bfseries Michalewicz5 (5)} 
    & \multicolumn{2}{c}{\bfseries StyblinskiTang5 (5)} 
    & \multicolumn{2}{c}{\bfseries Hartmann6 (6)} 
    & \multicolumn{2}{c}{\bfseries Rosenbrock7 (7)} \\ 
    & \multicolumn{1}{c}{Median} & \multicolumn{1}{c}{MAD}
    & \multicolumn{1}{c}{Median} & \multicolumn{1}{c}{MAD}
    & \multicolumn{1}{c}{Median} & \multicolumn{1}{c}{MAD}
    & \multicolumn{1}{c}{Median} & \multicolumn{1}{c}{MAD}
    & \multicolumn{1}{c}{Median} & \multicolumn{1}{c}{MAD}  \\ \midrule
    Faster & 5.00e+00 & 3.62e+00 & \best 1.36e+00 & \best 5.55e-01 & \best 1.45e+01 & \best 2.03e+01 & \statsimilar 2.59e-03 & \statsimilar 2.83e-03 & 5.34e+02 & 3.07e+02 \\
    Slower & \statsimilar 2.95e+00 & \statsimilar 7.71e-01 & \statsimilar 1.47e+00 & \statsimilar 4.00e-01 & \statsimilar 1.47e+01 & \statsimilar 3.93e+00 & \statsimilar 2.45e-03 & \statsimilar 3.39e-03 & \best 3.49e+02 & \best 2.82e+02 \\
    AEGiS & \best 2.90e+00 & \best 1.44e+00 & \statsimilar 1.39e+00 & \statsimilar 5.38e-01 & \statsimilar 1.46e+01 & \statsimilar 2.95e+00 & \best 2.21e-03 & \best 2.77e-03 & \statsimilar 4.20e+02 & \statsimilar 2.33e+02 \\
\bottomrule
\toprule
    \bfseries Method
    & \multicolumn{2}{c}{\bfseries StyblinskiTang7 (7)} 
    & \multicolumn{2}{c}{\bfseries Ackley10 (10)} 
    & \multicolumn{2}{c}{\bfseries Michalewicz10 (10)} 
    & \multicolumn{2}{c}{\bfseries Rosenbrock10 (10)} 
    & \multicolumn{2}{c}{\bfseries StyblinskiTang10 (10)} \\ 
    & \multicolumn{1}{c}{Median} & \multicolumn{1}{c}{MAD}
    & \multicolumn{1}{c}{Median} & \multicolumn{1}{c}{MAD}
    & \multicolumn{1}{c}{Median} & \multicolumn{1}{c}{MAD}
    & \multicolumn{1}{c}{Median} & \multicolumn{1}{c}{MAD}
    & \multicolumn{1}{c}{Median} & \multicolumn{1}{c}{MAD}  \\ \midrule
    Faster & \best 2.95e+01 & \best 1.98e+01 & 1.54e+01 & 3.28e+00 & \statsimilar 5.61e+00 & \statsimilar 7.34e-01 & 1.69e+03 & 1.44e+03 & \best 5.71e+01 & \best 2.00e+01 \\
    Slower & \statsimilar 3.08e+01 & \statsimilar 1.87e+01 & \best 1.16e+01 & \best 3.73e+00 & \statsimilar 5.48e+00 & \statsimilar 5.56e-01 & \best 9.86e+02 & \best 3.87e+02 & 6.20e+01 & 2.15e+01 \\
    AEGiS & \statsimilar 3.07e+01 & \statsimilar 1.89e+01 & \statsimilar 1.27e+01 & \statsimilar 2.38e+00 & \best 5.42e+00 & \best 6.94e-01 & \statsimilar 1.03e+03 & \statsimilar 7.26e+02 & \statsimilar 5.95e+01 & \statsimilar 2.10e+01 \\
\bottomrule
    \end{tabular}
}
\label{tbl:q1:epsbracket}
\end{table}

\section{Proportion of TS to Pareto Set Selection}
\label{sec:tsprop}
\begin{figure}[H]
\centering
\includegraphics[width=0.5\linewidth, clip, trim={5 0 5 7}]{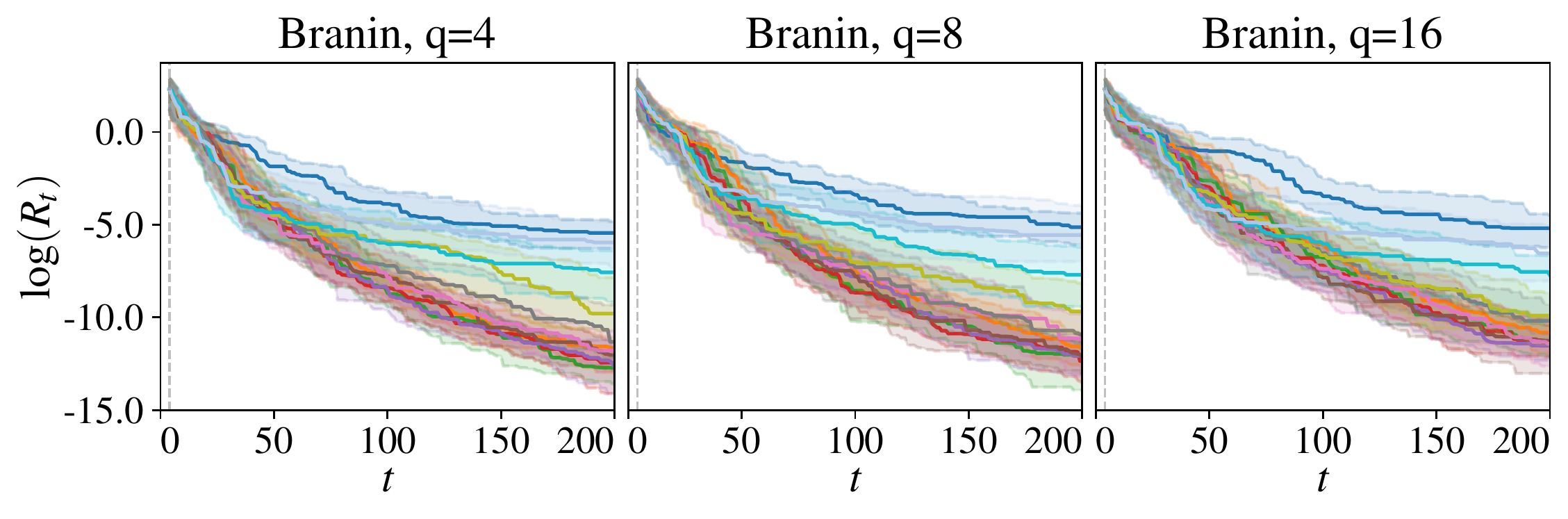}%
\includegraphics[width=0.5\linewidth, clip, trim={5 0 5 7}]{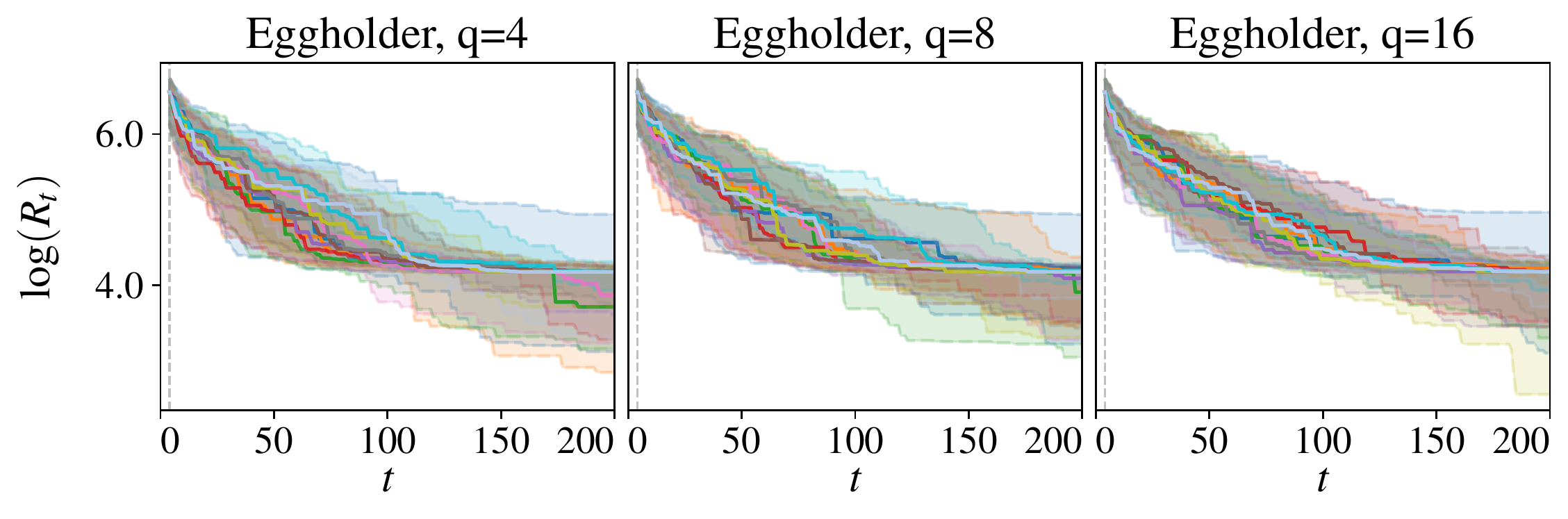}\\
\includegraphics[width=0.5\linewidth, clip, trim={5 0 5 7}]{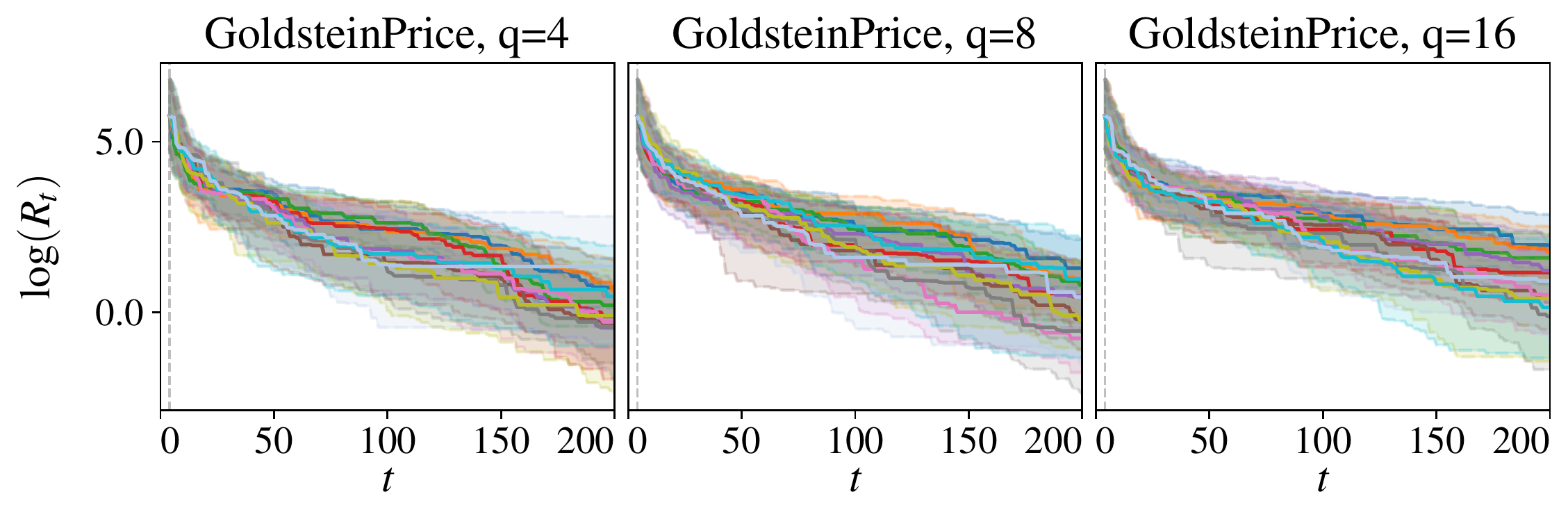}%
\includegraphics[width=0.5\linewidth, clip, trim={5 0 5 7}]{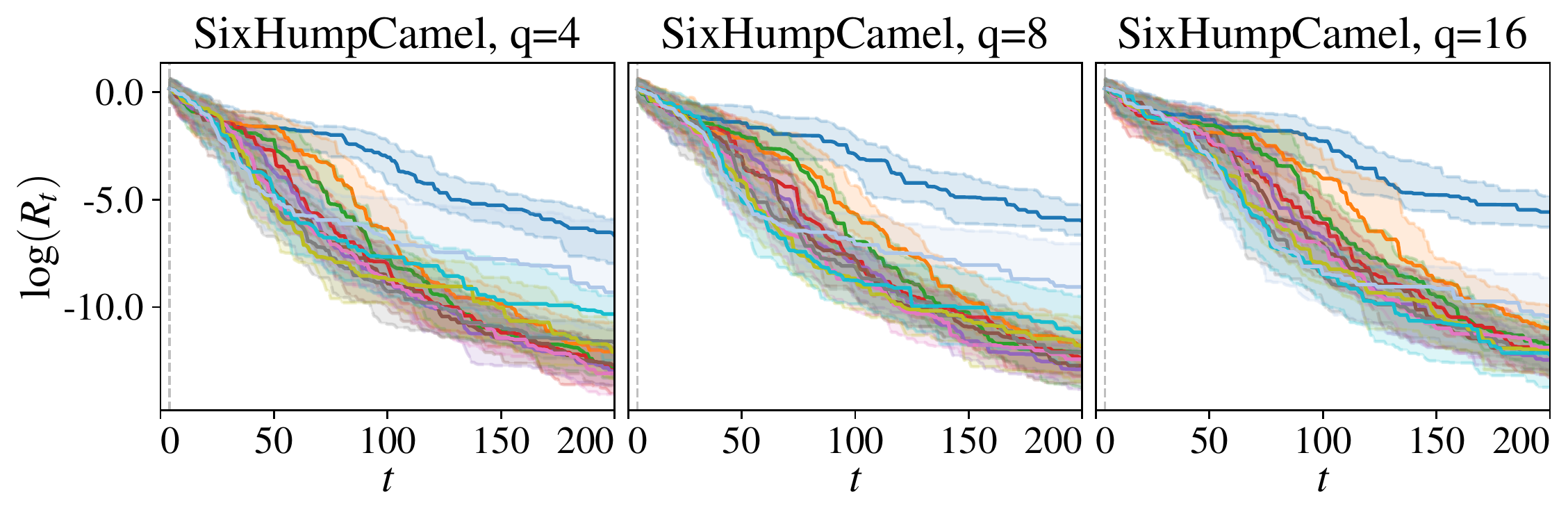}\\
\includegraphics[width=0.5\linewidth, clip, trim={5 0 5 7}]{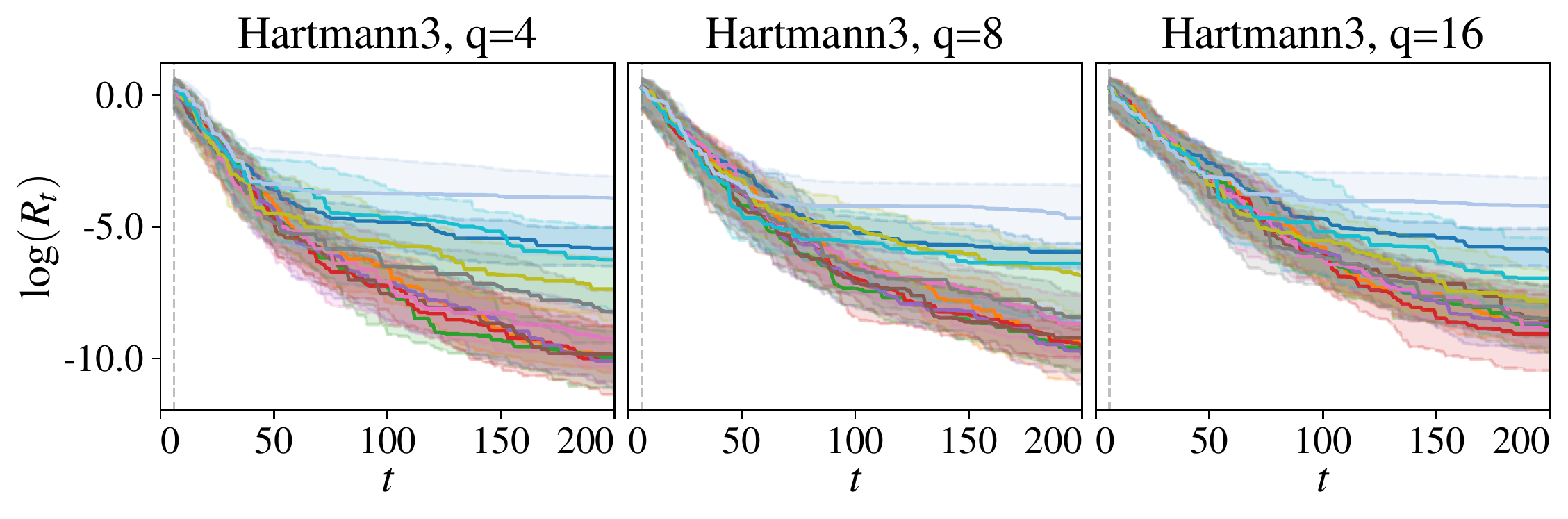}%
\includegraphics[width=0.5\linewidth, clip, trim={5 0 5 7}]{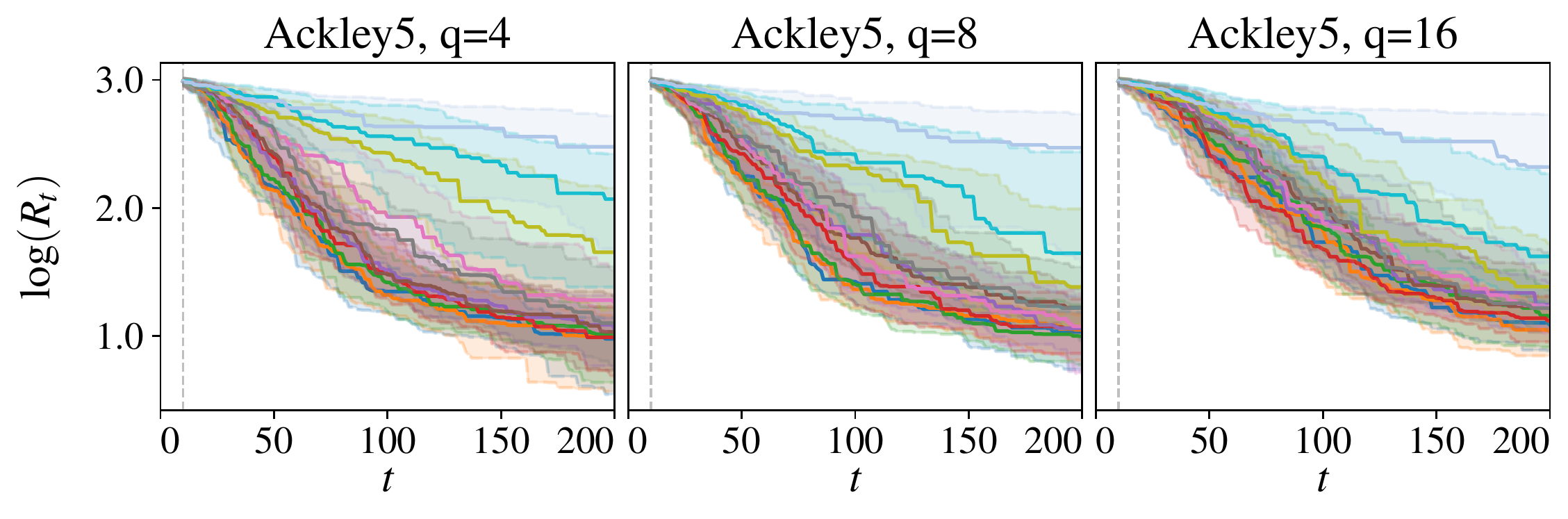}\\
\includegraphics[width=0.5\linewidth, clip, trim={5 0 5 7}]{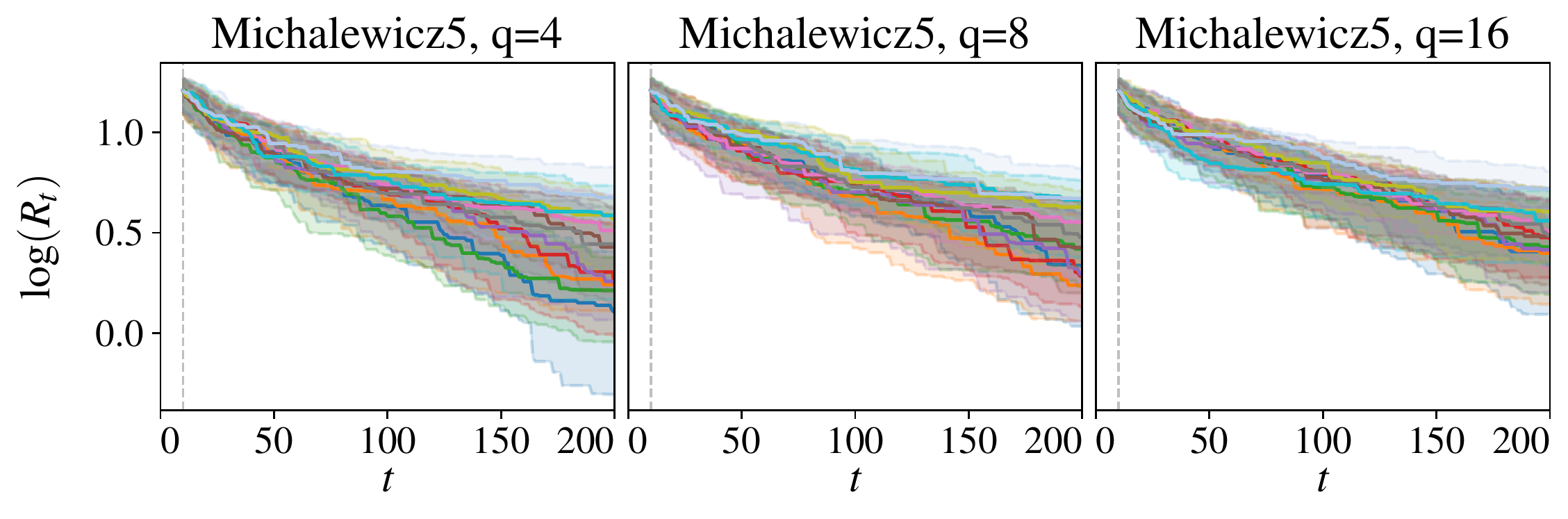}%
\includegraphics[width=0.5\linewidth, clip, trim={5 0 5 7}]{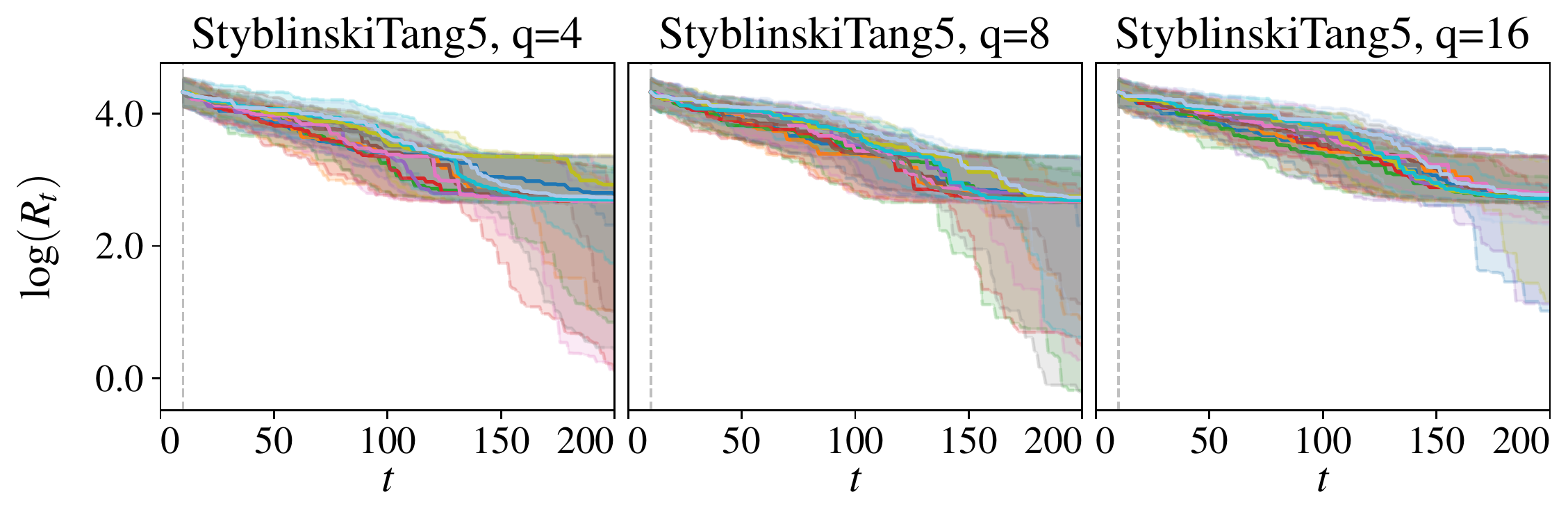}\\
\includegraphics[width=0.5\linewidth, clip, trim={5 0 5 7}]{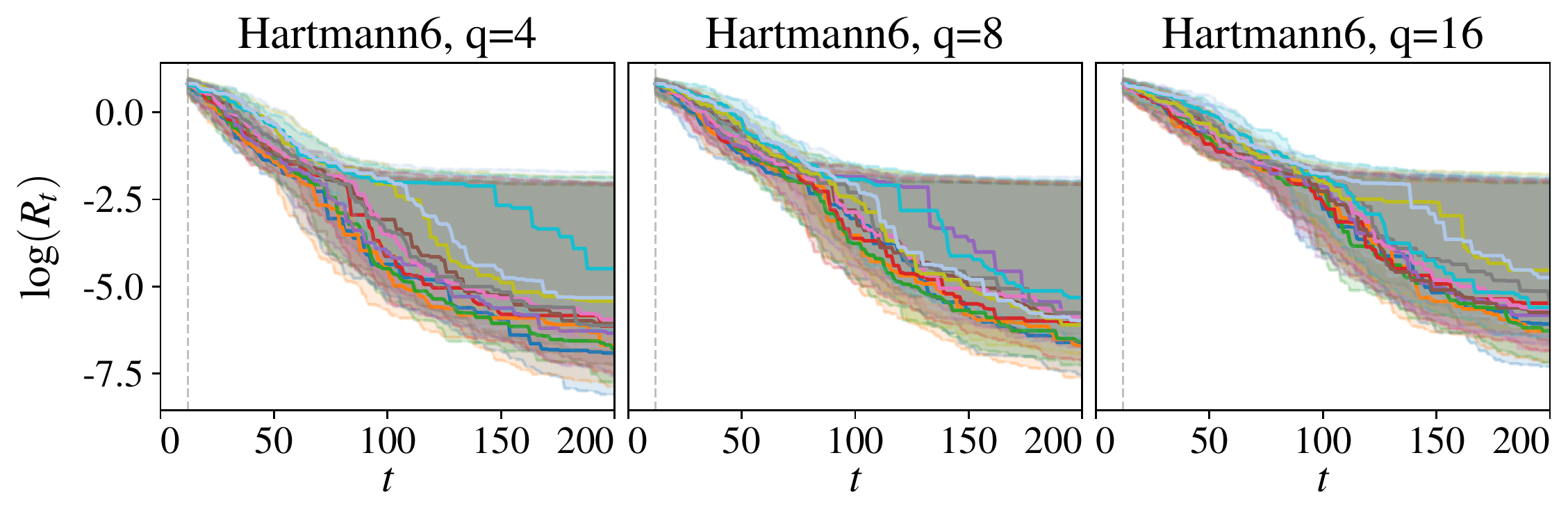}%
\includegraphics[width=0.5\linewidth, clip, trim={5 0 5 7}]{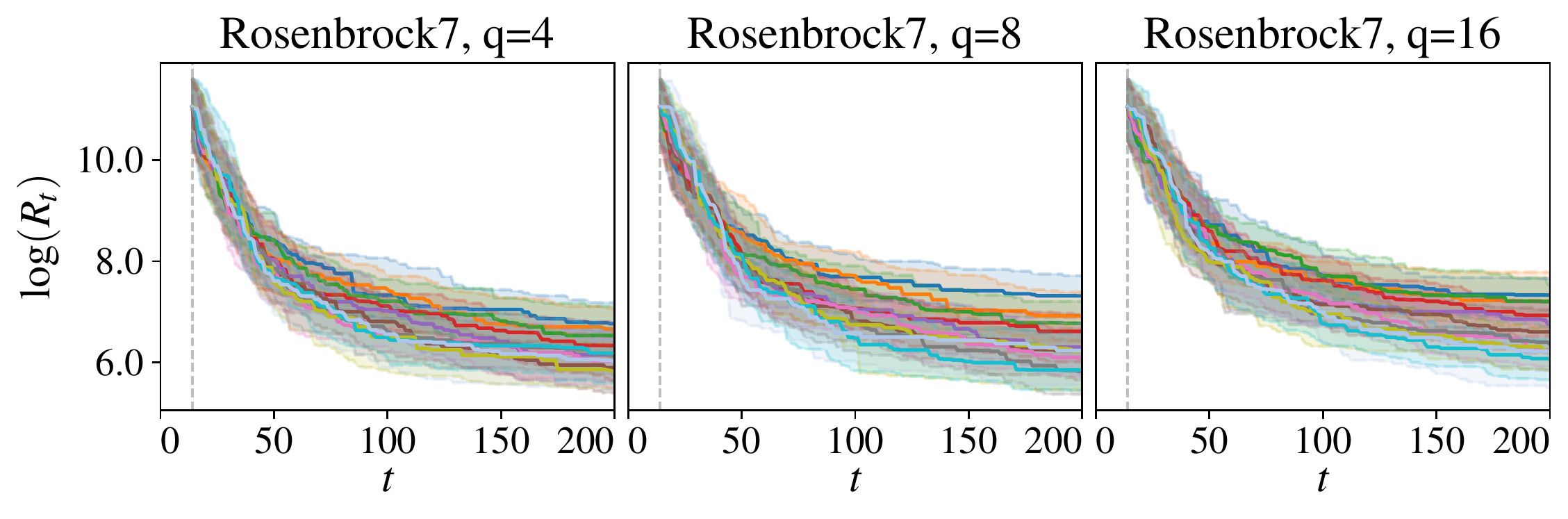}\\
\includegraphics[width=0.5\linewidth, clip, trim={5 0 5 7}]{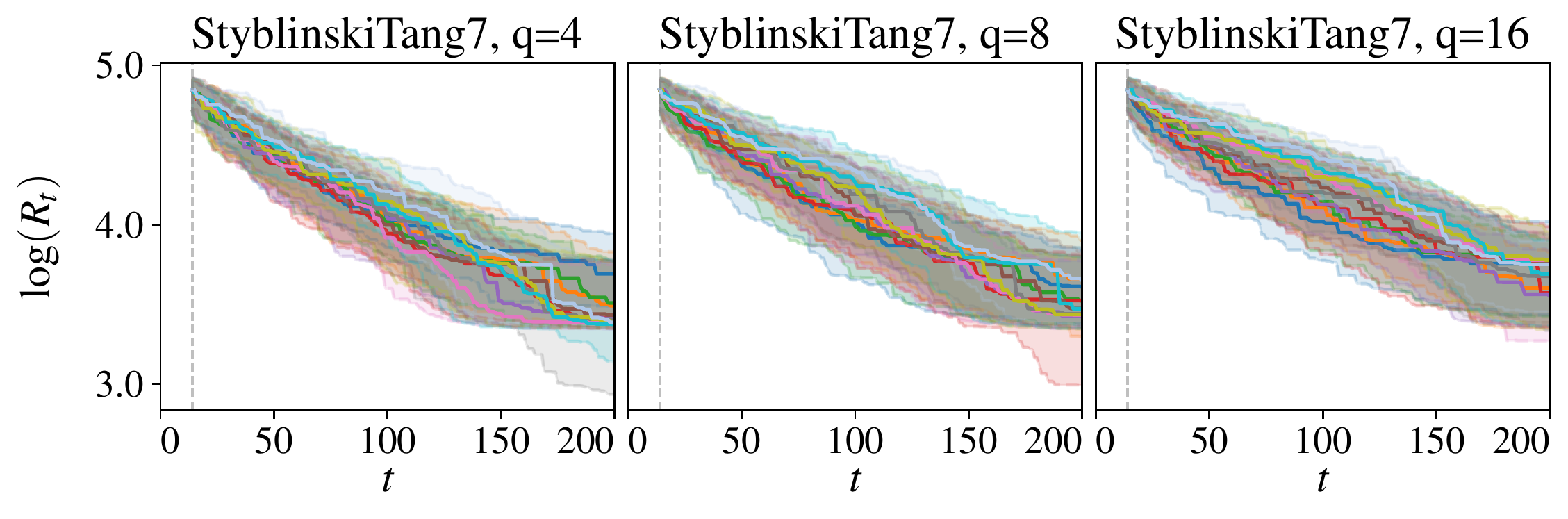}%
\includegraphics[width=0.5\linewidth, clip, trim={5 0 5 7}]{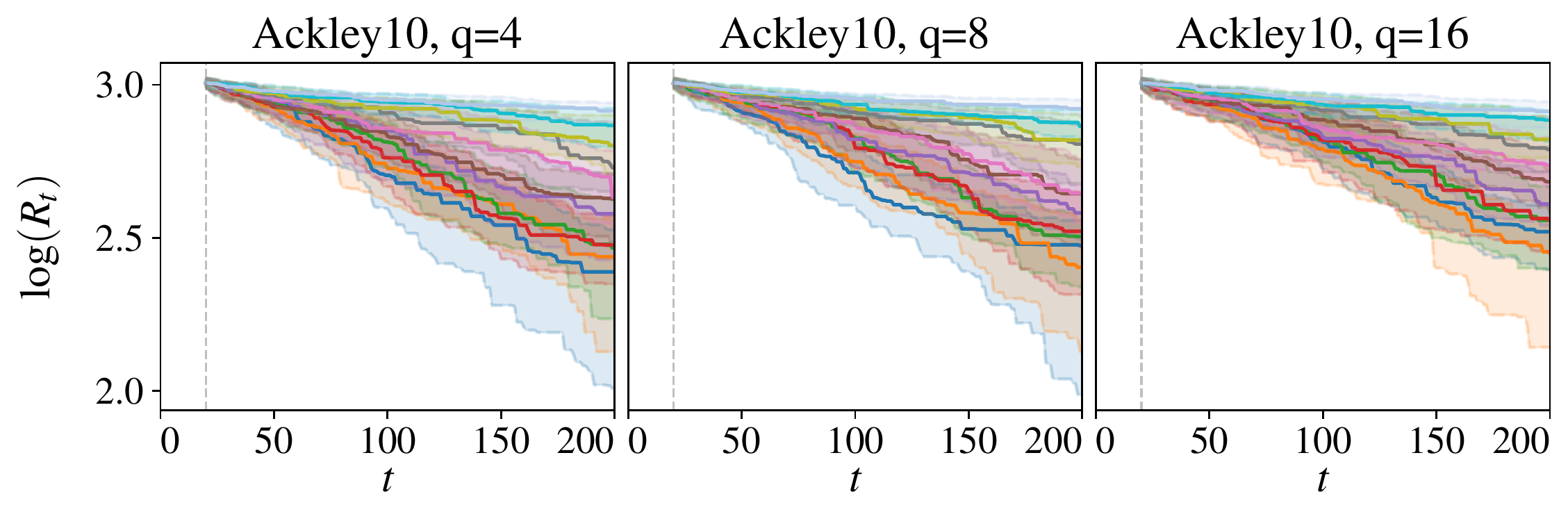}\\
\includegraphics[width=0.5\linewidth, clip, trim={5 0 5 7}]{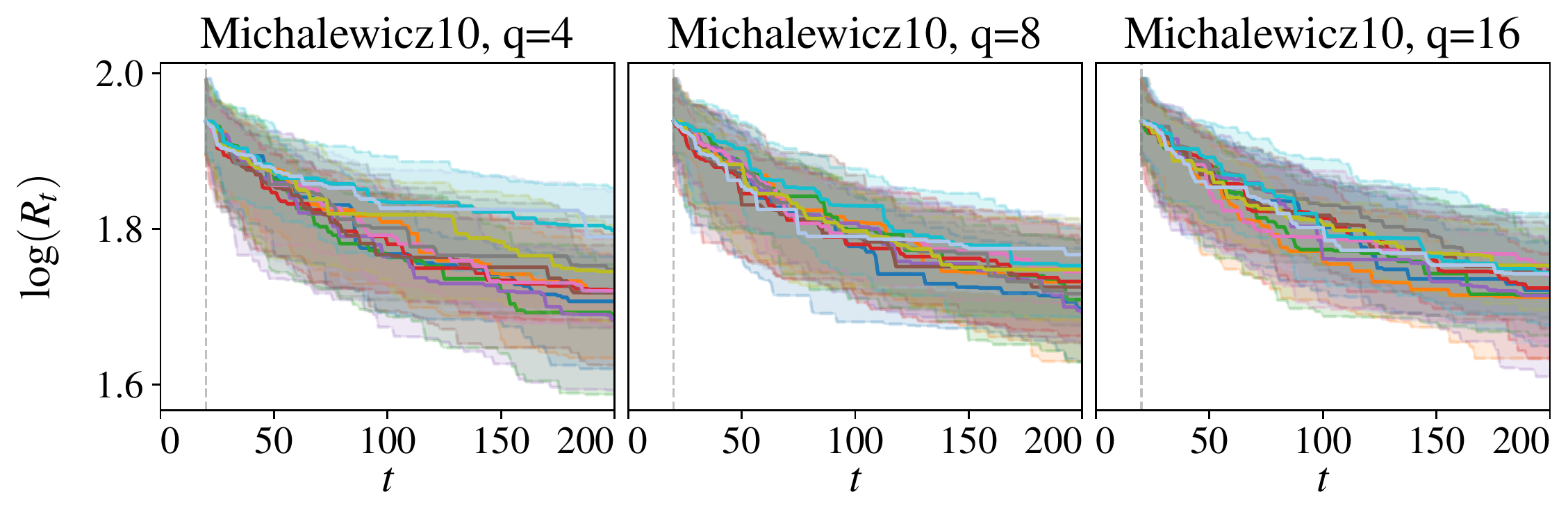}%
\includegraphics[width=0.5\linewidth, clip, trim={5 0 5 7}]{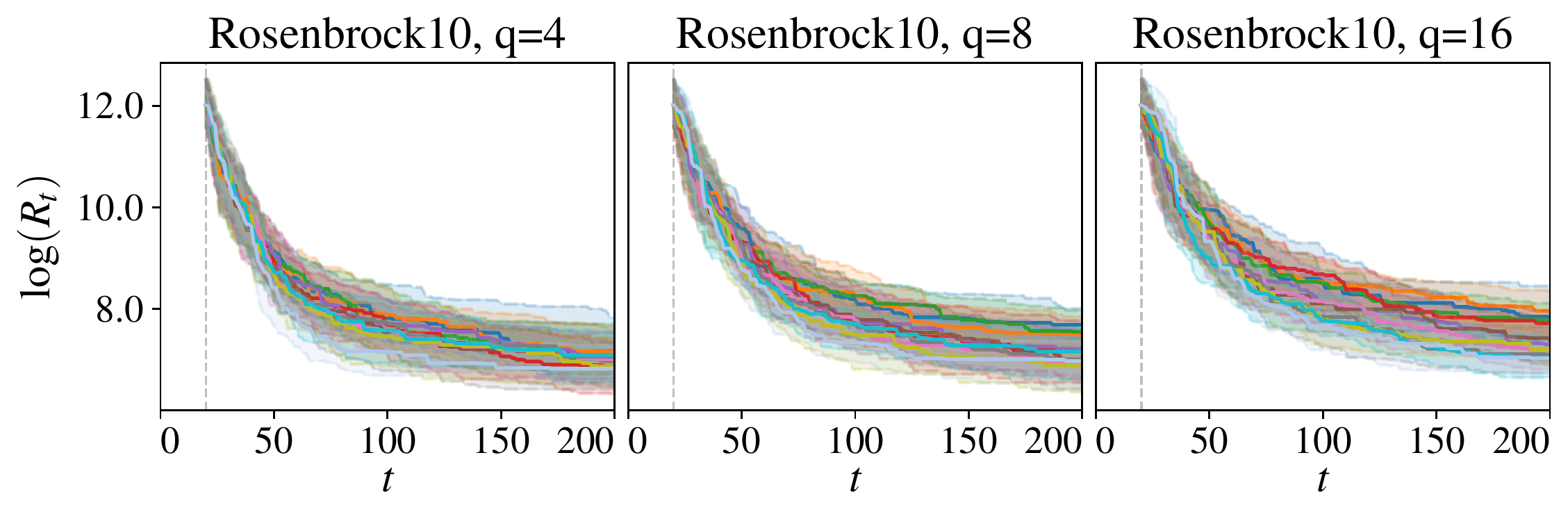}\\
\includegraphics[width=0.5\linewidth, clip, trim={5 0 5 7}]{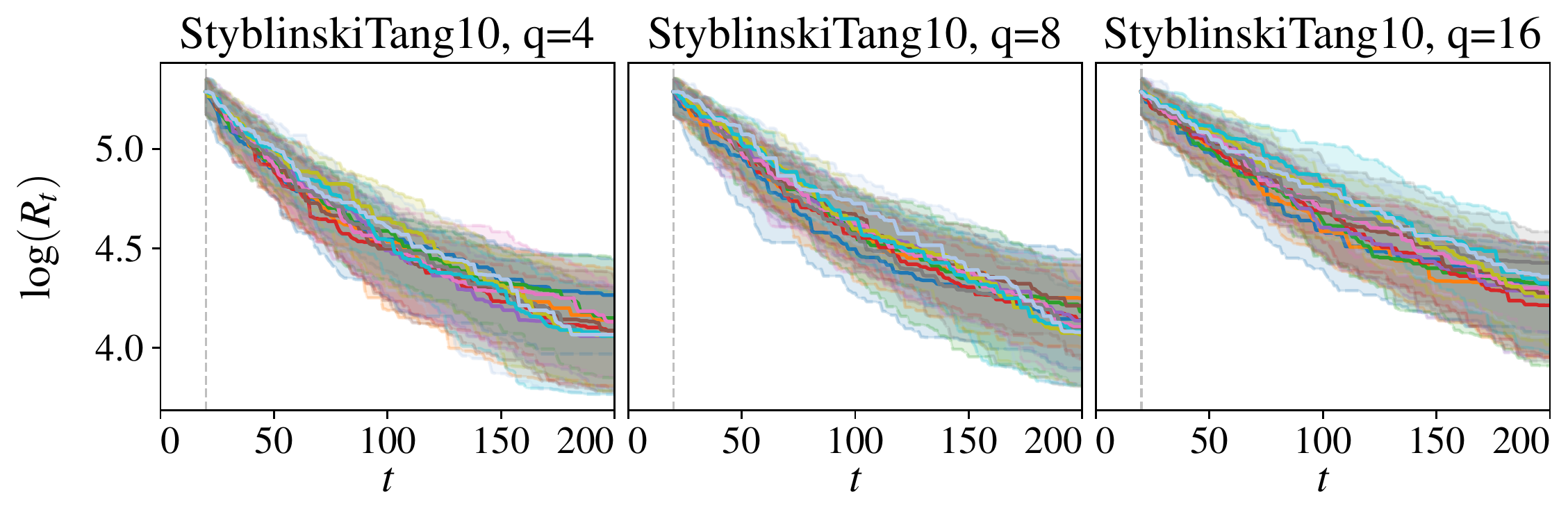}\\
\includegraphics[width=1\linewidth, clip, trim={10 15 10 13}]{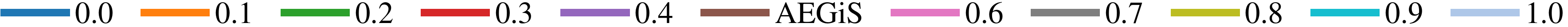}%
\caption{Convergence results for AEGiS with $\gamma \in \{ 0, 0.1, \dots, 1 \}$.}
\label{fig:results:tsprop}
\end{figure}

In this section we show all convergence plots and results tables for the
investigation into the optimal selection ratio between TS and Pareto set
selection on the synthetic benchmark functions for $q \in \{4,8,16\}$.
Specifically we evaluate AEGiS on the synthetic benchmark functions
for $q \in \{4,8,16\}$ with the split between exploitation, TS and Pareto set
selection being $1-\epsilon$, $\epsilon_T = \gamma \epsilon$ and
$\epsilon_P = (1 - \gamma)\epsilon$ respectively and $\gamma \in \{ 0,
0.1, \dots, 1 \}$.
Note that we use the default value $\epsilon = \min(2 / \sqrt{d}, 1)$.
Figure~\ref{fig:results:tsprop} shows the convergence plots for
the 15 synthetic benchmark problems and and Tables~\ref{tbl:tsprop_results_4},
\ref{tbl:tsprop_results_8} and \ref{tbl:tsprop_results_16} show median log
simple regret as well as the median absolute deviation from the median.

\begin{table}[H]
\setlength{\tabcolsep}{2pt}
\sisetup{table-format=1.2e-1,table-number-alignment=center}
\caption{Tabulated results for $q=4$ workers for 
$\gamma \in \{0, 0.1, \dots, 1\}$. The table shows the median log simple regret
(\emph{left}) and median absolute deviation from the median (MAD, \emph{right})
after 200 function evaluations across the 51 runs. The method with the lowest
median performance is shown in dark grey, with those with statistically
equivalent performance are shown in light grey.}
\resizebox{1\textwidth}{!}{%
\begin{tabular}{l Sz Sz Sz Sz Sz}
    \toprule
    \bfseries Method
    & \multicolumn{2}{c}{\bfseries Branin (2)} 
    & \multicolumn{2}{c}{\bfseries Eggholder (2)} 
    & \multicolumn{2}{c}{\bfseries GoldsteinPrice (2)} 
    & \multicolumn{2}{c}{\bfseries SixHumpCamel (2)} 
    & \multicolumn{2}{c}{\bfseries Hartmann3 (3)} \\ 
    & \multicolumn{1}{c}{Median} & \multicolumn{1}{c}{MAD}
    & \multicolumn{1}{c}{Median} & \multicolumn{1}{c}{MAD}
    & \multicolumn{1}{c}{Median} & \multicolumn{1}{c}{MAD}
    & \multicolumn{1}{c}{Median} & \multicolumn{1}{c}{MAD}
    & \multicolumn{1}{c}{Median} & \multicolumn{1}{c}{MAD}  \\ \midrule
    0.0 & 4.31e-03 & 4.14e-03 & \statsimilar 6.57e+01 & \statsimilar 8.01e+01 & 2.06e+00 & 2.23e+00 & 1.25e-03 & 1.45e-03 & 2.95e-03 & 3.03e-03 \\
    0.1 & 8.79e-06 & 9.91e-06 & \statsimilar 6.51e+01 & \statsimilar 5.55e+01 & 1.88e+00 & 2.24e+00 & 5.71e-06 & 6.51e-06 & \statsimilar 4.81e-05 & \statsimilar 5.92e-05 \\
    0.2 & \best 3.01e-06 & \best 3.01e-06 & \best 4.09e+01 & \best 3.61e+01 & \statsimilar 1.23e+00 & \statsimilar 1.10e+00 & \statsimilar 2.75e-06 & \statsimilar 2.61e-06 & \statsimilar 4.70e-05 & \statsimilar 5.47e-05 \\
    0.3 & \statsimilar 3.93e-06 & \statsimilar 4.77e-06 & \statsimilar 6.51e+01 & \statsimilar 1.15e+01 & \statsimilar 8.79e-01 & \statsimilar 1.08e+00 & \statsimilar 2.41e-06 & \statsimilar 2.67e-06 & \statsimilar 4.15e-05 & \statsimilar 5.23e-05 \\
    0.4 & \statsimilar 3.27e-06 & \statsimilar 4.19e-06 & \statsimilar 6.51e+01 & \statsimilar 8.93e+00 & \best 6.32e-01 & \best 7.85e-01 & \best 2.02e-06 & \best 2.58e-06 & \best 4.12e-05 & \best 5.61e-05 \\
    AEGiS & 5.99e-06 & 6.90e-06 & \statsimilar 6.52e+01 & \statsimilar 1.08e+01 & \statsimilar 6.99e-01 & \statsimilar 7.67e-01 & \statsimilar 2.93e-06 & \statsimilar 3.31e-06 & \statsimilar 5.29e-05 & \statsimilar 5.59e-05 \\
    0.6 & 6.94e-06 & 8.01e-06 & \statsimilar 4.78e+01 & \statsimilar 3.05e+01 & \statsimilar 7.51e-01 & \statsimilar 9.80e-01 & \statsimilar 2.08e-06 & \statsimilar 2.38e-06 & \statsimilar 9.57e-05 & \statsimilar 1.20e-04 \\
    0.7 & 1.22e-05 & 1.66e-05 & \statsimilar 6.51e+01 & \statsimilar 1.43e+01 & \statsimilar 6.47e-01 & \statsimilar 8.93e-01 & 8.43e-06 & 1.03e-05 & 2.67e-04 & 3.28e-04 \\
    0.8 & 5.61e-05 & 8.25e-05 & \statsimilar 6.51e+01 & \statsimilar 1.01e+01 & \statsimilar 8.99e-01 & \statsimilar 1.32e+00 & 5.63e-06 & 7.55e-06 & 6.34e-04 & 7.93e-04 \\
    0.9 & 5.15e-04 & 6.74e-04 & \statsimilar 6.51e+01 & \statsimilar 1.50e+01 & 1.58e+00 & 2.31e+00 & 3.28e-05 & 4.52e-05 & 1.92e-03 & 2.74e-03 \\
    1.0 & 2.59e-03 & 3.43e-03 & \statsimilar 6.51e+01 & \statsimilar 1.42e+01 & 3.83e+00 & 5.62e+00 & 9.05e-05 & 1.34e-04 & 1.97e-02 & 2.72e-02 \\
\bottomrule
\toprule
    \bfseries Method
    & \multicolumn{2}{c}{\bfseries Ackley5 (5)} 
    & \multicolumn{2}{c}{\bfseries Michalewicz5 (5)} 
    & \multicolumn{2}{c}{\bfseries StyblinskiTang5 (5)} 
    & \multicolumn{2}{c}{\bfseries Hartmann6 (6)} 
    & \multicolumn{2}{c}{\bfseries Rosenbrock7 (7)} \\ 
    & \multicolumn{1}{c}{Median} & \multicolumn{1}{c}{MAD}
    & \multicolumn{1}{c}{Median} & \multicolumn{1}{c}{MAD}
    & \multicolumn{1}{c}{Median} & \multicolumn{1}{c}{MAD}
    & \multicolumn{1}{c}{Median} & \multicolumn{1}{c}{MAD}
    & \multicolumn{1}{c}{Median} & \multicolumn{1}{c}{MAD}  \\ \midrule
    0.0 & \best 2.66e+00 & \best 9.48e-01 & \best 1.11e+00 & \best 5.75e-01 & \statsimilar 1.61e+01 & \statsimilar 9.85e+00 & \best 1.00e-03 & \best 1.28e-03 & 8.69e+02 & 6.58e+02 \\
    0.1 & \statsimilar 2.69e+00 & \statsimilar 1.09e+00 & \statsimilar 1.27e+00 & \statsimilar 5.23e-01 & \statsimilar 1.47e+01 & \statsimilar 1.48e+01 & \statsimilar 1.10e-03 & \statsimilar 1.57e-03 & 7.26e+02 & 5.25e+02 \\
    0.2 & \statsimilar 2.76e+00 & \statsimilar 1.00e+00 & \statsimilar 1.24e+00 & \statsimilar 4.11e-01 & \statsimilar 1.46e+01 & \statsimilar 1.88e+01 & \statsimilar 1.10e-03 & \statsimilar 1.26e-03 & 6.88e+02 & 4.17e+02 \\
    0.3 & \statsimilar 2.69e+00 & \statsimilar 1.08e+00 & \statsimilar 1.28e+00 & \statsimilar 5.30e-01 & \statsimilar 1.46e+01 & \statsimilar 1.99e+01 & \statsimilar 2.17e-03 & \statsimilar 2.91e-03 & \statsimilar 5.65e+02 & \statsimilar 4.25e+02 \\
    0.4 & \statsimilar 2.87e+00 & \statsimilar 1.19e+00 & 1.29e+00 & 5.00e-01 & \statsimilar 1.49e+01 & \statsimilar 2.04e+00 & \statsimilar 1.77e-03 & \statsimilar 2.33e-03 & \statsimilar 4.69e+02 & \statsimilar 4.00e+02 \\
    AEGiS & \statsimilar 2.81e+00 & \statsimilar 1.19e+00 & 1.54e+00 & 4.77e-01 & \statsimilar 1.51e+01 & \statsimilar 2.00e+01 & 2.28e-03 & 3.09e-03 & \statsimilar 3.64e+02 & \statsimilar 2.44e+02 \\
    0.6 & 3.52e+00 & 1.28e+00 & 1.67e+00 & 5.18e-01 & \best 1.45e+01 & \best 1.96e+01 & \statsimilar 2.62e-03 & \statsimilar 3.34e-03 & \statsimilar 4.15e+02 & \statsimilar 3.00e+02 \\
    0.7 & 3.01e+00 & 1.61e+00 & 1.56e+00 & 5.82e-01 & \statsimilar 1.50e+01 & \statsimilar 2.00e+01 & 2.20e-03 & 2.87e-03 & \statsimilar 4.59e+02 & \statsimilar 2.91e+02 \\
    0.8 & 5.23e+00 & 3.34e+00 & 1.77e+00 & 4.13e-01 & 1.87e+01 & 1.47e+01 & 4.43e-03 & 6.11e-03 & \best 3.39e+02 & \best 1.86e+02 \\
    0.9 & 7.92e+00 & 5.46e+00 & 1.80e+00 & 5.88e-01 & \statsimilar 1.49e+01 & \statsimilar 1.87e+01 & 1.12e-02 & 1.64e-02 & \statsimilar 4.86e+02 & \statsimilar 3.78e+02 \\
    1.0 & 1.19e+01 & 6.42e+00 & 1.96e+00 & 5.73e-01 & \statsimilar 1.53e+01 & \statsimilar 1.93e+01 & 4.90e-03 & 6.64e-03 & \statsimilar 4.17e+02 & \statsimilar 3.55e+02 \\
\bottomrule
\toprule
    \bfseries Method
    & \multicolumn{2}{c}{\bfseries StyblinskiTang7 (7)} 
    & \multicolumn{2}{c}{\bfseries Ackley10 (10)} 
    & \multicolumn{2}{c}{\bfseries Michalewicz10 (10)} 
    & \multicolumn{2}{c}{\bfseries Rosenbrock10 (10)} 
    & \multicolumn{2}{c}{\bfseries StyblinskiTang10 (10)} \\ 
    & \multicolumn{1}{c}{Median} & \multicolumn{1}{c}{MAD}
    & \multicolumn{1}{c}{Median} & \multicolumn{1}{c}{MAD}
    & \multicolumn{1}{c}{Median} & \multicolumn{1}{c}{MAD}
    & \multicolumn{1}{c}{Median} & \multicolumn{1}{c}{MAD}
    & \multicolumn{1}{c}{Median} & \multicolumn{1}{c}{MAD}  \\ \midrule
    0.0 & 4.01e+01 & 1.40e+01 & \best 1.09e+01 & \best 3.36e+00 & \statsimilar 5.51e+00 & \statsimilar 5.36e-01 & \statsimilar 1.29e+03 & \statsimilar 9.35e+02 & \statsimilar 7.12e+01 & \statsimilar 2.36e+01 \\
    0.1 & \statsimilar 3.27e+01 & \statsimilar 1.21e+01 & \statsimilar 1.14e+01 & \statsimilar 2.53e+00 & \statsimilar 5.59e+00 & \statsimilar 5.48e-01 & \statsimilar 1.28e+03 & \statsimilar 7.31e+02 & \statsimilar 6.23e+01 & \statsimilar 2.75e+01 \\
    0.2 & \statsimilar 3.34e+01 & \statsimilar 1.42e+01 & \statsimilar 1.18e+01 & \statsimilar 3.20e+00 & \statsimilar 5.42e+00 & \statsimilar 6.76e-01 & \statsimilar 1.12e+03 & \statsimilar 8.05e+02 & \statsimilar 6.34e+01 & \statsimilar 2.92e+01 \\
    0.3 & \statsimilar 3.01e+01 & \statsimilar 1.80e+01 & \statsimilar 1.19e+01 & \statsimilar 2.00e+00 & \statsimilar 5.59e+00 & \statsimilar 4.29e-01 & \statsimilar 9.57e+02 & \statsimilar 6.96e+02 & \statsimilar 5.95e+01 & \statsimilar 2.24e+01 \\
    0.4 & \statsimilar 3.03e+01 & \statsimilar 1.87e+01 & 1.32e+01 & 2.75e+00 & \best 5.38e+00 & \best 7.18e-01 & \statsimilar 1.08e+03 & \statsimilar 7.91e+02 & \statsimilar 5.79e+01 & \statsimilar 2.31e+01 \\
    AEGiS & \statsimilar 3.10e+01 & \statsimilar 1.79e+01 & 1.38e+01 & 2.30e+00 & \statsimilar 5.57e+00 & \statsimilar 7.16e-01 & \statsimilar 1.08e+03 & \statsimilar 7.27e+02 & \statsimilar 5.96e+01 & \statsimilar 2.41e+01 \\
    0.6 & \statsimilar 2.94e+01 & \statsimilar 9.29e+00 & 1.39e+01 & 2.91e+00 & 5.59e+00 & 5.22e-01 & \statsimilar 1.09e+03 & \statsimilar 7.03e+02 & \statsimilar 6.23e+01 & \statsimilar 2.17e+01 \\
    0.7 & \statsimilar 3.02e+01 & \statsimilar 2.03e+01 & 1.53e+01 & 2.87e+00 & 5.74e+00 & 6.28e-01 & \statsimilar 1.17e+03 & \statsimilar 9.73e+02 & \statsimilar 5.81e+01 & \statsimilar 2.21e+01 \\
    0.8 & \statsimilar 2.92e+01 & \statsimilar 2.00e+01 & 1.65e+01 & 2.47e+00 & 5.73e+00 & 5.23e-01 & \statsimilar 9.63e+02 & \statsimilar 6.63e+02 & \best 5.79e+01 & \best 2.11e+01 \\
    0.9 & \best 2.92e+01 & \best 1.98e+01 & 1.76e+01 & 1.66e+00 & 6.04e+00 & 5.92e-01 & \statsimilar 1.20e+03 & \statsimilar 8.13e+02 & \statsimilar 5.79e+01 & \statsimilar 2.18e+01 \\
    1.0 & \statsimilar 2.93e+01 & \statsimilar 1.96e+01 & 1.84e+01 & 8.55e-01 & 5.99e+00 & 6.75e-01 & \best 9.18e+02 & \best 6.00e+02 & \statsimilar 5.83e+01 & \statsimilar 2.25e+01 \\
\bottomrule
\end{tabular}
}
\label{tbl:tsprop_results_4}
\end{table}

\begin{table}[H]
\setlength{\tabcolsep}{2pt}
\sisetup{table-format=1.2e-1,table-number-alignment=center}
\caption{Tabulated results for $q = 8$ workers for 
$\gamma \in \{0, 0.1, \dots, 1\}$. The table shows the median log simple regret
(\emph{left}) and median absolute deviation from the median (MAD, \emph{right})
after 200 function evaluations across the 51 runs. The method with the lowest
median performance is shown in dark grey, with those with statistically
equivalent performance are shown in light grey.}
\resizebox{1\textwidth}{!}{%
\begin{tabular}{l Sz Sz Sz Sz Sz}
    \toprule
    \bfseries Method
    & \multicolumn{2}{c}{\bfseries Branin (2)} 
    & \multicolumn{2}{c}{\bfseries Eggholder (2)} 
    & \multicolumn{2}{c}{\bfseries GoldsteinPrice (2)} 
    & \multicolumn{2}{c}{\bfseries SixHumpCamel (2)} 
    & \multicolumn{2}{c}{\bfseries Hartmann3 (3)} \\ 
    & \multicolumn{1}{c}{Median} & \multicolumn{1}{c}{MAD}
    & \multicolumn{1}{c}{Median} & \multicolumn{1}{c}{MAD}
    & \multicolumn{1}{c}{Median} & \multicolumn{1}{c}{MAD}
    & \multicolumn{1}{c}{Median} & \multicolumn{1}{c}{MAD}
    & \multicolumn{1}{c}{Median} & \multicolumn{1}{c}{MAD}  \\ \midrule
    0.0 & 5.86e-03 & 6.80e-03 & 6.89e+01 & 8.42e+01 & 3.64e+00 & 3.10e+00 & 2.48e-03 & 2.67e-03 & 2.61e-03 & 1.75e-03 \\
    0.1 & \statsimilar 8.73e-06 & \statsimilar 9.79e-06 & \statsimilar 6.62e+01 & \statsimilar 4.71e+01 & 2.58e+00 & 2.69e+00 & 6.97e-06 & 7.29e-06 & \statsimilar 5.90e-05 & \statsimilar 7.27e-05 \\
    0.2 & \statsimilar 5.62e-06 & \statsimilar 7.18e-06 & \best 4.98e+01 & \best 3.54e+01 & \statsimilar 2.21e+00 & \statsimilar 2.34e+00 & \statsimilar 5.62e-06 & \statsimilar 6.64e-06 & \statsimilar 6.91e-05 & \statsimilar 8.71e-05 \\
    0.3 & \best 4.26e-06 & \best 5.78e-06 & \statsimilar 6.51e+01 & \statsimilar 8.78e+00 & \statsimilar 1.57e+00 & \statsimilar 1.78e+00 & \statsimilar 4.29e-06 & \statsimilar 4.97e-06 & \statsimilar 7.20e-05 & \statsimilar 6.30e-05 \\
    0.4 & \statsimilar 5.50e-06 & \statsimilar 5.91e-06 & \statsimilar 5.97e+01 & \statsimilar 2.15e+01 & \statsimilar 1.61e+00 & \statsimilar 1.92e+00 & \best 2.49e-06 & \best 2.32e-06 & \best 5.31e-05 & \best 6.85e-05 \\
    AEGiS & \statsimilar 5.32e-06 & \statsimilar 6.51e-06 & \statsimilar 6.51e+01 & \statsimilar 1.35e+01 & \statsimilar 8.10e-01 & \statsimilar 8.82e-01 & \statsimilar 2.98e-06 & \statsimilar 3.83e-06 & \statsimilar 9.99e-05 & \statsimilar 1.03e-04 \\
    0.6 & 1.44e-05 & 2.03e-05 & \statsimilar 6.51e+01 & \statsimilar 2.41e+01 & \best 4.63e-01 & \best 6.19e-01 & \statsimilar 4.02e-06 & \statsimilar 4.17e-06 & 1.48e-04 & 2.06e-04 \\
    0.7 & 1.82e-05 & 2.48e-05 & \statsimilar 6.51e+01 & \statsimilar 2.09e+01 & \statsimilar 5.56e-01 & \statsimilar 7.69e-01 & 5.56e-06 & 6.83e-06 & 2.17e-04 & 2.81e-04 \\
    0.8 & 6.26e-05 & 9.11e-05 & \statsimilar 6.51e+01 & \statsimilar 1.76e+01 & \statsimilar 7.84e-01 & \statsimilar 1.10e+00 & 7.66e-06 & 1.03e-05 & 1.05e-03 & 1.41e-03 \\
    0.9 & 4.51e-04 & 6.51e-04 & \statsimilar 6.55e+01 & \statsimilar 9.19e+00 & \statsimilar 2.66e+00 & \statsimilar 3.84e+00 & 1.41e-05 & 1.94e-05 & 1.65e-03 & 2.13e-03 \\
    1.0 & 3.87e-03 & 5.73e-03 & \statsimilar 6.51e+01 & \statsimilar 1.15e+01 & 1.56e+00 & 2.28e+00 & 1.16e-04 & 1.70e-04 & 9.29e-03 & 1.33e-02 \\
\bottomrule
\toprule
    \bfseries Method
    & \multicolumn{2}{c}{\bfseries Ackley5 (5)} 
    & \multicolumn{2}{c}{\bfseries Michalewicz5 (5)} 
    & \multicolumn{2}{c}{\bfseries StyblinskiTang5 (5)} 
    & \multicolumn{2}{c}{\bfseries Hartmann6 (6)} 
    & \multicolumn{2}{c}{\bfseries Rosenbrock7 (7)} \\ 
    & \multicolumn{1}{c}{Median} & \multicolumn{1}{c}{MAD}
    & \multicolumn{1}{c}{Median} & \multicolumn{1}{c}{MAD}
    & \multicolumn{1}{c}{Median} & \multicolumn{1}{c}{MAD}
    & \multicolumn{1}{c}{Median} & \multicolumn{1}{c}{MAD}
    & \multicolumn{1}{c}{Median} & \multicolumn{1}{c}{MAD}  \\ \midrule
    0.0 & \statsimilar 2.81e+00 & \statsimilar 8.15e-01 & \statsimilar 1.40e+00 & \statsimilar 5.86e-01 & \statsimilar 1.55e+01 & \statsimilar 1.33e+01 & \statsimilar 1.36e-03 & \statsimilar 1.71e-03 & 1.50e+03 & 8.73e+02 \\
    0.1 & \statsimilar 2.85e+00 & \statsimilar 6.58e-01 & \best 1.27e+00 & \best 4.24e-01 & \statsimilar 1.46e+01 & \statsimilar 1.05e+01 & \best 1.23e-03 & \best 1.43e-03 & 1.01e+03 & 6.21e+02 \\
    0.2 & \best 2.71e+00 & \best 9.03e-01 & \statsimilar 1.52e+00 & \statsimilar 4.98e-01 & \statsimilar 1.46e+01 & \statsimilar 2.04e+01 & \statsimilar 1.40e-03 & \statsimilar 1.66e-03 & 8.75e+02 & 6.10e+02 \\
    0.3 & \statsimilar 2.89e+00 & \statsimilar 8.64e-01 & \statsimilar 1.33e+00 & \statsimilar 6.12e-01 & \best 1.45e+01 & \best 1.65e+01 & \statsimilar 2.37e-03 & \statsimilar 3.09e-03 & 7.45e+02 & 4.88e+02 \\
    0.4 & \statsimilar 2.89e+00 & \statsimilar 1.23e+00 & \statsimilar 1.34e+00 & \statsimilar 6.47e-01 & \statsimilar 1.50e+01 & \statsimilar 5.53e+00 & \statsimilar 3.15e-03 & \statsimilar 4.57e-03 & 5.45e+02 & 3.73e+02 \\
    AEGiS & \statsimilar 3.39e+00 & \statsimilar 8.01e-01 & \statsimilar 1.53e+00 & \statsimilar 5.62e-01 & \statsimilar 1.53e+01 & \statsimilar 1.89e+01 & \statsimilar 2.67e-03 & \statsimilar 3.16e-03 & \statsimilar 5.04e+02 & \statsimilar 3.54e+02 \\
    0.6 & \statsimilar 2.93e+00 & \statsimilar 1.33e+00 & 1.74e+00 & 4.11e-01 & \statsimilar 1.45e+01 & \statsimilar 1.77e+01 & \statsimilar 2.60e-03 & \statsimilar 3.15e-03 & \statsimilar 4.21e+02 & \statsimilar 3.17e+02 \\
    0.7 & 3.39e+00 & 1.55e+00 & 1.61e+00 & 4.48e-01 & \statsimilar 1.47e+01 & \statsimilar 2.05e+01 & \statsimilar 3.17e-03 & \statsimilar 4.11e-03 & \best 3.40e+02 & \best 2.27e+02 \\
    0.8 & 3.99e+00 & 2.12e+00 & 1.87e+00 & 5.16e-01 & \statsimilar 1.57e+01 & \statsimilar 1.87e+01 & \statsimilar 2.24e-03 & \statsimilar 2.93e-03 & \statsimilar 5.02e+02 & \statsimilar 4.53e+02 \\
    0.9 & 5.19e+00 & 3.88e+00 & 1.94e+00 & 4.09e-01 & \statsimilar 1.47e+01 & \statsimilar 2.02e+01 & \statsimilar 4.91e-03 & \statsimilar 6.56e-03 & \statsimilar 3.48e+02 & \statsimilar 2.50e+02 \\
    1.0 & 1.18e+01 & 7.29e+00 & 1.96e+00 & 3.60e-01 & \statsimilar 1.54e+01 & \statsimilar 1.95e+01 & \statsimilar 2.52e-03 & \statsimilar 3.13e-03 & \statsimilar 5.11e+02 & \statsimilar 4.00e+02 \\
\bottomrule
\toprule
    \bfseries Method
    & \multicolumn{2}{c}{\bfseries StyblinskiTang7 (7)} 
    & \multicolumn{2}{c}{\bfseries Ackley10 (10)} 
    & \multicolumn{2}{c}{\bfseries Michalewicz10 (10)} 
    & \multicolumn{2}{c}{\bfseries Rosenbrock10 (10)} 
    & \multicolumn{2}{c}{\bfseries StyblinskiTang10 (10)} \\ 
    & \multicolumn{1}{c}{Median} & \multicolumn{1}{c}{MAD}
    & \multicolumn{1}{c}{Median} & \multicolumn{1}{c}{MAD}
    & \multicolumn{1}{c}{Median} & \multicolumn{1}{c}{MAD}
    & \multicolumn{1}{c}{Median} & \multicolumn{1}{c}{MAD}
    & \multicolumn{1}{c}{Median} & \multicolumn{1}{c}{MAD}  \\ \midrule
    0.0 & \statsimilar 3.69e+01 & \statsimilar 1.36e+01 & \statsimilar 1.19e+01 & \statsimilar 2.80e+00 & \statsimilar 5.46e+00 & \statsimilar 5.94e-01 & 2.18e+03 & 1.21e+03 & \statsimilar 6.28e+01 & \statsimilar 2.31e+01 \\
    0.1 & \statsimilar 3.89e+01 & \statsimilar 1.75e+01 & \best 1.11e+01 & \best 3.11e+00 & \statsimilar 5.65e+00 & \statsimilar 6.58e-01 & 1.77e+03 & 1.07e+03 & \statsimilar 7.01e+01 & \statsimilar 2.16e+01 \\
    0.2 & \statsimilar 3.39e+01 & \statsimilar 1.50e+01 & \statsimilar 1.22e+01 & \statsimilar 1.80e+00 & \statsimilar 5.52e+00 & \statsimilar 6.62e-01 & 1.90e+03 & 1.25e+03 & \statsimilar 6.55e+01 & \statsimilar 3.04e+01 \\
    0.3 & \statsimilar 3.39e+01 & \statsimilar 1.75e+01 & \statsimilar 1.24e+01 & \statsimilar 3.06e+00 & \statsimilar 5.66e+00 & \statsimilar 5.56e-01 & \statsimilar 1.31e+03 & \statsimilar 1.09e+03 & \statsimilar 6.26e+01 & \statsimilar 1.79e+01 \\
    0.4 & \best 3.08e+01 & \best 1.72e+01 & 1.32e+01 & 1.68e+00 & \best 5.43e+00 & \best 8.41e-01 & \statsimilar 1.35e+03 & \statsimilar 8.19e+02 & \statsimilar 6.26e+01 & \statsimilar 2.20e+01 \\
    AEGiS & \statsimilar 3.23e+01 & \statsimilar 1.69e+01 & 1.41e+01 & 2.83e+00 & \statsimilar 5.61e+00 & \statsimilar 5.60e-01 & \statsimilar 9.97e+02 & \statsimilar 5.96e+02 & \statsimilar 6.75e+01 & \statsimilar 2.31e+01 \\
    0.6 & \statsimilar 3.10e+01 & \statsimilar 1.73e+01 & 1.41e+01 & 2.07e+00 & \statsimilar 5.73e+00 & \statsimilar 4.28e-01 & \statsimilar 1.01e+03 & \statsimilar 4.93e+02 & \statsimilar 6.06e+01 & \statsimilar 2.81e+01 \\
    0.7 & \statsimilar 3.11e+01 & \statsimilar 1.81e+01 & 1.66e+01 & 2.23e+00 & \statsimilar 5.57e+00 & \statsimilar 4.39e-01 & \statsimilar 1.10e+03 & \statsimilar 7.36e+02 & \statsimilar 5.98e+01 & \statsimilar 2.15e+01 \\
    0.8 & \statsimilar 3.11e+01 & \statsimilar 1.77e+01 & 1.68e+01 & 2.08e+00 & \statsimilar 5.74e+00 & \statsimilar 5.77e-01 & \best 9.57e+02 & \best 7.16e+02 & \best 5.83e+01 & \best 2.18e+01 \\
    0.9 & \statsimilar 3.22e+01 & \statsimilar 1.77e+01 & 1.75e+01 & 1.41e+00 & \statsimilar 5.77e+00 & \statsimilar 5.40e-01 & \statsimilar 1.29e+03 & \statsimilar 9.01e+02 & \statsimilar 5.94e+01 & \statsimilar 2.15e+01 \\
    1.0 & \statsimilar 3.84e+01 & \statsimilar 1.30e+01 & 1.86e+01 & 6.40e-01 & \statsimilar 5.85e+00 & \statsimilar 4.42e-01 & \statsimilar 1.07e+03 & \statsimilar 7.34e+02 & \statsimilar 5.90e+01 & \statsimilar 2.27e+01 \\
\bottomrule
\end{tabular}
}
\label{tbl:tsprop_results_8}
\end{table}

\begin{table}[H]
\setlength{\tabcolsep}{2pt}
\sisetup{table-format=1.2e-1,table-number-alignment=center}
\caption{Tabulated results for $q = 16$ workers for 
$\gamma \in \{0, 0.1, \dots, 1\}$. The table shows the median log simple regret
(\emph{left}) and median absolute deviation from the median (MAD, \emph{right})
after 200 function evaluations across the 51 runs. The method with the lowest
median performance is shown in dark grey, with those with statistically
equivalent performance are shown in light grey.}
\resizebox{1\textwidth}{!}{%
\begin{tabular}{l Sz Sz Sz Sz Sz}
    \toprule
    \bfseries Method
    & \multicolumn{2}{c}{\bfseries Branin (2)} 
    & \multicolumn{2}{c}{\bfseries Eggholder (2)} 
    & \multicolumn{2}{c}{\bfseries GoldsteinPrice (2)} 
    & \multicolumn{2}{c}{\bfseries SixHumpCamel (2)} 
    & \multicolumn{2}{c}{\bfseries Hartmann3 (3)} \\ 
    & \multicolumn{1}{c}{Median} & \multicolumn{1}{c}{MAD}
    & \multicolumn{1}{c}{Median} & \multicolumn{1}{c}{MAD}
    & \multicolumn{1}{c}{Median} & \multicolumn{1}{c}{MAD}
    & \multicolumn{1}{c}{Median} & \multicolumn{1}{c}{MAD}
    & \multicolumn{1}{c}{Median} & \multicolumn{1}{c}{MAD}  \\ \midrule
    0.0 & 5.57e-03 & 5.89e-03 & \statsimilar 6.70e+01 & \statsimilar 8.02e+01 & 7.14e+00 & 8.27e+00 & 3.80e-03 & 3.63e-03 & 2.66e-03 & 2.61e-03 \\
    0.1 & \statsimilar 2.01e-05 & \statsimilar 2.23e-05 & \statsimilar 6.81e+01 & \statsimilar 1.13e+01 & 5.78e+00 & 5.54e+00 & 1.67e-05 & 2.07e-05 & \statsimilar 1.49e-04 & \statsimilar 1.61e-04 \\
    0.2 & \statsimilar 1.20e-05 & \statsimilar 1.13e-05 & \statsimilar 6.51e+01 & \statsimilar 3.07e+01 & 4.92e+00 & 4.59e+00 & \statsimilar 7.57e-06 & \statsimilar 9.49e-06 & \statsimilar 1.55e-04 & \statsimilar 1.71e-04 \\
    0.3 & \statsimilar 1.10e-05 & \statsimilar 1.29e-05 & \statsimilar 6.53e+01 & \statsimilar 3.61e+01 & \statsimilar 3.20e+00 & \statsimilar 3.73e+00 & \statsimilar 5.35e-06 & \statsimilar 6.14e-06 & \best 1.15e-04 & \best 1.44e-04 \\
    0.4 & \best 9.80e-06 & \best 9.11e-06 & \statsimilar 6.51e+01 & \statsimilar 1.97e+01 & 3.38e+00 & 3.58e+00 & \best 3.89e-06 & \best 4.71e-06 & \statsimilar 1.74e-04 & \statsimilar 1.91e-04 \\
    AEGiS & \statsimilar 1.28e-05 & \statsimilar 1.80e-05 & \statsimilar 6.53e+01 & \statsimilar 1.27e+01 & \statsimilar 1.56e+00 & \statsimilar 1.84e+00 & \statsimilar 5.82e-06 & \statsimilar 5.32e-06 & \statsimilar 1.93e-04 & \statsimilar 2.80e-04 \\
    0.6 & \statsimilar 1.10e-05 & \statsimilar 1.39e-05 & \statsimilar 6.60e+01 & \statsimilar 1.50e+01 & \statsimilar 1.66e+00 & \statsimilar 1.98e+00 & \statsimilar 6.89e-06 & \statsimilar 8.79e-06 & \statsimilar 1.35e-04 & \statsimilar 1.75e-04 \\
    0.7 & 3.74e-05 & 5.05e-05 & \statsimilar 6.51e+01 & \statsimilar 4.12e+01 & \best 8.82e-01 & \best 1.27e+00 & \statsimilar 5.54e-06 & \statsimilar 6.69e-06 & \statsimilar 2.07e-04 & \statsimilar 2.59e-04 \\
    0.8 & 4.98e-05 & 6.76e-05 & \best 6.51e+01 & \best 3.23e+01 & \statsimilar 1.50e+00 & \statsimilar 2.08e+00 & \statsimilar 5.95e-06 & \statsimilar 7.22e-06 & 4.03e-04 & 5.34e-04 \\
    0.9 & 4.40e-04 & 6.46e-04 & \statsimilar 6.52e+01 & \statsimilar 9.10e+00 & \statsimilar 1.15e+00 & \statsimilar 1.66e+00 & \statsimilar 4.92e-06 & \statsimilar 6.51e-06 & 9.59e-04 & 1.32e-03 \\
    1.0 & 1.43e-03 & 2.04e-03 & \statsimilar 6.52e+01 & \statsimilar 2.47e+01 & \statsimilar 2.45e+00 & \statsimilar 3.46e+00 & 2.98e-05 & 4.25e-05 & 1.47e-02 & 2.10e-02 \\
\bottomrule
\toprule
    \bfseries Method
    & \multicolumn{2}{c}{\bfseries Ackley5 (5)} 
    & \multicolumn{2}{c}{\bfseries Michalewicz5 (5)} 
    & \multicolumn{2}{c}{\bfseries StyblinskiTang5 (5)} 
    & \multicolumn{2}{c}{\bfseries Hartmann6 (6)} 
    & \multicolumn{2}{c}{\bfseries Rosenbrock7 (7)} \\ 
    & \multicolumn{1}{c}{Median} & \multicolumn{1}{c}{MAD}
    & \multicolumn{1}{c}{Median} & \multicolumn{1}{c}{MAD}
    & \multicolumn{1}{c}{Median} & \multicolumn{1}{c}{MAD}
    & \multicolumn{1}{c}{Median} & \multicolumn{1}{c}{MAD}
    & \multicolumn{1}{c}{Median} & \multicolumn{1}{c}{MAD}  \\ \midrule
    0.0 & \statsimilar 2.97e+00 & \statsimilar 8.93e-01 & \statsimilar 1.50e+00 & \statsimilar 6.17e-01 & \statsimilar 1.58e+01 & \statsimilar 1.96e+01 & \statsimilar 2.30e-03 & \statsimilar 2.74e-03 & 1.53e+03 & 8.24e+02 \\
    0.1 & \best 2.84e+00 & \best 8.76e-01 & \best 1.49e+00 & \best 5.01e-01 & \statsimilar 1.60e+01 & \statsimilar 1.90e+01 & \best 1.87e-03 & \best 2.23e-03 & 1.35e+03 & 9.36e+02 \\
    0.2 & \statsimilar 3.16e+00 & \statsimilar 1.03e+00 & \statsimilar 1.54e+00 & \statsimilar 5.19e-01 & \statsimilar 1.55e+01 & \statsimilar 7.75e+00 & \statsimilar 1.89e-03 & \statsimilar 2.28e-03 & 1.35e+03 & 8.08e+02 \\
    0.3 & \statsimilar 3.06e+00 & \statsimilar 8.66e-01 & \statsimilar 1.60e+00 & \statsimilar 4.88e-01 & \statsimilar 1.49e+01 & \statsimilar 3.01e+00 & \statsimilar 4.20e-03 & \statsimilar 5.80e-03 & 1.02e+03 & 5.18e+02 \\
    0.4 & 3.47e+00 & 9.78e-01 & \statsimilar 1.51e+00 & \statsimilar 4.74e-01 & \best 1.47e+01 & \best 1.84e+01 & \statsimilar 2.96e-03 & \statsimilar 3.84e-03 & 8.65e+02 & 6.87e+02 \\
    AEGiS & 3.39e+00 & 1.12e+00 & \statsimilar 1.62e+00 & \statsimilar 5.10e-01 & \statsimilar 1.56e+01 & \statsimilar 1.09e+01 & \statsimilar 3.21e-03 & \statsimilar 3.82e-03 & 7.38e+02 & 5.13e+02 \\
    0.6 & \statsimilar 3.43e+00 & \statsimilar 1.34e+00 & \statsimilar 1.66e+00 & \statsimilar 4.10e-01 & \statsimilar 1.64e+01 & \statsimilar 1.79e+01 & \statsimilar 3.57e-03 & \statsimilar 4.54e-03 & \statsimilar 5.17e+02 & \statsimilar 2.91e+02 \\
    0.7 & 3.38e+00 & 1.47e+00 & \statsimilar 1.63e+00 & \statsimilar 5.68e-01 & \statsimilar 1.52e+01 & \statsimilar 1.76e+01 & 4.13e-03 & 5.56e-03 & \statsimilar 5.78e+02 & \statsimilar 3.51e+02 \\
    0.8 & 3.99e+00 & 2.05e+00 & \statsimilar 1.83e+00 & \statsimilar 4.57e-01 & \statsimilar 1.51e+01 & \statsimilar 1.77e+01 & 1.07e-02 & 1.51e-02 & \statsimilar 5.00e+02 & \statsimilar 2.98e+02 \\
    0.9 & 5.07e+00 & 2.74e+00 & \statsimilar 1.75e+00 & \statsimilar 4.60e-01 & \statsimilar 1.52e+01 & \statsimilar 1.64e+00 & \statsimilar 3.70e-03 & \statsimilar 4.74e-03 & \best 4.28e+02 & \best 2.70e+02 \\
    1.0 & 1.02e+01 & 7.76e+00 & 2.05e+00 & 3.46e-01 & \statsimilar 1.60e+01 & \statsimilar 1.33e+01 & 8.71e-03 & 1.21e-02 & \statsimilar 5.02e+02 & \statsimilar 4.58e+02 \\
\bottomrule
\toprule
    \bfseries Method
    & \multicolumn{2}{c}{\bfseries StyblinskiTang7 (7)} 
    & \multicolumn{2}{c}{\bfseries Ackley10 (10)} 
    & \multicolumn{2}{c}{\bfseries Michalewicz10 (10)} 
    & \multicolumn{2}{c}{\bfseries Rosenbrock10 (10)} 
    & \multicolumn{2}{c}{\bfseries StyblinskiTang10 (10)} \\ 
    & \multicolumn{1}{c}{Median} & \multicolumn{1}{c}{MAD}
    & \multicolumn{1}{c}{Median} & \multicolumn{1}{c}{MAD}
    & \multicolumn{1}{c}{Median} & \multicolumn{1}{c}{MAD}
    & \multicolumn{1}{c}{Median} & \multicolumn{1}{c}{MAD}
    & \multicolumn{1}{c}{Median} & \multicolumn{1}{c}{MAD}  \\ \midrule
    0.0 & \statsimilar 4.27e+01 & \statsimilar 1.87e+01 & \statsimilar 1.24e+01 & \statsimilar 1.73e+00 & \statsimilar 5.59e+00 & \statsimilar 5.43e-01 & 2.54e+03 & 1.89e+03 & \statsimilar 7.53e+01 & \statsimilar 2.26e+01 \\
    0.1 & \statsimilar 3.67e+01 & \statsimilar 1.36e+01 & \best 1.16e+01 & \best 2.69e+00 & \best 5.54e+00 & \best 4.19e-01 & 2.87e+03 & 1.62e+03 & \statsimilar 7.16e+01 & \statsimilar 2.51e+01 \\
    0.2 & \statsimilar 4.31e+01 & \statsimilar 1.74e+01 & 1.29e+01 & 2.29e+00 & \statsimilar 5.56e+00 & \statsimilar 5.68e-01 & 2.38e+03 & 1.22e+03 & \statsimilar 7.49e+01 & \statsimilar 2.53e+01 \\
    0.3 & \statsimilar 3.55e+01 & \statsimilar 1.44e+01 & 1.29e+01 & 3.09e+00 & \statsimilar 5.61e+00 & \statsimilar 5.26e-01 & 2.24e+03 & 1.21e+03 & \best 6.76e+01 & \best 3.26e+01 \\
    0.4 & \best 3.48e+01 & \best 1.55e+01 & 1.36e+01 & 2.14e+00 & \statsimilar 5.55e+00 & \statsimilar 7.22e-01 & 1.48e+03 & 8.46e+02 & \statsimilar 7.11e+01 & \statsimilar 2.29e+01 \\
    AEGiS & \statsimilar 4.32e+01 & \statsimilar 1.80e+01 & 1.46e+01 & 2.22e+00 & \statsimilar 5.74e+00 & \statsimilar 5.74e-01 & 1.60e+03 & 1.01e+03 & \statsimilar 7.20e+01 & \statsimilar 2.83e+01 \\
    0.6 & \statsimilar 4.37e+01 & \statsimilar 1.98e+01 & 1.54e+01 & 1.85e+00 & \statsimilar 5.70e+00 & \statsimilar 6.21e-01 & \statsimilar 1.33e+03 & \statsimilar 6.92e+02 & \statsimilar 7.11e+01 & \statsimilar 1.95e+01 \\
    0.7 & \statsimilar 3.95e+01 & \statsimilar 1.75e+01 & 1.63e+01 & 1.99e+00 & 5.77e+00 & 5.68e-01 & \statsimilar 1.21e+03 & \statsimilar 7.68e+02 & \statsimilar 8.37e+01 & \statsimilar 3.66e+01 \\
    0.8 & \statsimilar 4.37e+01 & \statsimilar 2.09e+01 & 1.68e+01 & 1.74e+00 & 5.77e+00 & 4.51e-01 & \statsimilar 1.33e+03 & \statsimilar 8.95e+02 & \statsimilar 7.06e+01 & \statsimilar 2.13e+01 \\
    0.9 & \statsimilar 4.01e+01 & \statsimilar 1.62e+01 & 1.79e+01 & 1.09e+00 & \statsimilar 5.70e+00 & \statsimilar 6.43e-01 & \statsimilar 1.15e+03 & \statsimilar 7.28e+02 & \statsimilar 7.51e+01 & \statsimilar 3.21e+01 \\
    1.0 & \statsimilar 4.26e+01 & \statsimilar 1.41e+01 & 1.84e+01 & 1.13e+00 & \statsimilar 5.70e+00 & \statsimilar 5.35e-01 & \best 1.13e+03 & \best 5.77e+02 & \statsimilar 7.81e+01 & \statsimilar 2.74e+01 \\
\bottomrule
\end{tabular}
}
\label{tbl:tsprop_results_16}
\end{table}

\bibliography{de-ath_227}

%% file: algorithm.tex
\begin{algorithm} [t]
    \caption{AEGiS}
    \label{alg:ets}

    \begin{algorithmic}[1]
      \Statex\textbf{Inputs:} $\Data_t$: \small{Training data}; 
        $T$: \small{Evaluation budget}
        \smallskip

        \State{$M = |\Data_t|$}
        \For{$t = M+1, \dots, T$}
 
            \State{Condition model posterior on $\Data_t$;
            see~\eqref{eqn:gp:pred} and~\eqref{eqn:gp:var}}
            \label{alg:eS:fitgp}

            \State{$r \sim \Uniform(0, 1)$}
            
            \If{ $r < 1-(\epsilon_T + \epsilon_P)$ }
               \Comment{\small{Exploit surrogate model}}
                \State{$\xnext \gets \underset{\bx \in \mX}{\argmin} \,
                  \mu(\bx)$}
                \label{alg:optimise-mu}
            \ElsIf{$r < 1- \epsilon_P$}
                \Comment{\small{Thompson sample}}
                \State{$g \sim p(f \given \Data)$} \label{alg:ts-sample}
                \State{$\xnext \gets \underset{\bx \in \mX}{\argmin} \,
                  g(\bx)$} \label{alg:optimise-ts}
            \Else
            \Comment{\small{Approximate Pareto set selection}}
                    \State{$\Papprox \gets \mathtt{MOOptimise}_{\bx\in\mX}(\mu(\bx), \sigma^2(\bx))$}
                    \label{alg:eS:moo_optimise}
                    \State{$\xnext \gets \mathtt{randomChoice}(\Papprox)$}
                    \label{alg:eS:choice_PF}
            \EndIf%

            \State{Submit new job to evaluate $\xnext$}
            \While{all workers in use}
                 \State{Wait for a worker to finish}
            \EndWhile
            \State{$\Data_t \gets \Data_{t-1} \cup \{ (\xnext, f(\xnext)) \}$}
            \Comment{\small{Augment data}}

        \EndFor%
    \end{algorithmic}
\end{algorithm}